\documentclass{article} %
\usepackage{iclr2027_conference,times}

\usepackage[hidelinks]{hyperref}
\usepackage{url}
\usepackage{booktabs}
\usepackage{multirow}
\usepackage{subcaption} %
\usepackage{wrapfig}
\usepackage{minted}

\usepackage{amsmath,amsfonts, amsthm, amssymb, bm, bbm}
\usepackage{nicefrac}

\newcommand{\lmset}{\{\!\!\{}
\newcommand{\rmset}{\}\!\!\}}

\usepackage{tikz}
\usepackage{pgfplots}
\pgfplotsset{compat=1.18}
\usetikzlibrary{calc, shapes.geometric, shapes.misc, positioning, intersections, pgfplots.fillbetween, automata}

\definecolor{color1}{RGB}{1,133,50}
\definecolor{color2}{RGB}{205,50,50}
\definecolor{color3}{RGB}{41,134,204}
\definecolor{color4}{RGB}{200,145,28}
\definecolor{color5}{RGB}{140,50,190}

\newcommand{\tr}[2]{{$#1 {\scriptscriptstyle \!\pm #2}$}}
\newcommand{\trb}[2]{{$\mathbf{#1 {\scriptscriptstyle \!\pm #2}}$}}

\newtheorem{lemma}{Lemma}
\newtheorem{theorem}{Theorem}

\usepackage{xspace}

\title{Scone{\kern-0.05ex}:{\kern-0.25ex} S{\kern-0.1ex}ymmetr{\kern-0.18ex}y-aware Neural{\kern-0.05ex} Combinatoria\rlap{l} \\ Optimization for Graph Coloring}

\author{Paolo Pellizzoni \& Melanie Weber \\
Harvard University \\
\texttt{\{paolo\_pellizzoni,mweber\}@seas.harvard.edu} \\
}

\iclrfinalcopy %
\begin{document}

\maketitle
\lhead{Preprint}

\vspace{-1.8ex}
\begin{abstract}
\vspace{-0.5ex}
Graph coloring, which entails assigning colors to nodes such that no adjacent nodes have the same color, is a fundamental combinatorial optimization problem. 
In general, however, finding a node coloring with no monochromatic edges given a fixed palette is NP-hard.
This motivates learning-based methods that can learn to exploit recurring structure across instances to efficiently produce high-quality colorings.
Crucially, the validity of a coloring is invariant to permutations of the colors, and we show that exploiting this symmetry reduces the size of the effective state space over which the model has to learn its policy. At the same time, the coloring algorithm must be able to break some symmetries, for example to assign different colors to two adjacent nodes belonging to the same automorphism orbit. 
 In this paper, we introduce \textsc{Scone}, a model that uses a \textit{color-equivariant} graph neural network to construct colorings sequentially, allowing information from previous assignments to propagate and break node symmetries. 
\textsc{Scone} achieves perfect solution quality with low inference times across several graph coloring tasks, including on \texttt{sudoku\_extreme} and register allocations, and showcases generalization to graph families not encountered during training.
\end{abstract}

\section{Introduction}

Graph coloring is a fundamental combinatorial optimization problem, which, given a graph and a palette of colors, asks
for an assignment of colors to the nodes of the graph such that adjacent nodes receive different colors.
Given a palette of $k$ colors, deciding whether such
an assignment exists is NP-complete for every fixed $k\geq3$ \citep{garey1974some}. 
Moreover, computing the smallest number of colors required to obtain a conflict-free assignment, known as the chromatic number, is NP-hard. 
Graph coloring has applications in several practical
domains, such as register allocation, where a compiler has to choose how to assign the variables of a program to the computer registers \citep{chaitin1981register}, and timetabling problems, where conflicting activities must occupy distinct timeslots \citep{welsh1967upper, carter1996examination}. 
Sudoku puzzles can also be seen as graph coloring instances, with cells represented as nodes and digits as colors \citep{herzberg2007sudoku}.

Traditional exact approaches include integer programming \citep{mendez2006branch}, as well as SAT and constraint satisfaction problem (CSP) solvers \citep{heule2022cliques}. 
Moreover, classical heuristics, which are generally faster but provide no guarantee of finding a valid coloring, include constructive algorithms such as greedy coloring \citep{welsh1967upper} and \textsc{Dsatur} \citep{brelaz1979new}, as well as local-search methods \citep{hertz1987using,malaguti2010survey}. 

Intuitively, when instances are drawn from a distribution or feature recurring structure, machine-learning-based algorithms can learn to exploit these regularities and produce high-quality solutions efficiently. More generally, neural combinatorial optimization uses learned models as heuristics for constrained optimization problems like the traveling salesman \citep{bello2016neural,khalil2017learning} and knapsack \citep{drakulic2023bq}. When trained on instance distributions, these models amortize the cost of learning a heuristic and enable fast inference on new instances \citep{bengio2021machine}. 

\textbf{Related Work}$\ $ For graph coloring, neural approaches can optimize instances directly or learn across a distribution. \textsc{PI-SAGE} \citep{schuetz2022graph} casts coloring as unsupervised node classification, using a graph neural network (GNN) to produce soft color assignments, which are then rounded.
Building on \textsc{PI-SAGE}, \citet{vanderbush2026neural} improve approximate $k$-coloring using recursive warm starts obtained from $(k\!-\!1)$-colorings. \textsc{ROS} \citep{qiu2025ros} similarly relaxes assignments to continuous ones, optimizes a continuous objective, and samples discrete solutions. %
Other methods learn general procedures across instances. 
\textsc{AnyCSP} \citep{tonshoff2023anycsp} trains a GNN with reinforcement learning for solving general CSPs by performing global stochastic updates.
Similarly, \textsc{ConsFormer} \citep{xu2025selfsupervised} trains a Transformer to improve complete assignments using differentiable relaxations of the constraints, as a learned form of local search. 
\citet{le2026contrastive} learn color representations with a contrastive loss and decode them by clustering.
\citet{li2022rethinking} also propose a color-equivariant model using restricted linear equivariant updates, but predict all node colors simultaneously, without conditioning each assignment on previous ones as in sequential decoding. Instead, \citet{watkins2023generating} propose a sequential constructive heuristic that learns node selection, while a fixed rule assigns the lowest permissible color.
See Appendix~\ref{sec:related} for additional related work.

Graph tasks often exhibit symmetries \citep{gilmer2017neural}. Most commonly, if two graphs are isomorphic via a node permutation, node-level outputs should permute accordingly. In graph coloring, applying the same node permutation to a graph and to a valid coloring preserves validity. Importantly, however, an individual coloring isn't \textit{invariant} under node permutations, since nodes in the same automorphism orbit may receive different colors. A solver must therefore be able to break node symmetries.
Moreover, graph coloring exhibits a symmetry over its \textit{output}. Color labels are interchangeable, since permuting them transforms any coloring into an equivalent one. These observations suggest that a learned solver should exploit both forms of symmetry to learn efficiently.

\textbf{Contributions}$\ $
We introduce \textsc{Scone} (\textbf{S}ymmetry-aware \textbf{CO}loring via \textbf{N}eural \textbf{E}quivariance), a model that constructs colorings sequentially, one node at a time. At each step, it takes as input the current partial coloring, allowing information from previous assignments to propagate through the graph, which distinguishes any nodes that were previously symmetric. Node colors are scored by a \textit{color-equivariant} GNN, so relabeling the colors permutes the predictions accordingly without changing the behavior of the model. We show theoretically that equivariance reduces the state space that the model must learn (Theorem~\ref{thm:full-reduction}), and that sequential decoding yields expressive models (Theorems~ \ref{thm:express}, \ref{thm:dsatur}). 
In summary, \textsc{Scone} combines sequential predictions, equivariance and classical search for expressivity and efficiency.
We evaluate \textsc{Scone} on real-world coloring applications, including Sudoku, register allocation, and exam timetabling. 
Here, it achieves state-of-the-art solution quality among neural methods, 
achieving the best solution quality while also having the lowest solve times.

\section{Preliminaries}

A graph is a tuple $G=(V_G,E_G)$, where $V_G = \{ 1, \dots, |V_G|\}$ is a set of nodes and
$E_G \subseteq \{\{u,v\}:u,v\in V_G,\ u\neq v\}$
is a set of undirected edges. We denote by $\mathcal N_G(u)$ the set of neighbors of $u$.

\textbf{Graph coloring}$\ $ 
Let $\mathcal C=\{1,\dots,k\}$ be a palette of $k$ colors, and let $\varepsilon$ denote the absence of a color. A partial $k$-coloring of $G$ is a map
$c:V_G\rightarrow \mathcal C\cup\{\varepsilon\}$.
We refer to $(G,c)$ as a $k$-coloring instance. A complete coloring $c^\star:V_G\rightarrow\mathcal C$ is a \emph{proper coloring} if
$c^\star(u)\neq c^\star(v)
\ \forall\,\{u,v\}\in E_G$.
It is a valid solution of $(G,c)$ if it is proper and extends the coloring $c$, i.e.,
$c(u)\!\neq\!\varepsilon \Rightarrow c^\star(u)\!=\!c(u)$. A graph is $k$-colorable if it has a proper $k$-coloring, and its chromatic number $\chi(G)$ is the smallest such $k$. %

\textbf{Groups and permutations} $\ $ 
For group $\Gamma$, let $\gamma\cdot x$ be the action of $\gamma\in\Gamma$ on $x$. Its
$\Gamma$-\emph{orbit} is
$\Gamma\cdot x=\{\gamma\cdot x:\gamma\in\Gamma\}$, and
$\mathcal X/\Gamma$ denotes the set of orbits.
A function is $\Gamma$-\emph{equivariant} if
$f(\gamma\cdot x)=\gamma\cdot f(x)$ for all $\gamma\in\Gamma$ and $x\in\mathcal X$.
We denote by $S_m$ the group of permutations of $\{1,\dots,m\}$. For $\tau\in S_m$, let
$P_\tau\in\{0,1\}^{m\times m}$ be its permutation matrix, such that $(P_\tau x)_i=x_{\tau^{-1}(i)}$.
A node permutation $\pi$ acts on graphs by relabeling their nodes as
$\pi\cdot G=\left(V_G,\{\{\pi(u),\pi(v)\}:\{u,v\}\in E_G\}\right)$,
and on a coloring as $(\pi\cdot c)(\pi(u))=c(u), \forall u$.
Two coloring instances are isomorphic if one is obtained from the other by such a node permutation $\pi$, written
$(G,c)\simeq(G',c')$. If such a $\pi$ satisfies $\pi(u)=u'$, then
$u\in V_G$ and $u'\in V_{G'}$ have the same \textit{rooted isomorphism} type.
We denote by
$\operatorname{Aut}(G,c)=\{\pi:\pi\cdot G=G,\ \pi\cdot c=c\}$
the automorphism group of $(G,c)$. Its orbits on $V_G$ are called node orbits. Intuitively, two nodes share an orbit if an isomorphism from $(G,c)$ to itself maps one to the other.

\section{Graph Coloring and Symmetries} \label{sec:symmetries}

Our goal is to design a machine learning model that colors a graph from a fixed palette while minimizing monochromatic edges, ideally producing a proper coloring.
This requires exploiting the symmetries of the task for sample efficiency, while breaking them when necessary for expressivity.

\vspace{-0.1ex}

\subsection{Node permutation symmetries and when to break them} \label{sub:node-symm}

Consider a graph coloring instance $(G, c)$ and a valid solution (i.e., a valid coloring completing $c$) $c^* : V_G \rightarrow \mathcal{C}$.
Then, if another graph $(G', c')$ is isomorphic to $(G, c)$ via an isomorphism $\pi$, the coloring $\pi \cdot c^*$ is a valid solution for $(G', c')$. We therefore want to use an architecture that is equivariant to permutations of the nodes, like a message-passing GNN \citep{gilmer2017neural}. 

The most immediate strategy for graph coloring with a model is to treat color assignment as a multiclass classification task with labels in $\mathcal{C}$. We would predict the color scores simultaneously for each node, and then select the argmax color for each node independently from each other, as if it was a node classification task. 
However, as noted also in \citet{li2022rethinking}, this strategy has a fundamental limitation, as nodes in the same node orbit of $(G, c)$ are assigned the same color even if neighboring.

\begin{figure}[t]
    \centering
    \includegraphics[width=\linewidth]{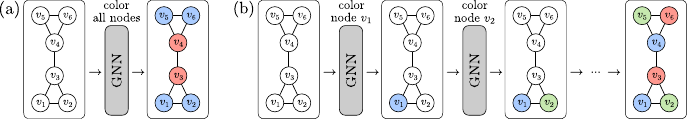} 
    \caption{ \textbf{(a)} Nodes $v_1$ and $v_2$ are symmetric and thus indistinguishable, so a deterministic model that predicts colors simultaneously for each node must assign them the same color, even if adjacent. \textbf{(b)} \textsc{Scone} sequentially colors one node at a time, conditioned on previous assignments. Once $v_1$ has been colored, $v_2$ can be distinguished from it, allowing the model to assign a distinct color to $v_2$.}
    \label{fig:symmetries}
    \vspace{-1.5ex}
\end{figure}

Consider, for example, the adjacent nodes $v_1$ and $v_2$ in Figure~\ref{fig:symmetries}~(a). Since they belong to the same orbit, any node-equivariant predictor must assign them the same vector of color scores, even if it can distinguish arbitrary rooted-graph isomorphism types.
Therefore, an adversary can always break any score ties so that these nodes are assigned the same color, thus failing to yield a proper coloring.

Therefore, although the predictor itself remains permutation-equivariant, the decoding procedure must be able to \textit{break} symmetries. We do so by constructing the solution sequentially, coloring one node at a time and recomputing the scores after every assignment, see Fig.~\ref{fig:symmetries}~(b). 
When automorphic nodes are tied, selecting any one of them breaks the symmetry of the resulting state, allowing previously indistinguishable nodes to become distinguishable. %
In particular, given the graph and a partial coloring, the model produces color scores to every uncolored node, selects one node, and updates the coloring by assigning it the highest-scoring color. The process ends when all nodes are colored.

More formally, given a $k$-coloring instance $(G, c_0)$, we define the state space $\mathcal{S}_{(G, c_0)}$ as the set of instances (up to isomorphism) that can be reached by coloring some uncolored nodes of $G$, and the set of extendable states $\mathcal{S}^+_{(G, c_0)}$ as states that can lead to a proper coloring $c^\star$ of $(G, c_0)$, i.e. \vspace{-0.1ex}
\[
\mathcal{S}_{(G, c_0)} = \big\{ (G, c) : \, c_0(v) \neq \varepsilon \Rightarrow c(v)=c_0(v) \ \forall v \in V_G \big\} /\!\simeq \vspace{-0.1ex}
\]
\[
\mathcal{S}^+_{(G, c_0)} = \big\{ (G, c) : \, \exists \text{ valid solution } c^\star \text{ of } (G, c_0) \text{ s.t. } c(v)\neq\varepsilon \Rightarrow c(v)=c^\star(v) \ \forall v \in V_G \big\} /\!\simeq .
\]
From each state $(G, c) \in \mathcal{S}_{(G, c_0)}$, the policy assigns to each uncolored node a score for each color. Based on these scores the policy selects a single uncolored node $u \in V_G$ and assigns a color $i$ to this node (breaking ties arbitrarily), updating the state to $(G, c')$, with $c'(u) \!=\! i$ and $c'(v) \!=\! c(v) \, \forall v \neq u$.

Supposing that the color predictor can distinguish any \textit{rooted isomorphism}, a single parameter setting can then express every $k$-colorable graph of bounded size. %
At each extendable state, the predictor can memorize, for every
uncolored node, the colors leading to an extendable successor.
Since every uncolored node has at least one such color, selecting any
node and any such color preserves extendability, so induction
produces a proper coloring.
(Thm.~\ref{thm:sequential-universality}). While practical models generally cannot distinguish arbitrary rooted-graph isomorphism types \citep{morris2019weisfeiler}, this shows that, at least in principle, sequential coloring has enough expressivity to represent valid coloring strategies.

\subsection{Color symmetries}

\begin{figure}[t]
    \centering
    \includegraphics[width=\linewidth]{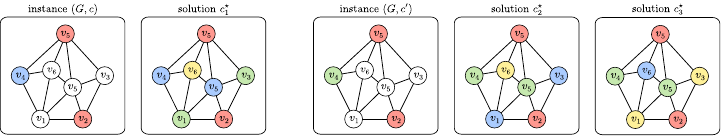}
    \caption{Graph coloring is equivariant to color permutations. The instances $(G,c)$ and $(G,c')$ differ only by exchanging blue and green, so the solutions $c_1^\star$ and $c_2^\star$ are related by the same permutation. Moreover, blue and yellow are unused in $c'$ and thus interchangeable, so $c_2^\star$ and $c_3^\star$ are equivalent. %
    }
    \label{fig:symmetries2}
    \vspace{-1ex}
\end{figure}

While the equivariance to permutations of the nodes is common in most graph tasks, graph coloring presents another symmetry: the color labels are arbitrary. Indeed, consistently permuting the color labels leaves the validity and the number of monochromatic edges of a coloring unchanged. Figure~\ref{fig:symmetries2} shows two instances that are exactly the same up to a permutation of the partial colors, and illustrates that their solutions are also the same up to the same permutation. 
Formally, let a color permutation $\tau \in S_k$ act as $(\tau \cdot c)(v) = \tau(c(v))$ if $c(v) \neq \varepsilon$ and $(\tau \cdot c)(v) = \varepsilon$ otherwise. 
If $c^*$ is a valid solution for $(G, c)$, then for any $\tau \in S_k$ 
we have that $\tau \cdot c^* $ is a valid solution for $(G, \tau \cdot c)$. 

We therefore seek a \emph{color-equivariant} model in order to exploit the symmetries of the problem. 
Intuitively, if the model outputs the highest score for color blue on node $v_6$ for the instance $(G, c)$ in Figure~\ref{fig:symmetries2}, then the model should output exactly the same score for color green on node $v_6$ for the instance $(G, c')$.
Let $p_u(G,c)\in\mathbb{R}^k$ denote the color scores for an uncolored node $u$. For every color permutation $\tau\in S_k$, we require
$p_u(G,\tau\cdot c)=P_\tau p_u(G,c)$.
Importantly, color equivariance couples the policy across states that differ only by a permutation of the color labels. If $s'=\tau\cdot s$, then the action scores at $s'$ are determined by those at $s$, up to the same permutation $\tau$. Therefore, learning the policy on one representative determines its behavior on the entire orbit $\{\tau\cdot s:\tau\in S_k\}$, reducing the size of the effective state space.
Moreover, color equivariance does not hinder the model when several colors are symmetric, as the policy will select any one of the tied colors arbitrarily.

For an instance distribution supported on a set $\mathcal I$ closed under color permutations, define the corresponding state space as
$\mathcal{S}_{\mathcal I} = \bigcup_{(G,c)\in\mathcal I}\mathcal{S}_{(G,c)}$.
Then, the effective state space of a color-equivariant model is the quotient $\mathcal{S}_{\mathcal I}/S_k$, while an unconstrained model may need to learn the appropriate behavior separately for all states in $\mathcal{S}_{\mathcal I}$, including states belonging to the same equivalence class. %
In fact, for the purpose of solving a feasible instance exactly, it is sufficient for the policy to learn how to act on extendable states. Indeed, if every choice preserves extendability, the decoding trajectory remains in $\mathcal{S}^+_{\mathcal I}$ until it reaches a proper coloring. However, non-extendable states are still relevant when considering %
approximate solutions or suboptimal decisions, so we consider both $\mathcal{S}_{\mathcal I}$ and $\mathcal{S}^+_{\mathcal I}$.

Our goal is thus to quantify the reduction in effective state space induced by color equivariance. We begin by bounding how many states related by color permutations form a single equivalence class.

\begin{lemma} \label{lem:fixed-reduction}
    Consider a state $(G, c) \in \bigcup_{\tau \in S_k} \mathcal{S}_{\tau \cdot (G, c_0)}$ with exactly $r$ colors appearing in $c$. Let $R(G, c) = |S_k\cdot(G,c)|$ denote the size of the orbit of $(G,c)$ under color permutations. Then,  \vspace{-0.2ex}
    { 
    \[
    \frac{k!}{(k-r)! \, \min(|{\rm Aut}(G)|, r!)} \leq R(G, c) \leq \frac{k!}{(k-r)!}.
    \] }
\end{lemma}

Intuitively, there are exactly $k!/(k-r)!$ injective ways to relabel the $r$ colors currently used by $c$, and graph automorphisms can cause some of these relabelings to yield the same state up to isomorphism.
For example, for standard $9\times9$ Sudoku problems, we have $k=9$. Any uniquely solvable instance contains at least eight distinct fixed numbers, i.e. $r \in \{ 8, 9 \}$. A careful application of the lemma yields $R(G,c)\ge 9!/648=560$ \citep{mcguire2014there}. Since this is valid for any extendable state, color equivariance reduces the number of distinct states by a factor $\smash{|\mathcal{S}^+_\mathcal{I}| / |\mathcal{S}^+_\mathcal{I}/ S_k|}$ of at least $560$.

In general, however, no nontrivial lower bound on $r$ holds uniformly over the state space. For instance, register-allocation and timetabling instances typically begin with no assigned colors, so early decoding states may use only one color. Applying the lemma uniformly over all states would therefore use the worst case $r=1$, yielding only an overly pessimistic orbit-size bound of order $k$. 
The following more nuanced analysis, which exploits the fact that states using few colors are rare, yields a much stronger lower bound on the state space reduction, which is at least exponential in $k$.

\begin{theorem} \label{thm:full-reduction}
    Consider a finite $k$-colorability instance set $\mathcal{I}$ with no starting colors. Let $|{\rm Aut}(G)| \leq B \ \forall G \in \mathcal{I}$. Let $\mathcal{S_I}$ be its associated state space and $\mathcal{S}^+_{\mathcal I}$ the space of extendable states. Then:
    \begin{enumerate}
        \item If all graphs $G \in \mathcal{I}$ have more than $k$ nodes, then $|\mathcal{S_I} / S_k | \leq B \, \smash{\dfrac{16}{e^k}} \, |\mathcal{S_I}|  \,$.
        \item If all graphs $G \in \mathcal{I}$ satisfy $0 \leq k - \chi(G) \leq \kappa$, then $|\mathcal{S}^+_{\mathcal I} / S_k | \leq B  \, \smash{\dfrac{ \kappa! \, 2^{k - \kappa } }{ k! }} \, |\mathcal{S}^+_{\mathcal I}| \, $.
    \end{enumerate}
\end{theorem}

Thus, for graph families with bounded $B$, color equivariance yields an exponential reduction over the full state space, and an even stronger reduction over the extendable states when $k$ is close to the chromatic numbers. This reduction in effective state space suggests a corresponding improvement in sample efficiency, as the model needs to infer appropriate behavior over substantially fewer states.

\section{\textsc{Scone}} \label{sec:model}

In this section, we describe the design and theoretical properties of \textsc{Scone}, which follows the sequential framework introduced in the previous section. Starting from an initial partial coloring, \textsc{Scone} constructs a complete coloring one node at a time. 
At each step, the color predictor produces color scores for every uncolored node. 
A decoding rule then selects one uncolored node and assigns a color according to that node's scores.
The resulting partial coloring becomes the next state. %

The main component of \textsc{Scone} is its color predictor, which we require to be equivariant to both node and color permutations. While message-passing GNNs naturally provide node equivariance, enforcing color equivariance requires that, for every parameter setting, input $(G,c)$, and $\tau\in S_k$,
${\rm GNN}(G,\tau\cdot c)=\tau\cdot{\rm GNN}(G,c)$.
Standard GNN constructions do not enforce this property. For example, representing color as one-hot encodings followed by unconstrained linear or MLP layers assign distinct parameters to each color, so permuting colors does not necessarily 
permute the output scores accordingly.
We therefore design a predictor that is color-equivariant by construction.

\subsection{A color-equivariant message-passing architecture}

Our color predictor is based on a message-passing mechanism. 
Each node $u$ will maintain, instead of a single embedding $h^\ell_u \in \mathbb{R}^d$, a collection of $k$ embeddings, one per color, as a matrix $H^\ell_u = (h^\ell_{u, 1}, \dots, h^\ell_{u, k}) \in \mathbb{R}^{k \times d}$. 
Each slot $h^\ell_{u, i}$ represents the state of node $u$ relative to color $i$.

We initialize the embeddings of all nodes with color $i$ with a vector $e_{\rm on}$ in the $i$-th slot and $e_{\rm off}$ in all other slots, while uncolored nodes receive $e_{\rm unk}$ in every slot. Formally, if $c(u)=\varepsilon$, then $H^0_u=(e_{\rm unk},\dots,e_{\rm unk})$, whereas if $c(u)\neq\varepsilon$, then $H^0_{u,c(u)}=e_{\rm on}$ and $H^0_{u,i}=e_{\rm off}$ for all $i\neq c(u)$. The vectors $e_{\rm on}$, $e_{\rm off}$, and $e_{\rm unk}$ are learned parameters. This initialization is color-equivariant by construction, as for every $\tau\in S_k$,
$H^0_u(G,\tau\cdot c)=P_\tau H^0_u(G,c)$.

As in standard message-passing GNNs, we aggregate the embeddings of neighboring nodes, i.e.,
\[
\hat H^\ell_u := f_{\rm agg}( \lmset H^{\ell-1}_v : \ v \in \mathcal{N}(u) \rmset ) \ = \bigoplus_{v \in \mathcal{N}(u)} H^{\ell-1}_v \ \in \mathbb{R}^{k \times d}. \vspace{-1ex} 
\]
Moreover, common element-wise aggregators $\oplus$ such as sum, mean, and max commute with permutations of the color slots, and are therefore $S_k$-equivariant.

In a traditional GNN one would then process the aggregated messages to produce the new embeddings via the update function $f_{\rm upd}$, which is typically an MLP. However, this is in general not $S_k$-equivariant, and therefore we process them with DeepSet-like layers \citep{zaheer2017deep}. In particular we can build an $S_k$-equivariant linear map $L: \mathbb{R}^{k \times d_1} \rightarrow \mathbb{R}^{k \times d_2}$ with the form 
\vspace{-0.5ex}
\[ 
    L(H)[i] = W_{\rm self}\, H[i] + W_{\rm all}\, \sum\nolimits_{j=1}^k H[j] + b \quad \text{with} \quad W_{\rm self}, W_{\rm all} \in \mathbb{R}^{d_2 \times d_1}, b \in \mathbb{R}^{d_2}.
    \vspace{-0.5ex}
\]
Since the same parameters are used for every slot and the sum over all slots is permutation-invariant, $L$ is $S_k$-equivariant.
Then, we set ${\rm MLP}_{S_k}(H) = L_2\big( \sigma\big(L_1(H) \big) \big)$, with $\sigma$ a point-wise nonlinearity like ${\rm ReLU}$ or ${\rm SiLU}$. 
Then, our update function is $H^\ell_u := f_{\rm upd}\big( H^{\ell -1}_u, \hat H^{\ell}_u \big) = {\rm MLP}_{S_k}( H^{\ell -1}_u \| \hat H^{\ell}_u)$, where $\|$ denotes concatenation on the $d$-axis. 
This is guaranteed to be $S_k$-equivariant.

Moreover, we apply LayerNorm independently to each color slot, with shared parameters, and add a residual connection, as they have been shown to improve training stability \citep{luo2025can}, as 
\[
H^\ell_u = {\rm LayerNorm}_d\bigg( {\rm MLP}_{S_k}\bigg( H^{\ell -1}_u \bigg\| \bigoplus_{v \in \mathcal{N}(u)} H^{\ell-1}_v \bigg) \bigg) +  H^{\ell-1}_u.
\vspace{-1.5ex}
\]

Finally, after $\mathcal{L}$ message-passing layers, for each node $u$ we produce the color scores from the final layer embeddings $H^{\mathcal{L}}_u$ %
via a $S_k$-equivariant readout ${\rm MLP}_{\!S_k}\!: \mathbb{R}^{k \times d} \rightarrow \mathbb{R}^{k \times 1}$. %

All operations are color-equivariant, and they compose with the permutation equivariance acting on nodes typical of message-passing architectures. Therefore, we obtain the following result. 

\begin{theorem} \label{thm:equi}
    The color predictor of \textsc{Scone} is equivariant to the joint action of $S_{|V_G|}\!\times\! S_k$.  
    Let $(G,c)$ be any graph $k$-coloring instance, and let $p_u(G,c)\in\mathbb{R}^k$ be the output of the predictor for node $u$. Then, for every parameter setting, every $\pi\in S_{|V_G|}$, every $\tau\in S_k$, and every $u\in V_G$, \vspace{-0.5ex}
    \[
        p_{\pi(u)}\bigl(\pi\cdot G,\, \tau\cdot(\pi\cdot c)\bigr)
        =
        P_\tau\,p_u(G,c).
    \] 
\end{theorem}

Moreover, the parameterization of our color-equivariant message-passing architecture is independent of $k$. Since all transformations are shared across color slots, the same parameters apply to any value of $k$. This permits zero-shot evaluation on instances with palette sizes unseen during training.

\subsection{Symmetry-aware training} \label{sec:training}

We train \textsc{Scone} in a supervised fashion, as a node prediction task, from target proper colorings. Given a partial coloring $(G,c)$ and a target completion $c^\star$, a natural approach would be to train each uncolored node $u$ against the one-hot label corresponding to $c^\star(u)$. However, such labels can conflict with color symmetries that are still present in the current state.

Let $U(c)=\{i\in\mathcal{C}:\exists u\in V_G,\ c(u)=i\}$ and $F(c)=\mathcal C\setminus U(c)$ denote the used and free
colors, and let
$\operatorname{Stab}(c)=\{\tau\in S_k:\tau\cdot c=c\}$ be the stabilizer of $c$, i.e., all the color relabelings that leave the current partial coloring $c$ unchanged.
These permutations fix every used color and freely permute the remaining ones.
Color equivariance therefore implies
$p_u(G,c)=P_\tau p_u(G,c) \ \forall\,\tau\in\operatorname{Stab}(c),
$
so all free colors receive the same score. Hence, if
$c^\star(u)\in F(c)$ and $|F(c)|>1$, a one-hot target for $c^\star(u)$ cannot
be matched and produces an irreducible loss.

This ambiguity is intrinsic to the problem rather than a limitation of the architecture. Indeed, for every $\tau\in\operatorname{Stab}(c)$, the coloring $\tau\cdot c^\star$ is an equally valid completion of $(G,c)$. Figure~\ref{fig:symmetries2} illustrates an example of this phenomenon. There, since only red and green have appeared in $c'$, blue and yellow remain interchangeable, so completions obtained by swapping these two colors are equivalent.

While full symmetry-consistent supervision would also require averaging targets over automorphic nodes, doing so requires computing node orbits at each state, which can be expensive. %
In practice,
we therefore replace the one-hot target by its average over all color-equivalent completions, i.e., 
{ \small
\[
y_u(i)
:=
\frac{1}{|\operatorname{Stab}(c)|}
\sum_{\tau\in\operatorname{Stab}(c)}
\mathbbm{1}\!\left[i=\tau(c^\star(u))\right]
=
\begin{cases}
1, &                            \text{if } c^\star(u)\in U(c)\ \text{and}\ i=c^\star(u),\\
\smash{\nicefrac{1}{|F(c)|}}, & \text{if } c^\star(u)\in F(c)\ \text{and}\ i\in F(c),\\
0, &                            \text{otherwise}.
\end{cases}
\] 
}
\vspace{-2ex}

We train the predictor by minimizing the cross-entropy of the softmax of color scores with these symmetry-aware labels $y$. 
The loss supervises every uncolored node rather than only a node selected by an expert ordering. The predictor is thus trained to provide a useful color distribution for whichever node the decoder selects next, without requiring a separate node-selection objective.

\textbf{Ground truth construction}$\ $ For graphs drawn from a given distribution, we can obtain proper colorings using an exact solver, for example via integer linear programs (ILP). While computationally expensive, this is a one-time cost which is amortized by the much cheaper model at inference time. 

Moreover, there is an asymmetry in the difficulty of solving given $k$-coloring instances and \textit{generating} instances with valid solutions. Indeed, we can first draw a certain number of nodes, each with a color from $\mathcal{C}$, and then draw edges according to a distribution of interest, \textit{only} between nodes of different colors. Masking some or all of the assigned colors yields a $k$-coloring instance together with a proper coloring by construction, avoiding the need to solve the generated instance.

Given an instance $(G,c)$ and a target completion $c^\star$, we generate multiple training states by partially revealing $c^\star$. Specifically, for a subset $R$ of the uncolored nodes, we reveal $c^\star(u)$ for every $u\in R$, obtaining a partial coloring $c'$. All such instances $(G,c')$ belong to the state space $\mathcal{S}_{(G,c)}$, with small sets $R$ representing states near the beginning of the decoding process, and larger sets representing later stages. For a graph with $m$ uncolored nodes, this construction yields up to $2^m$ training states. %

\subsection{Inference strategies} \label{sec:inference}

We train our model to predict a correct color for \textit{every} uncolored node at any stage of the decoding process. Therefore, we could simply color the nodes in their input order. 
However, we find (see Appendix~\ref{sec:sweeps}) that in practice it is more effective to select nodes using the common \emph{most-constrained-first} principle \citep{haralick1980increasing} from constraint satisfaction. 
In particular, we select an uncolored node with the largest number of distinct colors already assigned to its neighbors. In case of ties, we select the node with the highest maximum softmax probability, as a proxy for confidence. 
Then, we assign to this node the color with the maximum score, breaking any ties arbitrarily.

Moreover, we can improve over greedy decoding using beam search \citep{pearl1984heuristics}. We maintain a beam of $B$ partial colorings and, at each step, expand each of them with multiple color assignments, retaining the $B$ states with highest cumulative score. This allows the model to maintain multiple decoding trajectories and recover from wrong choices made at earlier steps.
We further increase diversity by ensembling independent decodings from $E$ starting nodes, yielding up to $B\times E$ candidate colorings. If an instance is not solved, we can repeat the decoding at increasing width, e.g., by doubling $B$ or $E$. We return the coloring with the fewest monochromatic edges across all attempts.

Finally, we can avoid some model evaluations by directly propagating \textit{forced assignments}. If an uncolored node has only one color that would not create a monochromatic edge, we assign it without running the color predictor. We repeatedly apply this propagation rule until no forced assignment remains, and then resume model-based decoding.
When no conflict-free color is available for the selected node, \textsc{Scone} assigns a conflicting color, so it always produces a complete coloring. %

\subsection{\textsc{Scone} is expressive}

In Section~\ref{sub:node-symm}, we argued that sequential coloring is expressive in
principle. However, the message-passing-based predictor of \textsc{Scone} is limited
by $1$-WL at each step \citep{morris2019weisfeiler}, so it is interesting to characterize on which graph families the model can implement a policy solving them.

The $1$-WL algorithm iteratively updates each node's label from its current label and those of its neighbors. A graph is \emph{amenable} if its $1$-WL multiset distinguishes it from every non-isomorphic graph \citep{arvind2017graph}. We call a graph family \emph{cover-amenable} if no two non-isomorphic connected components occurring in the family have isomorphic universal covers, and each is amenable.

On cover-amenable families of graphs, stable rooted $1$-WL types determine rooted isomorphism types throughout decoding. This yields the following result. The condition holds, e.g., for families of trees or discrete graphs, which include almost all dense Erdos-Renyi graphs \citep{babai1980random}.

\begin{theorem}
\label{thm:express}
Let $n,k\in\mathbb N$, and let $\mathcal G$ be a cover-amenable family of graphs with at most $n$ nodes. There exists a parameter setting of \textsc{Scone} s.t. every $k$-colorable $(G,c_0)$ with $G\in\mathcal G$ can be solved.
\end{theorem}

Note that the result, as common in expressivity results, is in the non-uniform regime, i.e. the model works for graphs of size up to $n$, but might not generalize to larger graphs. 
However, for some classes of graphs, we can actually show that \textsc{Scone} can simulate classical algorithms for graph coloring, and therefore obtain \textit{uniform} results. In particular, we can show that \textsc{Scone} returns a proper coloring for all graphs in some graph families, even for ones of unbounded size.

\begin{theorem} \label{thm:dsatur}
    Let $k, \Delta \in \mathbb{N}$. There exists a parameter setting of \textsc{Scone} s.t. every $k$-colorable cycle,
chordal or bipartite graph with degree at most $\Delta$ can be properly colored from the empty coloring.
\end{theorem}

\section{Experimental Results}

In this section, we experimentally evaluate the solution quality and efficiency of \textsc{Scone}. We evaluate our method and baselines on both real-world graph coloring instances and on controlled synthetic data. 
Our code and the curated graph-coloring
benchmark datasets will be provided publicly. %

\textbf{Datasets}$\ $ We place particular emphasis on graph-coloring instances arising from practical applications. We consider Sudoku instances, register-allocation graphs, and timetabling graphs, which exhibit substantially different graph structures, numbers of colors, and partial-coloring patterns. In addition, we use structured synthetic graph families to study controlled settings in which particular symmetries or expressivity limitations become relevant. See Appendix~\ref{sec:datasets} for additional details. 

\textbf{Metrics}$\ $ We evaluate each method using the percentage of instances solved (Solved\%), the percentage of monochromatic edges in the returned colorings (Mono\%), and the average inference time per graph. The solve percentage shows how often a method returns a proper coloring, while the percentage of monochromatic edges measures the solution quality when a proper coloring is not found. Unless specified, we report averages and standard deviations over five independent repeats.

\textbf{Methods}$\ $ We compare \textsc{Scone} against both classical methods and neural solvers. As classical baselines, we consider %
the greedy, \textsc{Dsatur} and RLF \citep{leighton1979graph} heuristics, an ILP formulation with timelimit, and \textsc{ACE}, a frontier CSP solver.
We also compare against the learning-based \textsc{Ros} \citep{qiu2025ros}, \textsc{Pi-SAGE} \citep{schuetz2022graph}, \textsc{AnyCSP} \citep{tonshoff2023anycsp}, \textsc{ConsFormer} \citep{xu2025selfsupervised}, and \textsc{ReLCol} \citep{watkins2023generating}. 
For \textsc{Scone}, we fix the number of layers to $\mathcal{L}=10$ and the embedding dimension to $d=128$, which provides a good trade-off between solution quality and inference time. As aggregation $\oplus$, we use a $\smash{\sqrt{\mathcal{N}(u)}}$-normalized sum, which retains degree information while controlling the scale of the aggregated messages. We train the model using the AdamW optimizer (${\rm lr}=10^{-3}$). Checkpoints and the choice between beams and ensembling are selected on validation Solved\%. See also Appendix~\ref{sec:setup} for additional details. 
\vspace{-0.5ex}

\subsection{Sudokus}

\begin{table}[t!]
\centering \vspace{-0.1ex}
\resizebox{\textwidth}{!}{%
{\setlength{\tabcolsep}{6pt}
\begin{tabular}{l*{4}{ccc}}
\toprule
\multirow{2}{*}{Method} & \multicolumn{3}{c}{\texttt{sudoku\_17}} & \multicolumn{3}{c}{\texttt{sudoku\_satnet}} & \multicolumn{3}{c}{\texttt{sudoku\_rrn}} & \multicolumn{3}{c}{\texttt{sudoku\_top1465}} \\
\cmidrule(lr){2-4} \cmidrule(lr){5-7} \cmidrule(lr){8-10} \cmidrule(lr){11-13}
 & Solved\% & Mono\% & Time (s) & Solved\% & Mono\% & Time (s) & Solved\% & Mono\% & Time (s) & Solved\% & Mono\% & Time (s) \\
\midrule
\textsc{Greedy} & \tr{0.0}{0.0} & \tr{1.83}{0.00} & \tr{0.00}{0.00} & \tr{1.4}{0.0} & \tr{1.48}{0.00} & \tr{0.00}{0.00} & \tr{0.0}{0.0} & \tr{1.83}{0.00} & \tr{0.00}{0.00} & \tr{0.0}{0.0} & \tr{1.93}{0.00} & \tr{0.00}{0.00} \\
\textsc{DSatur} & \tr{0.1}{0.0} & \tr{0.91}{0.00} & \tr{0.00}{0.00} & \trb{100.0}{0.0} & \trb{0.00}{0.00} & \tr{0.00}{0.00} & \tr{32.0}{0.0} & \tr{0.63}{0.00} & \tr{0.00}{0.00} & \tr{0.7}{0.0} & \tr{0.88}{0.00} & \tr{0.00}{0.00} \\
\textsc{RLF} & \tr{0.0}{0.0} & \tr{0.94}{0.00} & \tr{0.00}{0.00} & \tr{7.0}{0.0} & \tr{0.77}{0.00} & \tr{0.00}{0.00} & \tr{0.1}{0.0} & \tr{1.10}{0.00} & \tr{0.00}{0.00} & \tr{0.0}{0.0} & \tr{1.05}{0.00} & \tr{0.00}{0.00} \\
\textsc{ILP} {\scriptsize (1s)} & \tr{97.6}{0.0} & \tr{0.01}{0.00} & \tr{0.72}{0.00} & \trb{100.0}{0.0} & \trb{0.00}{0.00} & \tr{0.02}{0.00} & \tr{99.7}{0.0} & \tr{0.00}{0.00} & \tr{0.24}{0.00} & \tr{97.9}{0.0} & \tr{0.01}{0.00} & \tr{0.59}{0.00} \\
\textsc{ACE} & \trb{100.0}{0.0} & \trb{0.00}{0.00} & \tr{0.51}{0.00} & \trb{100.0}{0.0} & \trb{0.00}{0.00} & \tr{0.51}{0.00} & \trb{100.0}{0.0} & \trb{0.00}{0.00} & \tr{0.51}{0.00} & \trb{100.0}{0.0} & \trb{0.00}{0.00} & \tr{0.53}{0.00} \\
\midrule
\textsc{AnyCSP} & \tr{26.5}{1.2} & \tr{0.18}{0.00} & \tr{4.37}{0.03} & \trb{100.0}{0.0} & \trb{0.00}{0.00} & \tr{0.04}{0.00} & \tr{91.3}{0.6} & \tr{0.02}{0.00} & \tr{0.87}{0.00} & \tr{82.7}{0.2} & \tr{0.04}{0.00} & \tr{1.99}{0.00} \\
\textsc{ConsFormer} & \tr{4.1}{0.5} & \tr{0.24}{0.00} & \tr{0.22}{0.00} & \trb{100.0}{0.0} & \trb{0.00}{0.00} & \tr{0.01}{0.01} & \tr{68.0}{0.3} & \tr{0.08}{0.00} & \tr{0.21}{0.00} & \tr{20.5}{1.1} & \tr{0.20}{0.00} & \tr{0.22}{0.00} \\
\textsc{ReLCol} & \tr{0.0}{0.0} & \tr{0.71}{0.00} & \tr{5.09}{0.00} & \tr{63.6}{0.8} & \tr{0.11}{0.00} & \tr{2.84}{0.03} & \tr{2.0}{0.0} & \tr{0.66}{0.00} & \tr{5.07}{0.01} & \tr{0.0}{0.0} & \tr{0.69}{0.00} & \tr{5.13}{0.00} \\
\midrule
\textsc{Ours} & \trb{100.0}{0.0} & \trb{0.00}{0.00} & \tr{0.01}{0.00} & \trb{100.0}{0.0} & \trb{0.00}{0.00} & \tr{0.00}{0.00} & \trb{100.0}{0.0} & \trb{0.00}{0.00} & \tr{0.00}{0.00} & \trb{100.0}{0.0} & \trb{0.00}{0.00} & \tr{0.02}{0.00} \\
\bottomrule
\end{tabular}%
}
}
\caption{Solution quality on Sudoku-graph coloring: Solved\% ($\uparrow$), Mono\% ($\downarrow$) and inference times.$\!\!$}
\label{tab:sudoku}
\vspace{-1.5ex}
\end{table}

Sudoku can be naturally formulated as a graph-coloring problem, with one node for each cell, edges connecting cells in the same row, column, or block, and digits represented as colors.
Our main dataset is \texttt{sudoku\_17}, which contains all known non-equivalent Sudoku puzzles with $17$ clues, split into $70\%/15\%/15\%$. %
We also evaluate our model (without retraining it) on three datasets of increasing difficulty, \texttt{satnet}, \texttt{rrn}, and \texttt{top1465}. Since \textsc{AnyCSP} does not natively support fixed colors, we extended it accordingly and retrained it. %
\textsc{Pi-SAGE} and \textsc{ROS} do not support fixed colors. \\
Table~\ref{tab:sudoku} shows that \textsc{Scone} performs strongly both in distribution and under increasingly difficult distribution shifts. \textsc{AnyCSP} is the strongest competing learning method, but its solution quality deteriorates on the harder instances and its iterative search is roughly two orders of magnitude slower. %
Classical heuristics struggle, while \textsc{ILP} and \textsc{ACE} have near-perfect quality, but are usually slower.

\begin{wrapfigure}{r}{0.69\textwidth}
\centering
\begin{minipage}{0.35\textwidth}
\centering
\vspace{-0.5ex}
\scalebox{0.8}{%
{\setlength{\tabcolsep}{6pt}
\begin{tabular}{l*{1}{cc}}
\toprule
\multirow{2}{*}{Method} & \multicolumn{2}{c}{\texttt{sudoku\_extreme}} \\
\cmidrule(lr){2-3}
 & Solved\% & Time (s) \\
\midrule
\textsc{AnyCSP} & \tr{79.03}{0.30} & \tr{2.02}{0.02} \\
\textsc{TRM} & \tr{94.87}{0.00} & \tr{0.02}{0.00} \\ 
\textsc{LDT} & \tr{99.95}{0.03} & \tr{0.12}{0.01} \\
\midrule
\textsc{Ours} & \trb{100.00}{0.00} & \tr{0.02}{0.00} \\
\bottomrule
\end{tabular}%
}
}
\captionof{table}{Solution quality and inference times on \texttt{sudoku\_extreme}}
\label{tab:lattice}
\end{minipage}
\hfill
\begin{minipage}{0.32\textwidth}
    \centering
    \vspace{-1.8ex}
    \includegraphics[width=\textwidth]{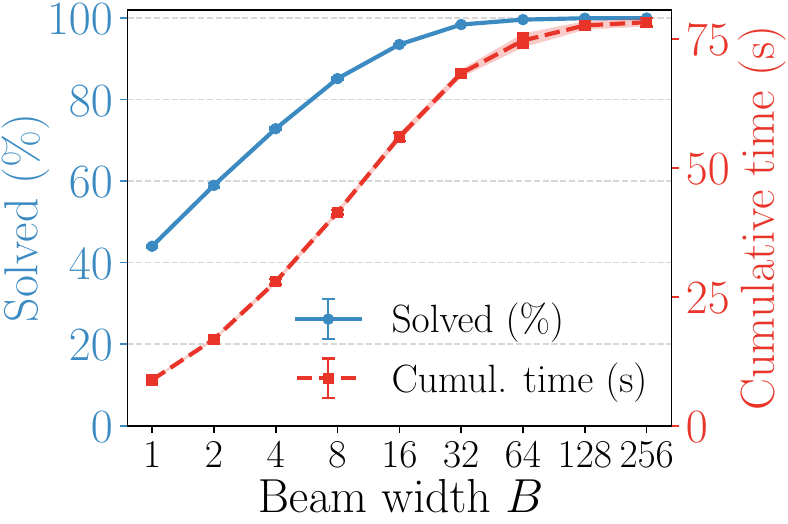} \vspace{-4.45ex}
    \captionof{figure}{Cumulative solutions and runtime with adaptive width}
    \label{fig:cumul_width}
\end{minipage}
\vspace{-2ex}
\end{wrapfigure}
Motivated by these results, we evaluate \textsc{Scone} on \texttt{sudoku\_extreme} \citep{wang2025hierarchicalreasoningmodel}, the most challenging public Sudoku benchmark.
Table~\ref{tab:lattice} compares it with the 
state-of-the-art reasoning models \textsc{LDT} \citep{davis2026lattice} and \textsc{TRM} \citep{jolicoeur2025less}, using \textsc{AnyCSP} as a reference point. We use the splits introduced by \textsc{LDT}, so we train all the models on 1000 training instances, and evaluate on 4000 test instances.
While the baselines leave some puzzles unsolved, \textsc{Scone} consistently solves every test instance, with only a few requiring as many as $256$ beams.
Moreover, despite being a general graph-coloring model, \textsc{Scone} is at least five times faster than LDT.
Figure~\ref{fig:cumul_width} shows that most instances are solved at small beam widths, as widths up to 32 solve 98.5\% of instances, while only a small fraction require wider search.
Adaptive widening therefore solves these harder instances with only a modest increase in total runtime.

\subsection{Register allocation graphs}

Register allocation can be formulated as graph coloring. Indeed, nodes represent program variables, edges connect two variables that are used simultaneously and thus cannot occupy the same register, and 
colors represent registers. A proper coloring gives a conflict-free register allocation. %

\begin{wraptable}{rt}{0.45\textwidth}
\centering
\vspace{-2ex}
\scalebox{0.74}{%
{\setlength{\tabcolsep}{6pt}
\begin{tabular}{l*{1}{ccc}}
\toprule
\multirow{2}{*}{Method} & \multicolumn{3}{c}{\texttt{registers}} \\
\cmidrule(lr){2-4}
 & Solved\% & Mono\% & Time (s) \\
\midrule
\textsc{Greedy} & \tr{90.1}{0.0} & \tr{0.09}{0.00} & \tr{0.00}{0.00} \\
\textsc{DSatur} & \tr{99.0}{0.0} & \tr{0.01}{0.00} & \tr{0.00}{0.00} \\
\textsc{RLF} & \tr{98.0}{0.0} & \tr{0.04}{0.00} & \tr{0.00}{0.00} \color{black} \\
\textsc{ILP} {\scriptsize (1s)} & \tr{99.7}{0.0} & \tr{0.01}{0.00} & \tr{0.02}{0.00} \\
\textsc{ACE} & \trb{100.0}{0.0} & \trb{0.00}{0.00} & \tr{0.41}{0.00} \\
\midrule
\textsc{PI-SAGE} & \tr{53.8}{2.0} & \tr{0.52}{0.02} & \tr{2.85}{0.01} \\
\textsc{ROS} & \tr{35.5}{1.0} & \tr{0.70}{0.01} & \tr{0.39}{0.00} \\
\textsc{AnyCSP} & \trb{100.0}{0.0} & \trb{0.00}{0.00} & \tr{0.02}{0.00} \\
\textsc{ConsFormer} & \tr{80.5}{0.8} & \tr{0.51}{0.01} & \tr{0.13}{0.01} \\ %
\textsc{ReLCol} & \tr{98.6}{0.1} & \tr{0.03}{0.00} & \tr{0.19}{0.00} \\
\midrule
\textsc{Ours} & \trb{100.0}{0.0} & \trb{0.00}{0.00} & \tr{0.00}{0.00} \\
\bottomrule
\end{tabular}%
}
}
\caption{Solution quality on \texttt{registers}: Solved\% ($\uparrow$), Mono ($\downarrow$) and inference times.}
\label{tab:registers21}
\vspace{-2.5ex}
\end{wraptable}

We evaluate on a filtered dataset of 8999 register-interference graphs collected by \citet{george1996iterated}, split $70\%/15\%/15\%$. See Appendix~\ref{sec:datasets} for additional details. Rather than using the original budget of $21$ registers, we set the palette size for each graph to its chromatic number, thereby testing whether each method can recover an \textit{optimal} coloring. As shown in Table~\ref{tab:registers21}, \textsc{ACE}, \textsc{AnyCSP} and \textsc{Scone} solve every test instance, while the other learning-based methods produce lower-quality solutions. 
\textsc{DSatur} and greedy also perform strongly, indicating that the structure of these graphs is well suited to constraint-aware node ordering.
Among the learning-based methods, \textsc{Scone} also has the lowest inference time, at approximately $4$ ms per graph.

\subsection{Generalization from synthetic training} \label{sec:synth_train}

Many important coloring benchmarks contain only a few isolated instances and therefore provide no natural training distribution. To evaluate transfer in this setting, we train a single model exclusively on synthetic data, as described in Sec.~\ref{sec:training}, and apply it without finetuning to multiple instance families. The Toronto timetabling graphs \citep{carter1996examination} represent exams as nodes, conflicts due to common students as edges, and timeslots as colors. We aggregate 4 smaller instances and 5 larger instances. 
We also consider queen graphs \citep{vasquez2004new}, for $p\!\in\!\{8,9,10\}$, where nodes represent squares of a $p\times p$ chessboard and edges connect squares sharing a row, column or diagonal.

Table~\ref{tab:toronto_queens_cycles}, using 20 repeats, shows that the synthetically trained model transfers reliably across these families, with \textsc{Scone} being the only learned method to solve every instance. On the smaller Toronto instances, \textsc{Scone} matches the 100\% solve rate of \textsc{AnyCSP} while being slightly faster. The gap widens on the larger instances, where \textsc{Scone} solves every graph, while \textsc{AnyCSP} fails on some and takes five times longer. On queen graphs, \textsc{Scone} and \textsc{AnyCSP} both solve all instances, with the former being more consistent and faster. Overall, \textsc{ACE} also achieves near-perfect solution quality.

We test the same model on structured synthetic families that expose potential failure modes, reporting cycle graphs here and full results in Table~\ref{tab:structured}. With greedy decoding alone, \textsc{Scone} solves every graph with fast inference times. \textsc{AnyCSP} also succeeds but is slower, while other models struggle.

\begin{table}[t!]
\centering
\resizebox{\textwidth}{!}{%
{\setlength{\tabcolsep}{6pt}
\begin{tabular}{l*{4}{ccc}}
\toprule
\multirow{2}{*}{Method} & \multicolumn{3}{c}{toronto (small)} & \multicolumn{3}{c}{toronto (large)} & \multicolumn{3}{c}{queens} & \multicolumn{3}{c}{cycles} \\
\cmidrule(lr){2-4} \cmidrule(lr){5-7} \cmidrule(lr){8-10} \cmidrule(lr){11-13}
 & Solved\% & Mono\% & Time (s) & Solved\% & Mono\% & Time (s) & Solved\% & Mono\% & Time (s) & Solved\% & Mono\% & Time (s) \\
\midrule
\textsc{Greedy} & \tr{0.0}{0.0} & \tr{0.40}{0.00} & \tr{0.00}{0.00} & \tr{0.0}{0.0} & \tr{0.35}{0.00} & \tr{0.01}{0.00} & \tr{0.0}{0.0} & \tr{1.44}{0.00} & \tr{0.00}{0.00} & \trb{100.0}{0.0} & \trb{0.00}{0.00} & \tr{0.00}{0.00} \\
\textsc{DSatur} & \tr{50.0}{0.0} & \tr{0.02}{0.00} & \tr{0.00}{0.00} & \tr{20.0}{0.0} & \tr{0.04}{0.00} & \tr{0.02}{0.00} & \tr{0.0}{0.0} & \tr{1.24}{0.00} & \tr{0.00}{0.00} & \trb{100.0}{0.0} & \trb{0.00}{0.00} & \tr{0.00}{0.00} \\
\textsc{RLF} & \tr{0.0}{0.0} & \tr{0.07}{0.00} & \tr{0.02}{0.01} & \tr{0.0}{0.0} & \tr{0.07}{0.00} & \tr{0.11}{0.03} & \tr{0.0}{0.0} & \tr{0.54}{0.00} & \tr{0.01}{0.01} & \trb{100.0}{0.0} & \trb{0.00}{0.00} & \tr{0.01}{0.00} \\
\textsc{ILP} {\scriptsize (600s)} & \tr{50.0}{0.0} & \tr{0.05}{0.00} & \tr{328.9}{0.00} & \tr{20.0}{0.0} & \tr{0.56}{0.00} & \tr{548.8}{0.00} & \trb{100.0}{0.0} & \trb{0.00}{0.00} & \tr{170.8}{0.00} & \trb{100.0}{0.0} & \trb{0.00}{0.00} & \tr{0.05}{0.00} \\
\textsc{ACE} & \trb{100.0}{0.0} & \trb{0.00}{0.00} & \tr{0.71}{0.01} & \trb{100.0}{0.0} & \trb{0.00}{0.00} & \tr{65.98}{1.60} & \tr{66.7}{0.0} & -- & \tr{206.5}{0.04} & \trb{100.0}{0.0} & \trb{0.00}{0.00} & \tr{0.45}{0.00} \\
\midrule
\textsc{PI-SAGE} & \tr{0.0}{0.0} & \tr{0.66}{0.03} & \tr{7.09}{0.22} & \tr{0.0}{0.0} & \tr{0.58}{0.01} & \tr{14.07}{0.86} & \tr{0.0}{0.0} & \tr{1.79}{0.12} & \tr{5.65}{0.37} & \tr{53.1}{0.0} & \tr{4.30}{0.02} & \tr{3.89}{0.01} \\
\textsc{ROS} & \tr{0.0}{0.0} & \tr{0.52}{0.04} & \tr{0.66}{0.05} & \tr{0.0}{0.0} & \tr{0.53}{0.02} & \tr{0.82}{0.06} & \tr{0.0}{0.0} & \tr{1.36}{0.14} & \tr{0.69}{0.05} & \tr{53.1}{0.0} & \tr{3.96}{0.04} & \tr{0.78}{0.00} \\
\textsc{AnyCSP} & \trb{100.0}{0.0} & \trb{0.00}{0.00} & \tr{0.94}{0.04} & \tr{70.0}{14.8} & \tr{0.01}{0.00} & \tr{25.91}{1.45} & \tr{96.6}{10.0} & \trb{0.00}{0.01} & \tr{3.96}{2.41} & \trb{100.0}{0.0} & \trb{0.00}{0.00} & \tr{0.19}{0.01} \\
\textsc{ConsFormer} & \tr{0.0}{0.0} & \tr{2.06}{0.06} & \tr{24.83}{0.10} & \tr{0.0}{0.0} & \tr{1.93}{0.02} & \tr{24.77}{0.14} & \tr{11.7}{19.6} & \tr{1.32}{0.19} & \tr{23.17}{2.93} & \tr{61.5}{3.2} & \tr{0.32}{0.03} & \tr{2.12}{0.01} \\ 
\textsc{ReLCol} & \tr{0.0}{0.0} & \tr{0.42}{0.02} & \tr{31.22}{0.23} & \tr{0.0}{0.0} & \tr{0.35}{0.00} & \tr{719.2}{0.2} & \tr{0.0}{0.0} & \tr{0.51}{0.06} & \tr{30.24}{0.08} & \tr{53.1}{0.0} & \tr{6.75}{0.10} & \tr{8.14}{0.04} \\
\midrule
\textsc{Ours} & \trb{100.0}{0.0} & \trb{0.00}{0.00} & \tr{0.76}{0.00} & \trb{100.0}{0.0} & \trb{0.00}{0.00} & \tr{4.88}{0.82} & \trb{100.0}{0.0} & \trb{0.00}{0.00} & \tr{0.54}{0.14} & \trb{100.0}{0.0} & \trb{0.00}{0.00} & \tr{0.04}{0.00} \\
\bottomrule
\end{tabular}%
}
}
\caption{Solution quality on isolated instances: Solved\% ($\uparrow$), Mono\% ($\downarrow$) and inference time ($\downarrow$).}
\label{tab:toronto_queens_cycles}
\vspace{-2ex}
\end{table}

We additionally evaluate on \texttt{book} graphs connecting characters, Erdos-Renyi \texttt{DSJC} graphs, and random \texttt{le450} graphs from the DIMACS collection. These experiments are deferred to Appendix~\ref{sec:additional_exp}, and they confirm the strong solution quality observed in Table~\ref{tab:toronto_queens_cycles}. In particular, \textsc{Scone} solves all book graphs and \texttt{le450\_15} instances, including cases rarely solved by baselines.

\subsection{Ablation study}

\begin{figure}[t]
    \centering
    \includegraphics[width=\linewidth]{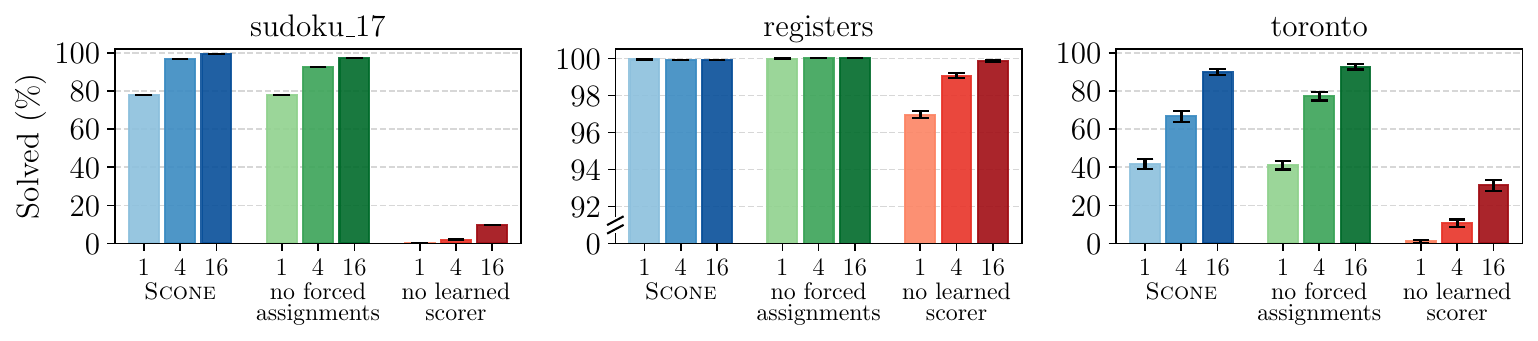}
    \vspace{-3.9ex}
    \caption{Solved rates at decoding widths $W\in\{1,4,16\}$ for \textsc{Scone}, for a version without forced assignments, and for a version with the neural scorer replaced by a minimum-conflict heuristic.}
    \label{fig:nolearned}
    \vspace{-2.5ex}
\end{figure}

First, we study the contributions of learned color selection and forced assignments by ablating each at decoding widths $W\in\{1,4,16\}$. In particular, we compare \textsc{Scone} to a variant with no forced assignments and, most importantly, to a variant that replaces the learned color scorer with a simple heuristic rule that assigns to the selected node the color with minimum number of conflicts. As shown in Figure~\ref{fig:nolearned}, 
removing the learned scorer lowers solved rates across all benchmarks, with particularly large drops on Sudoku and Toronto that persist not only on greedy decoding, but also when increasing the decoding width.
Even on registers, removing the learned scorer reduces the greedy decoding solve rate below the one of \textsc{Dsatur}. Removing forced assignments has no effect on solution quality for greedy decoding  or ensembling, and slightly helps to prune beams at $W > 1$.

\begin{wrapfigure}{r}{0.562\textwidth}
    \vspace{-2.3ex}
    \centering
    \includegraphics[width=0.99\linewidth]{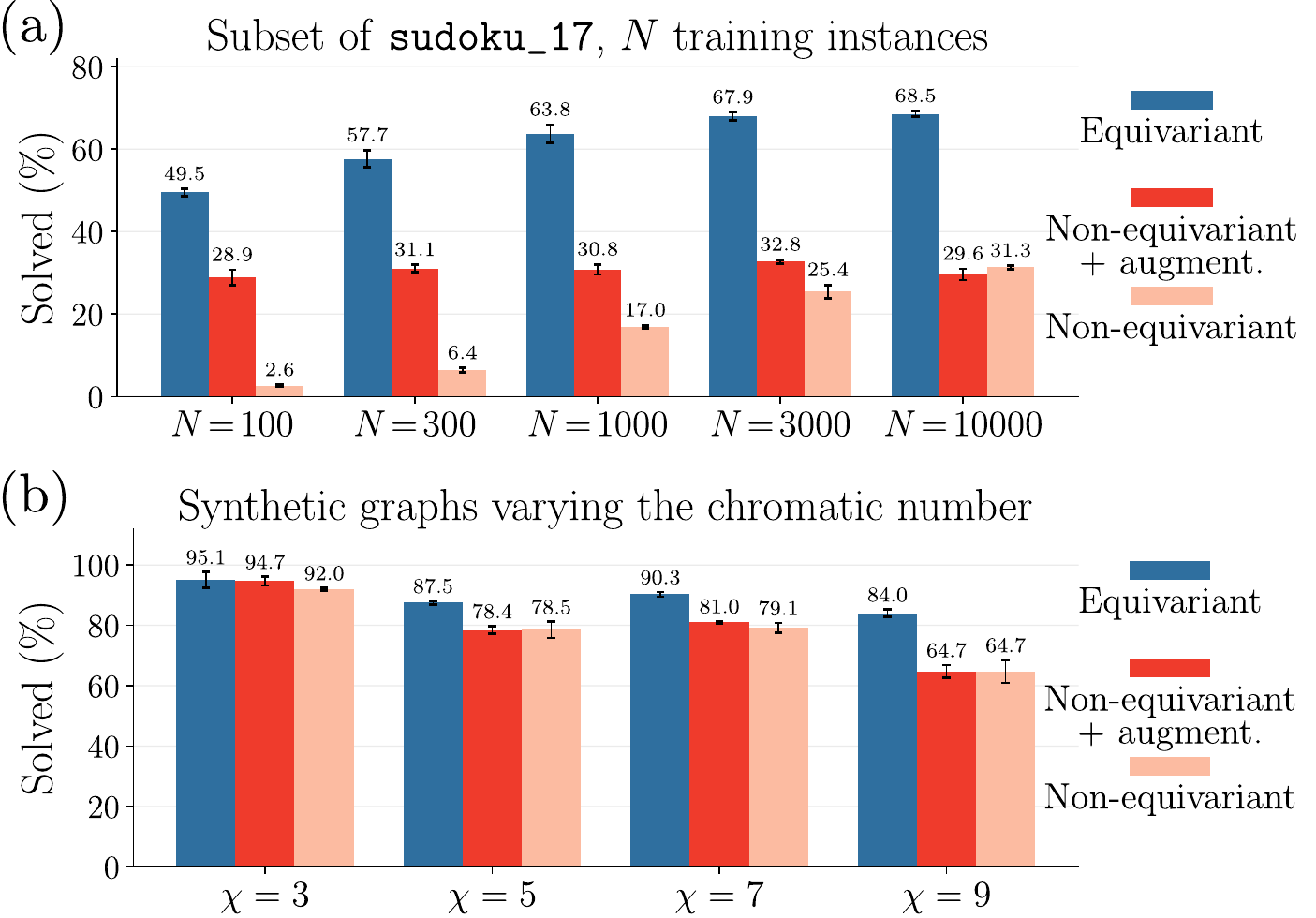}\vspace{-1.6ex}
    \caption{Ablation on Sudoku and synthetic graph\rlap{s.}}
    \label{fig:ablation}
    \vspace{-2ex}
\end{wrapfigure}

Moreover, we study the effect of equivariance by comparing \textsc{Scone} and a variant without color equivariance, with and without data augmentation by color permutations. First, we train on increasing subsets of \texttt{sudoku\_17}, see Figure~\ref{fig:ablation}~(a). 
With only $N=100$ training instances, \textsc{Scone} solves $49\%$ of the test set, compared with just $2\%$ for the non-equivariant model. Data augmentation narrows the gap in the low-data regime, but even at the largest subset it remains substantial, highlighting the greater sample efficiency of architectural equivariance.
Second, we generate datasets {targeting} increasing chromatic numbers $\chi$, setting the palette size $k=\chi$. While the non-equivariant variants are competitive for small palette sizes $k$, their performance degrades as $k$ increases, see Figure~\ref{fig:ablation}~(b). Data augmentation provides no improvement, and the gap to \textsc{Scone} widens with $k$, indicating that color equivariance becomes increasingly important as %
$k$ grows, as suggested by Theorem~\ref{thm:full-reduction}.

\section{Conclusions}

We introduced \textsc{Scone}, a graph-coloring method combining color
equivariance with sequential decoding, as supported by our theory. Across
real-world and synthetic benchmarks, it achieves near-perfect solution quality with low inference times.
As future work, these positive results motivate extending \textsc{Scone} to handle richer constraints and objectives beyond standard graph coloring, such as room capacities in timetabling and move preferences in register allocation, making it applicable to a broader
range of practical coloring-based applications in operations research. Moreover, it is interesting to study which symmetries can be exploited in other combinatorial problems.

\newpage

\subsection*{Acknowledgements}
The project is sponsored by the Defense Advanced Research Projects Agency under cooperative agreement HR0011262E027. The content of the information does not necessarily reflect the position or the policy of the Government, no official endorsement should be inferred.
PP and MW further acknowledge support from the Aramont Fellowship Fund for Emerging Science Research. MW was partially funded by NSF award DMS-2406905, an Alfred P. Sloan Research Fellowship in Mathematics, and the AI2050 program at Schmidt Sciences (Grant G-25-69786).

\subsection*{AI use statement}

In this work, we used generative AI tools to: assist in the writing of proofs, provide feedback on research methodology or experiments, propose or refine hypotheses, implement methods.

We have not used generative AI tools to: generate synthetic data sets, help develop theoretical models or conceptual frameworks, formulate mathematical claims, provide critical ingredients for proving mathematical claims, design the research methodology or experiments, assist with translation, clean and reformat dataset, support qualitative and thematic data analysis, interpret results, create or modify scientific figures or images, draft parts of a research paper. 

Additionally, we used generative AI tools to: create or edit software code, edit the paper to improve readability, identify relevant literature, propose a title or keywords for a research paper. 

We have reviewed all AI-assisted work. We have implemented the core parts of the model ourselves, and verified all the code for correctness. 
We take responsibility for the final content of this work,
including text, claims or artifacts produced with the aid of generative AI.

\subsection*{Ethics statement}

This paper presents work whose goal is to advance the field
of Machine Learning. There are potential and indirect societal
consequences of our work, none of which we feel must be
specifically highlighted here.

\subsection*{Reproducibility statement}

For all our theoretical results, clear explanations of any assumptions and a complete
proof of the claims are included in Appendix~\ref{sec:proofs}.
Descriptions and references for datasets and baselines are reported in Sections~\ref{sec:datasets} and \ref{sec:baselines} respectively. A description of our model's implementation and of the hyperparameters we used is reported in Section~\ref{sec:setup}. Section~\ref{sec:infrastructure} describes our computing infrastructure to ensure reproducible inference times. The code, checkpoints and datasets to reproduce our results will be made publicly available on GitHub.

\bibliography{biblio}
\bibliographystyle{iclr2027_conference}

\newpage

\appendix

\section*{\Large Appendix}

\section{Limitations and Future Work}

While \textsc{Scone} achieves state-of-the-art performance, there are some limitations, which suggest promising directions for future work.
For example, \textsc{Scone} constructs a coloring sequentially and therefore requires up to
$O(n)$ forward passes for a graph with $n$ nodes, which limits the applicability to larger graphs. 
A promising direction is to color multiple
compatible nodes speculatively, retaining their assignments when they are
consistent and revising them otherwise.
Moreover, \textsc{Scone} is an approximate solver. It always returns a complete
coloring but cannot guarantee that it is proper or certify that no proper
coloring exists for the given palette. Combining the model with backtracking
or an exact solver could provide such guarantees while retaining its ability
to identify promising assignments.

\section{Additional related work} \label{sec:related}

\textbf{Classical graph coloring.}$\ $
The $k$-colorability problem is NP-complete for every fixed $k\geq 3$ \citep{garey1974some}, while it can be solved in linear time for $k=2$. Exact algorithms can however be faster than enumerating all $k^n$ assignments. For example, $3$-colorability can be decided in $O(1.33^n)$ time using reductions to constraint satisfaction \citep{beigel20053}.
Moreover, several graph classes admit polynomial-time algorithms. Chordal graphs can be optimally colored in linear time \citep{tarjan1984simple}. Graphs of bounded treewidth can also be colored efficiently by dynamic programming over a tree decomposition \citep{arnborg1989linear}.
Brooks' theorem shows that every connected graph of maximum degree $\Delta$ can be colored with $\Delta$ colors unless it is a complete graph or an odd cycle \citep{brooks1941colouring}.
The four-color theorem guarantees that every planar graph admits a coloring with at most four colors \citep{appel1977every}, and the later proof of \citet{robertson1997four} also yields a quadratic-time coloring algorithm.

\textbf{Neural combinatorial optimization on graphs.}$\ $
Other works have addressed the graph coloring problem with machine learning. %
\citet{huang2019coloring} target very large graphs, and combine Monte Carlo tree search with a model that predicts a distribution over valid colors for the current node.
\citet{lemos2019graph} train a GNN to predict whether a graph is $k$-colorable, representing both nodes and candidate colors within the message-passing graph, and colorings are recovered indirectly by clustering the learned node embeddings. Finally, \citep{li2022rethinking} also identify node symmetry as a limitation of simultaneous GNN-based coloring and address it through randomly initialized node features. Their model is also color-equivariant, but uses a substantially more restricted class of linear equivariant updates. In contrast, \textsc{Scone} employs expressive nonlinear equivariant layers and breaks node symmetries sequentially, as previous color assignments become part of the state.

Neural combinatorial optimization has mostly focused on routing problems \citep{vinyals2015pointer, bello2016neural, drakulic2023bq}, but it has also been applied to several graph problems closely related to coloring, such as Max-Cut \citep{karalias2026geometric}. 
\citet{khalil2017learning} learn sequential construction policies for Max-Cut and minimum vertex cover using reinforcement learning, and \citet{barrett2020exploratory} learn a local-search policy that may revise previous decisions at inference time. Sequential policies have been used also for TSP \citep{khalil2017learning, drakulic2023bq}, graph edit distance \citep{pellizzoni2026gelato} and for knapsack \citep{afshar2020state}. See also \citet{cappart2023combinatorial} for an overview of the field.

\textbf{Symmetry in combinatorial reasoning.}$\ $
Symmetry has long been exploited in constraint programming and SAT to reduce redundant search.
For example, Lex-leader constraints select one representative from each orbit \citep{crawford1996symmetry}, while specialized methods handle interchangeable values \citep{puget2006efficient,walsh2012symmetry}.
Symmetry-aware architectures have also been used for neural combinatorial optimization, notably in \textsc{NeuroSAT} \citep{selsam2018learning}, which respects the natural permutations of SAT formulas by encoding them as graphs and using a message-passing mechanism.
Methods for more general constraint-satisfaction problems based on message-passing paradigms like \textsc{RUN-CSP} \citep{toenshoff2021graph} and \textsc{AnyCSP} \citep{tonshoff2023anycsp} also inherit equivariance to the natural permutations of the constraint graphs. 
Finally, \citet{kim2022sym} introduce a regularization-based training scheme that exploits problem and solution symmetries in neural routing problems, including TSP and CVRP.

\section{Proofs} \label{sec:proofs}

\subsection{Section \ref{sec:symmetries}}

\begin{theorem}[Expressivity of sequential coloring]
\label{thm:sequential-universality}
Fix $n,k\in\mathbb{N}$, and let
$s_\theta(G,c,u)\in\mathbb{R}^k$ denote the color scores assigned to an
uncolored node $u$. Assume that the predictor is universal over rooted isomorphism
types of size at most $n$.
Consider any sequential decoder that repeatedly selects an arbitrary
uncolored node $u$, chooses
$i\in\operatorname*{arg\,max}_{j\in\mathcal C} [s_\theta(G,c,u)]_j$,
and assigns color $i$ to $u$, breaking ties arbitrarily. Then there
exists a single parameter setting such that, for every extendable
partial $k$-coloring $(G,c_0)$ with $|V_G|\leq n$, every decoding
trajectory yields a proper coloring extending $c_0$.
\end{theorem}
\begin{proof}
For an uncolored node $u$, let $c^{u\leftarrow i}$ denote the partial
coloring obtained by assigning color $i$ to $u$. Define the target
score function as $s(G,c,u)_i =1 $ if $(G,c^{u\leftarrow i})$ is extendable 
and $0$ otherwise.

This function depends only on the rooted partial-coloring isomorphism
type of $(G,c,u)$. In fact, it is also color-equivariant.
There are finitely many rooted partial-coloring isomorphism types with
at most $n$ nodes. By the assumed universality, a single
parameter setting $\theta$ realizes this score function on all of
them.

Now consider any extendable state $(G,c)$ and any uncolored node $u$.
Let $c^\star$ be a proper completion of $c$. Assigning
$i=c^\star(u)$ to $u$ preserves extendability, so the maximum score at
$u$ is $1$. Consequently, every maximizing color preserves
extendability.

The selected node and all score ties may therefore be resolved
arbitrarily. Every decoding step preserves extendability and colors
one additional node. By induction, decoding reaches a complete
extendable state, which is a proper coloring extending $c_0$.
\end{proof}

\newtheorem*{lem:fixed-reduction}{Lemma~\ref{lem:fixed-reduction}}
\begin{lem:fixed-reduction}
    Consider a state $(G, c) \in \bigcup_{\tau \in S_k} \mathcal{S}_{\tau \cdot (G, c_0)}$ with exactly $r$ colors appearing in $c$. Let $R(G, c) = |S_k\cdot(G,c)|$ denote the size of the orbit of $(G,c)$ under color permutations. Then, 
    \[
    \frac{k!}{(k-r)! \, \min(|{\rm Aut}(G)|, r!)} \leq R(G, c) \leq \frac{k!}{(k-r)!}.
    \] 
\end{lem:fixed-reduction}
\begin{proof}
    Let $C_1,\dots,C_r$ be the used colors, and let $\Pi_c\leq S_r$ be the group of permutations of the used colors that can be realized by automorphisms of $G$, that is, 
    \[ 
    \Pi_c = \left\{ \tau\in S_r : \exists\,\varphi\in{\rm Aut}(G) \text{ s.t. } c(v)=C_i \Rightarrow c(\varphi(v))=C_{\tau(i)} \ \ \forall v\in V_G,\ \forall i=1,\dots,r \right\}. 
    \]
    A permutation $\tau\in S_k$ stabilizes the state $(G,c)$ if the partial coloring obtained by relabeling the colors according to $\tau$ is isomorphic to $(G,c)$. 
    Hence, on the $r$ used colors, $\tau$ must induce a permutation belonging to $\Pi_c$, while the remaining $k-r$ unused colors may be permuted arbitrarily. Therefore, $\left|{\rm Stab}_{S_k}(G,c)\right| = |\Pi_c|(k-r)!$. 
    By the orbit-stabilizer theorem \citep[Chapter 3]{rotman2012introduction}, the orbit $S_k\cdot(G,c)$ of $(G,c)$ has cardinality 
    \[ 
    R(G, c) = \frac{|S_k|}{|{\rm Stab}_{S_k}(G,c)|} = \frac{k!}{(k-r)!\,|\Pi_c|}. 
    \] 
    Since $1 \leq |\Pi_c| \leq \min(|{\rm Aut}(G)|, r!)$, we have the claim.
\end{proof}

In the main paper, we claim that, for Sudoku puzzles, a careful application of the lemma yields a reduction in the state space size by a factor of $R(G,c)= \frac{9!}{|\Pi_c|}\geq \frac{9!}{648} = 560$. This requires first the following lemma. 

\begin{lemma}[Automorphisms of a uniquely completable Sudoku]
\label{lem:sudoku-automorphisms}
Let $G, c$ be a $9\times 9$ Sudoku puzzle with unique completion
$c^\star$, and let $\Pi_c$ be as defined in
Lemma~\ref{lem:fixed-reduction}. Then
\[
    |\Pi_c|
    \le
    |\Pi_{c^\star}|
    \le 648.
\]
\end{lemma}
\begin{proof}
First, $c$ must contain $8$ or $9$ distinct digits. Otherwise, two
digits would be absent from $c$, and swapping those digits in
$c^\star$ would produce a second completion of $c$. Therefore, if $r$
is the number of digits occurring in $c$, then $r\in\{8,9\}$.

Let $\tau\in\Pi_c$. By the definition of $\Pi_c$, there exists
$\phi\in\operatorname{Aut}(G)$ such that
$c(\phi(v))=\tau(c(v))$ for every colored vertex $v$. 
Note that even if $r=8$, the permutation $\tau$ has a unique extension in $S_9$
by fixing the remaining digit.

Define a completed coloring $c'$ by
$c'(\phi(v))= \tau(c^\star(v))$. The automorphism $\phi$
preserves the Sudoku constraint graph, and $\tau$ only
relabels the digits, so $c'$ is a valid completed Sudoku grid.

Let $V_c$ be the set of colored nodes and note that $\phi(V_c) = V_c$.
Now let $x\in V_c$. Since $\phi(V_c)=V_c$, there is some $v\in V_c$
such that $x=\phi(v)$. Because $c^\star$ extends $c$, we have
\[
    c'(x)
    = {\tau}(c^\star(v))
    = \tau(c(v))
    = c(\phi(v))
    = c(x).
\]
Hence $c'$ extends $c$. Since $c$ has the unique completion
$c^\star$, it follows that $c'=c^\star$. Therefore,
$c^\star(\phi(v)) = {\tau}(c^\star(v))$
for every $v\in V_G$. Hence
${\tau}\in\Pi_{c^\star}$ and
$|\Pi_c|\leq|\Pi_{c^\star}|$.

Let $\Gamma_{c^\star}$ denote the full automorphism group of the
completed grid, whose elements are the pairs
$(\phi,\sigma)\in\operatorname{Aut}(G)\times S_9$ satisfying
$c^\star(\phi(v))=\sigma(c^\star(v))$ for every $v$. The projection
of $\Gamma_{c^\star}$ onto its digit-permutation component has image
$\Pi_{c^\star}$. Therefore,
$|\Pi_{c^\star}|\leq|\Gamma_{c^\star}|$.
Finally, \citet{mcguire2014there} argue that
$|\Gamma_{c^\star}|\leq 648$, which proves the result.
\end{proof}

\newtheorem*{thm:full-reduction}{Theorem~\ref{thm:full-reduction}}
\begin{thm:full-reduction}[Exponential reduction in the state space]
    Consider a finite $k$-colorability instance set $\mathcal{I}$ with no starting colors. Let $|{\rm Aut}(G)| \leq B \ \forall G \in \mathcal{I}$. Let $\mathcal{S_I}$ be its associated state space and $\mathcal{S}^+_{\mathcal I}$ the space of extendable states. Then:
    \begin{enumerate}
        \item If all graphs $G \in \mathcal{I}$ have more than $k$ nodes, then $|\mathcal{S_I} / S_k | \leq B \, \smash{\dfrac{16}{e^k}} \, |\mathcal{S_I}|  \,$.
        \item If all graphs $G \in \mathcal{I}$ satisfy $0 \leq k - \chi(G) \leq \kappa$, then $|\mathcal{S}^+_{\mathcal I} / S_k | \leq B  \, \smash{\dfrac{ \kappa! \, 2^{k - \kappa } }{ k! }} \, |\mathcal{S}^+_{\mathcal I}| \, $.
    \end{enumerate}
\end{thm:full-reduction}
\begin{proof}
    Let $\mathcal{\tilde S}_G$ be the state space before quotienting by graph isomorphisms, i.e. where each node has an unique identifier. 
    We then have that $\mathcal{S}_G = \mathcal{\tilde S}_G / {\rm Aut}(G)$ by definition, and since ${\rm Aut}(G)$ commutes with the color permutations $S_k$, we have that $\mathcal{S}_G / S_k = \mathcal{\tilde S}_G / ({\rm Aut}(G) \times S_k)$. 
    Moreover, $|\mathcal{\tilde S}_G| \leq |{\rm Aut}(G)| \, |\mathcal{\tilde S}_G / {\rm Aut}(G)|$ as every ${\rm Aut}(G)$-orbit can only have size at most $|{\rm Aut}(G)|$, and $\mathcal{\tilde S}_G / ({\rm Aut}(G) \times S_k) \leq \mathcal{\tilde S}_G / S_k$ as quotienting by ${\rm Aut}(G)$ can only decrease the number of elements. 
    Therefore, we have that 
    \[
    \frac{|\mathcal{S}_G|}{|\mathcal{S}_G / S_k|} = 
    \frac{|\mathcal{\tilde S}_G / {\rm Aut}(G)|}{|\mathcal{\tilde S}_G / ({\rm Aut}(G) \times S_k)|} \geq 
    \frac{|\mathcal{\tilde S}_G|} {|\mathcal{\tilde S}_G / S_k| \, |{\rm Aut}(G)|} \geq 
    \frac{1}{B} \frac{|\mathcal{\tilde S}_G|} {|\mathcal{\tilde S}_G / S_k|}.
    \]
    The same derivation follows for the extendable versions, yielding $\frac{|\mathcal{S}^+_G|}{|\mathcal{S}^+_G / S_k|} \geq 
    \frac{1}{B} \frac{|\mathcal{\tilde S}^+_G|} {|\mathcal{\tilde S}^+_G / S_k|}.$ Since the $1/B$ factor does not depend on $G$, we can get the same factor uniformly over $\mathcal{I}$. 
    Therefore, we will work with the states with identified nodes, which ensures that coloring different nodes leads to different states. 

    We first prove the reduction for $\mathcal{\tilde S_I}$. Let's consider an graph $G$ with $n$ nodes. In the resulting state space $\mathcal{\tilde S}_{G}$, each of those nodes can be either uncolored or have one of the $k$ available colors, leading to a total of $(k+1)^n$ possible states. The quotient space $\mathcal{\tilde S}_{G} / S_k$ ignores the identities of the colors. Therefore, to count the number of states using $r$ colors, we count the number of partitions of $n$ elements (the nodes) into $r$ equivalent non-empty classes (the colors) and one distinct and possibly empty class (the uncolored nodes). To count these, we use the following construction. We add a dummy node, and we say that whichever class this node is assigned to, is the class of uncolored nodes. We can then count the number of partitions of $n+1$ elements into $r+1$ equivalent and non-empty classes,
    which is the Stirling number of second kind $n+1 \brace r+1$ \citep{graham1994concrete}. We then have that $|\mathcal{\tilde S}_{G} / S_k| = \sum_{r=0}^k {n+1 \brace r+1}$. Therefore, we obtain the exact formula $R := |\mathcal{\tilde S}_{G}| \, / \, |\mathcal{\tilde S}_{G} / S_k| = \, \frac{(k+1)^n}{\sum_{r=0}^k {n+1 \brace r+1}}$. 

    Then, a standard bound yields that ${n+1 \brace r+1} \leq \frac{(r+1)^n}{r!}$, which then implies a reduction of factor 
    { \footnotesize \[
    R \geq \left( \sum_{r=0}^k \frac{1}{r!} \bigg(\frac{r+1}{k+1}\bigg)^n \right)^{\!\!-1}\!\! \geq \left( \sum_{r=0}^k \frac{1}{r!} \bigg(1-\frac{k-r}{k+1}\bigg)^{k+1} \right)^{\!\!-1}\!\!  \geq \left( \sum_{r=0}^k \frac{e^{r-k}}{r!} \right)^{\!\!-1}\!\! = e^{k} \left( \sum_{r=0}^k \frac{e^{r}}{r!} \right)^{\!\!-1}\!\! \geq \frac{e^k}{e^e}.
    \]}
    Taking the bound over all $G \in \mathcal{I}$ yields the claim.

    We then prove the reduction for $\mathcal{\tilde S}^+_{\mathcal I}$. 
    Let's consider a graph $G$ with $\chi(G) = q$. 
    Let $E_r$ be the number of extendable quotient states (i.e., states in $\mathcal{\tilde S}^+_{G} / S_k$) that use exactly $r$ colors. 
    If $r < q$, then at least $q-r$ colors absent from the current state occur on uncolored vertices. Choosing one vertex carrying each such color produces at least $q-r$ distinct $(r+1)$-color successor states. On the other hand, each $r+1$-color state has an $r$-color predecessor of this type only by uncoloring the unique vertex of a singleton color class. Hence it has at most $r+1$ predecessors. This implies the inequality $(q-r)E_r \leq (r+1)E_{r+1}$.  
    Importantly, if we didn't work on $\mathcal{\tilde S}$ and didn't have the unique node identifiers, several such successors may collapse to the same quotient state under graph automorphisms, so the lower bound $(q-r)$ on the number of distinct successors could fail.
    We therefore have 
    \[
    \frac{E_r}{\binom{q}{r}} \leq \frac{E_{r+1}}{\binom{q}{r+1}} \quad \forall \, 0 \leq r < q \, ,
    \]
    i.e., $a_r := {E_r}\big/{q \choose r}$ is nondecreasing in $r$. 

    We claim that, for every $R\le q$,
    \[
    \frac{\sum_{r=0}^R E_r}{\sum_{r=0}^k E_r}
    \leq
    \frac{\sum_{r=0}^R E_r}{\sum_{r=0}^q E_r}
    \leq
    2^{-q}\sum_{r=0}^R\binom{q}{r}.
    \]
    To prove it, set
    $A:=\sum_{r=0}^R a_r\binom{q}{r}$,
    $B:=\sum_{r=R+1}^q a_r\binom{q}{r}$,
    $C:=\sum_{r=0}^R \binom{q}{r}$, and
    $D:=\sum_{r=R+1}^q \binom{q}{r}$.
    Then it suffices to show
    \[
    \frac{A}{A+B}\leq \frac{C}{C+D},
    \]
    or equivalently,
    $AD\leq BC$.
    
    Since $a_r\leq a_R$ for every $r\leq R$, we have
    \[
    A = \sum_{r=0}^R a_r\binom{q}{r} \leq
    a_R\sum_{r=0}^R \binom{q}{r} = a_R C.
    \]
    Similarly, since $a_r\geq a_R$ for every $r>R$,
    \[
    B = \sum_{r=R+1}^q a_r\binom{q}{r}
    \geq a_R\sum_{r=R+1}^q \binom{q}{r} = a_R D.
    \]
    Multiplying the first inequality by $D$ and the second by $C$ we obtain
    $ AD\leq a_R CD\leq BC$. 
    We then have that
    \[
    \frac{\sum_{r=0}^R a_r\binom{q}{r}}
    {\sum_{r=0}^q a_r\binom{q}{r}}
    \leq
    \frac{\sum_{r=0}^R \binom{q}{r}}
    {\sum_{r=0}^q \binom{q}{r}}.
    \]
    Finally, using
    $\sum_{r=0}^q \binom{q}{r}=2^q$, we obtain
    \[
    \frac{\sum_{r=0}^R E_r}{\sum_{r=0}^k E_r}
    \leq
    \frac{\sum_{r=0}^R E_r}{\sum_{r=0}^q E_r}
    \leq 2^{-q}\sum_{r=0}^R \binom{q}{r}.
    \]
    Therefore the distribution with density $f(r) = \frac{ E_r}{\sum_{j=0}^k E_j}$ stochastically dominates ${\rm Bin}(q, 1/2)$, which stochastically dominates ${\rm Bin}(k - \kappa, 1/2)$, for $\kappa\in\{0,\dots,k\}$. Since $0 \leq k - \chi(G) \leq \kappa \leq k$, this is well defined. 
    Since $\frac{k!}{(k-r)!}$ is nondecreasing in $r$, we then have that 
    {\small
    \[
    \frac{|\mathcal{\tilde S}^+_{G}|}{|\mathcal{\tilde S}^+_{G} / S_k|} =
    \frac{\sum_{r=0}^k \!E_r \frac{k!}{(k-r)!}\!}{\sum_{r=0}^k E_r}
    =
    \sum_{r=0}^k \frac{ E_r}{\sum_{j=0}^k \!E_j\!} \frac{k!}{(k\!-\!r)!}
    \geq 
    \sum_{r=0}^q \frac{1}{2^q} \binom{q}{r} \frac{k!}{(k\!-\!r)!}
    \geq \sum_{r=0}^{k-\kappa} \frac{1}{2^{k-\kappa}} \binom{k\!-\!\kappa}{r} \frac{k!}{(k\!-\!r)!} .
    \]
    }
    If we take only the $r = k-\kappa$ term, we obtain a $\frac{k!}{2^{k-\kappa} \kappa!}$ bound. 
    Taking the bound over all instances then yields the claim.
    
\end{proof}

\subsection{Section \ref{sec:model}}

We prove that the color predictor of \textsc{Scone} is equivariant to
node and color permutations. Importantly, this guarantee applies only
to the predicted color scores. We do not require tie-breaking among
colors or the complete decoding procedure to be equivariant.

\newtheorem*{thm:equi}{Theorem~\ref{thm:equi}}
\begin{thm:equi} 
    The color predictor of \textsc{Scone} is equivariant to the joint action of $S_{|V_G|}\!\times\! S_k$.  
    Let $(G,c)$ be any graph $k$-coloring instance, and let $p_u(G,c)\in\mathbb{R}^k$ be the output of the predictor for node $u$. Then, for every parameter setting, every $\pi\in S_{|V_G|}$, every $\tau\in S_k$, and every $u\in V_G$, \vspace{-0.5ex}
    \[
        p_{\pi(u)}\bigl(\pi\cdot G,\, \tau\cdot(\pi\cdot c)\bigr)
        =
        P_\tau\,p_u(G,c).
    \] 
\end{thm:equi}
\begin{proof}
    Let $(G',c')=(\pi\cdot G,\tau\cdot(\pi\cdot c))$. We show by induction over the message-passing layers that
    \[
        H^\ell_{\pi(u)}(G',c')=P_\tau H^\ell_u(G,c)
        \qquad \forall u\in V_G.
    \]
    For $\ell=0$, if $c(u)=\varepsilon$, then $c'(\pi(u))=\varepsilon$, and both sides contain $e_{\rm unk}$ in every color slot. Otherwise, $c'(\pi(u))=\tau(c(u))$. Hence the slot containing $e_{\rm on}$ is permuted from $c(u)$ to $\tau(c(u))$, while every other slot contains $e_{\rm off}$. Therefore, we have that $H^0_{\pi(u)}(G',c')=P_\tau H^0_u(G,c)$.

    Suppose that the claim holds at layer $\ell-1$. Since node relabeling preserves neighborhoods,
    \[
        \mathcal N_{G'}(\pi(u))
        =
        \{\pi(v):v\in\mathcal N_G(u)\}.
    \]
    Using the induction hypothesis and the equivariance of the element-wise aggregator, we obtain
    \[
    \begin{aligned}
        \hat H^\ell_{\pi(u)}(G',c')
        &=
        \bigoplus_{w\in\mathcal N_{G'}(\pi(u))}
        H^{\ell-1}_w(G',c') \\
        &=
        \bigoplus_{v\in\mathcal N_G(u)}
        P_\tau H^{\ell-1}_v(G,c) \\
        &=
        P_\tau \hat H^\ell_u(G,c).
    \end{aligned}
    \]

    We next verify that every linear map $L$ used in ${\rm MLP}_{S_k}$ is color-equivariant. For every slot $i$,
    \[
    \begin{aligned}
        L(P_\tau H)[i]
        &=
        W_{\rm self}H[\tau^{-1}(i)]
        +
        W_{\rm all}\sum_{j=1}^k H[\tau^{-1}(j)]
        +b \\
        &=
        W_{\rm self}H[\tau^{-1}(i)]
        +
        W_{\rm all}\sum_{j=1}^k H[j]
        +b \\
        &=
        [P_\tau L(H)]_i.
    \end{aligned}
    \]
    Point-wise nonlinearities also commute with permutations of the color slots. Therefore, their composition ${\rm MLP}_{S_k}$ is $S_k$-equivariant. Moreover, concatenation along the feature axis satisfies
    \[
        P_\tau(A\|B)=(P_\tau A)\|(P_\tau B).
    \]
    LayerNorm is applied independently to each color slot with shared parameters, so it also commutes with $P_\tau$. The same holds for the residual connection. Consequently,
    \[
        H^\ell_{\pi(u)}(G',c')
        =
        P_\tau H^\ell_u(G,c),
    \]
    completing the induction.

    Finally, the readout ${\rm MLP}_{S_k}$ is color-equivariant by the same argument. If a softmax is applied to obtain probabilities, it also commutes with color permutations. We therefore have
    \[
        p_{\pi(u)}\bigl(\pi\cdot G,\tau\cdot(\pi\cdot c)\bigr)
        =
        P_\tau p_u(G,c)
    \]
    for every node and parameter setting, proving equivariance to the joint action of
    $S_{|V_G|}\times S_k$.
\end{proof}

\newtheorem*{thm:dsatur}{Theorem~\ref{thm:dsatur}}
\begin{thm:dsatur} 
    Let $k, \Delta \in \mathbb{N}$. There exists a parameter setting of \textsc{Scone} s.t. every $k$-colorable cycle,
chordal or bipartite graph with degree at most $\Delta$ can be properly colored from the empty coloring.
\end{thm:dsatur}
\begin{proof}
Let $c$ be the current proper partial coloring, and let $\varepsilon$ denote
the label of an uncolored node. For every uncolored node $v$, define
\[
F_c(v)
=
\{c(u):u\in\mathcal N_G(v),\ c(u)\neq\varepsilon\},
\]
\[
T_c(v)
=
\{c(w):\exists u,\ \{v,u\},\{u,w\}\in E_G,\ 
c(u),c(w)\neq\varepsilon\}\setminus F_c(v).
\]
Thus, $F_c(v)$ contains the colors appearing on colored neighbors of $v$,
while $T_c(v)$ contains the available colors observed at distance two through
a colored neighbor.

The most-constrained-first decoder of \textsc{Scone} selects an 
uncolored node maximizing $|F_c(v)|$. 

The color predictor realizing the claim gives
maximum score to the colors in $T_c(v)$ when this set is nonempty, and to the
colors in $[k]\setminus F_c(v)$ otherwise. 
We first show that this is realizable by \textsc{Scone}. 
The set $F_c(v)$ is determined by the one-hop neighborhood of $v$. In the
first message-passing layer, each colored node can additionally record the
colors appearing in its neighborhood. Aggregating these representations in a
second layer determines $T_c(v)$ and whether it is empty. Since $k$ and
$\Delta$ are fixed, only finitely many relevant aggregate inputs can occur.
With sufficient width, the color-equivariant MLPs can assign distinct codes
to these inputs. The codes can be chosen to remain distinct after LayerNorm,
while the residual term can be incorporated into the finite lookup performed
by the following layer. Any additional layers can implement the identity by
setting their residual update to zero. Hence the rule can be realized by a
single parameter setting of the full architecture \textsc{Scone}. We then show
that it succeeds on each graph family.

First, let $G$ be bipartite. We treat each connected component separately, and
isolated nodes are immediate. 
Let's then consider a component with bipartition $(A,B)$, and assume without loss of generality that its first
selected node lies in $A$ and receives color $a$.
The connectivity of the component guarantees that while there are uncolored nodes, 
at least one has $|F_c(v)| \geq 1$. Moreover, all nodes other components have $|F_c(v)| = 0$,
so the decoder will color the current component before moving to another one. 
The first selected node in $B$ receives some color $b\neq a$.

We show inductively that 
every colored node in $A$ has color $a$ and every
colored node in $B$ has color $b$. The base case is given by the first two selected nodes.
Let's then suppose that the inductive hypothesis holds.
Suppose that the decoder selects the next node to be
$v \in A$. Then it must have a colored neighbor $u \in B$, so $F_c(v) = \{b \}$. 
Moreover, $u$ has a previously colored neighbor in $A$.
Thus $v$ has a colored node of
$A$ at distance two. Since every node at distance two from $v$ lies in $A$,
we have $T_c(v)=\{a\}$, and the decoder assigns color $a$ to $v$. The argument
for $v\in B$ is symmetric. Hence, inductively, the component is colored properly using only
$a$ and $b$.

Every even cycle is bipartite, so the preceding argument also applies to it.
Now consider an odd cycle. After the first two adjacent nodes receive
distinct colors $a$ and $b$, the same two-hop rule extends an alternating
coloring along the cycle. Until only one node remains uncolored, the colored
nodes form a path whose colors alternate between $a$ and $b$. The path
between the two neighbors of the final node has an odd number of edges, so
these neighbors have different colors. Thus both $a$ and $b$ belong to
$F_c(v)$ for the final node and $T_c(v)=\emptyset$. Since an odd cycle is
$k$-colorable only if $k\geq3$, some color in
$[k]\setminus\{a,b\}$ remains available and completes a proper coloring.

Finally, let $G$ be chordal. Let $x_1,\ldots,x_n$ denote the order in which
the decoder colors the nodes. We show inductively that every prefix
$x_1,\ldots,x_t$ is a valid prefix of maximum cardinality search (MCS), as introduced in \citet{tarjan1984simple}. The first
choice is valid because all nodes initially have no colored neighbors.

Suppose that $x_1,\ldots,x_{t-1}$ is an MCS prefix and that the current
partial coloring is proper. Continue this prefix arbitrarily according to
MCS, and number the resulting ordering by $\alpha(x_j)=n-j+1$. On a chordal
graph, every ordering produced by MCS is a perfect elimination ordering
\citep{tarjan1984simple}. With this numbering convention, the higher-numbered
neighbors of every node form a clique. Therefore, the already-colored
neighbors of every uncolored node form a clique.

Since the partial coloring is proper, these neighbors have pairwise distinct
colors. Therefore, for every uncolored node $v$, its saturation
$|F_c(v)|$ equals its number of already-colored neighbors. Maximizing
saturation is therefore exactly the MCS selection rule. The node selected by
the decoder is a valid next MCS choice, while its confidence-based secondary
criterion only breaks ties between valid choices.

The already-colored neighbors of the selected node form a clique of size at
most $\omega(G)-1$. Chordal graphs are perfect, so
$\omega(G)=\chi(G)\leq k$. Hence at most $k-1$ colors are forbidden. The set
$[k]\setminus F_c(v)$ is therefore nonempty, and every color preferred by the
predictor is available. Assigning one of these colors preserves properness
and completes the induction.

Thus, the same parameter setting properly colors every $k$-colorable
bipartite, chordal, or cycle graph of maximum degree at most $\Delta$,
independently of its number of nodes.
\end{proof}

\subsection{Proof of Theorem \ref{thm:express}} \label{sec:proofthmexpress}

Theorem~\ref{thm:sequential-universality} shows that sequential
coloring succeeds whenever the predictor can realize arbitrary scoring
rules on the relevant rooted isomorphism types. We
now show that the sum-aggregating version of \textsc{Scone} has this
property on cover-separated families of amenable graphs, and results easily extend to any injective aggregation $\oplus$. Throughout
this section, the depth and width of the network may depend on $n$ and
$k$.

Recall that the $1$-dimensional Weisfeiler-Leman algorithm
($1$-WL) iteratively refines node labels according to the labels of
neighboring nodes \citep{arvind2017graph, xu2018powerful, morris2019weisfeiler}. Given a partially colored graph $(G,c)$, we
initialize
\[
{\rm WL}^0(G,c,v)=c(v)
\]
for every $v\in V_G$, where $\varepsilon$ is treated as an additional
label. For $\ell\geq 1$, we define
\[
{\rm WL}^{\ell}(G,c,v)
=
{\rm HASH}\!\left(
{\rm WL}^{\ell-1}(G,c,v),
\lmset
{\rm WL}^{\ell-1}(G,c,w)
:
w\in\mathcal N_G(v)
\rmset
\right),
\]
where ${\rm HASH}$ is injective and shared across graphs. We denote
the resulting multiset of labels by
\[
{\rm WL}^{\ell}(G,c)
=
\lmset
{\rm WL}^{\ell}(G,c,v)
:
v\in V_G
\rmset.
\]
Two partially colored graphs are distinguished by $1$-WL if these
multisets differ at some iteration. We say that $(G,c)$ is
\emph{identified} by $1$-WL if it is distinguished from every
nonisomorphic partially colored graph.

Since the partition induced by the $1$-WL labels is refined at every
iteration, it eventually stabilizes. We refer to the resulting label
of a node as its \emph{stable $1$-WL label}. Two rooted partially
colored graphs $(G,c,u)$ and $(G',c',u')$ have the same
\emph{stable rooted $1$-WL type} if $u$ and $u'$ receive the same
stable $1$-WL label when color refinement is run with a common
labeling scheme.
An uncolored graph $G$ is \emph{amenable} if $(G,c_\varepsilon)$ is
identified by $1$-WL, where $c_\varepsilon$ denotes the empty
coloring.

We first assess that the color-equivariant MLPs used by
\textsc{Scone} can realize arbitrary equivariant functions on finite
sets.

\begin{lemma}
\label{lem:equivariant-lookup}
Let $\mathcal{X}\subseteq\mathbb{R}^{k\times d}$ be finite and closed
under row permutations, and let
$F:\mathcal{X}\to\mathbb{R}^{k\times p}$ be $S_k$-equivariant. Then
$F$ can be realized on $\mathcal{X}$ by a finite-width ReLU network
whose linear layers have the form
$
    L(H)[i]
    =
    W_{\mathrm{self}}H[i]
    +
    W_{\mathrm{all}}\sum_{j=1}^k H[j]
    +
    b.
$
\end{lemma}
\begin{proof}
We first use the fact that a ReLU MLP can realize any function on a
finite set. Indeed, for distinct vectors $a_1,\ldots,a_m$, choose $w$
such that the scalars $w^\top a_1,\ldots,w^\top a_m$ are distinct.
A piecewise-linear function can interpolate arbitrary prescribed
values at these points, and every such function is representable by a
one-hidden-layer ReLU network.

Let $\mathcal{A}=\{a_1,\ldots,a_m\}$ be the set of all rows occurring
in matrices in $\mathcal{X}$. Using the preceding fact, construct a
row-wise ReLU MLP $\psi$ such that $\psi(a_\ell)=e_\ell$. For
$H\in\mathcal{X}$ and $i\in[k]$, define
\[
    z(H,i)
    =
    \psi(H[i])
    \bigg\|
    \sum_{j=1}^k\psi(H[j]).
\]
The second component records the multiplicity of every row type in
$H$.

If $z(H,i)=z(H',i')$, then $H[i]=H'[i']$ and $H,H'$ have the same
multiset of rows. Hence there exists $\tau\in S_k$ such that
$H'=P_\tau H$ and $i'=\tau(i)$. Equivariance gives
$F(H')[i']=F(H)[i]$. Therefore, the assignment
$\rho(z(H,i))=F(H)[i]$ is well defined on the finite set of pointed
signatures. Another row-wise ReLU MLP realizes $\rho$ exactly.
Then, we have that 
\[
    N(H)[i]
    =
    \rho\Bigg(
        \psi(H[i])
        \bigg\|
        \sum_{j=1}^k\psi(H[j])
    \Bigg)
\]
satisfies $N(H)=F(H)$ on $\mathcal{X}$. Both $\psi$ and $\rho$ are
implemented with $W_{\mathrm{all}}=0$, while the intermediate
concatenation is one linear layer of the required form. Thus $N$ has
the stated $S_k$-equivariant architecture.
\end{proof}

\begin{lemma}
\label{lem:name-s-wl}
For fixed $n$ and $k$, there exist a depth, a width, and a parameter setting of \textsc{Scone} such that the score predictor can realize any $S_k$-equivariant scoring rule on the finite set of stable rooted $1$-WL types occurring in partially colored graphs with at most $n$ nodes.
\end{lemma}
\begin{proof}
For fixed $n$ and $k$, only finitely many $1$-WL labels and
neighborhood multisets can occur. The initialization of
\textsc{Scone} can be chosen to give distinct, $S_k$-equivariant
representations to the initial node labels.

Suppose that the embeddings after layer $\ell-1$ distinguish all
$(\ell-1)$-round $1$-WL labels. Since every degree is at most $n-1$,
the embeddings can be chosen so that their sums uniquely determine all
multisets of at most $n-1$ neighboring labels. There are only finitely
many such multisets. Lemma~\ref{lem:equivariant-lookup} therefore
allows the equivariant update MLP to map each pair consisting of a
node embedding and its neighborhood aggregate to a distinct
equivariant code for the corresponding $\ell$-round $1$-WL label.

LayerNorm and the residual connections do not affect this finite-set
construction. With sufficiently large width, choose the update codes
orbit by orbit so that every color slot has zero mean and the same
nonzero variance. The codes can be chosen generically so that adding
the residual embeddings and summing the resulting node embeddings
introduces no unintended collisions among the finitely many inputs
under consideration. The update MLP realizes these normalized codes
by Lemma~\ref{lem:equivariant-lookup}. Shared per-slot LayerNorm
preserves them up to its learnable affine transformation, which can
absorb the normalization constant, and the residual term produces
the chosen next-layer embeddings.

Induction therefore shows that sufficiently many message-passing
layers distinguish all stable rooted $1$-WL types occurring on graphs
with at most $n$ nodes. Finally, an equivariant readout of sufficient
width realizes any prescribed $S_k$-equivariant scoring rule on this
finite set of types.
\end{proof}

We next record the closure of identification under refinements of the
initial coloring. Throughout the remainder of the proof, partially
colored graphs are regarded as vertex-colored graphs, with
$\varepsilon$ as an additional color, and all automorphisms and
fractional automorphisms are required to preserve these colors.
The definitions and implications from amenability to compactness and
the Godsil property extend to vertex-colored graphs
\citep{arvind2017graph}. In particular, every partially colored graph
identified by $1$-WL is compact and Godsil.

\begin{lemma}
\label{lem:amenable-refinement}
Let $(G,c)$ be a partially colored graph identified by $1$-WL, and let $c'$ be a refinement of $c$. Then $(G,c')$ is also identified by $1$-WL.
\end{lemma}
\begin{proof}
Suppose that $(G,c')$ and $(G',d')$ are not distinguished by $1$-WL. Since $c'$ refines $c$, there exists a map $\rho$ such that $c=\rho\circ c'$. Applying the same coarsening to $d'$, let $d=\rho\circ d'$. The partially colored graphs $(G,c)$ and $(G',d)$ are also indistinguishable by $1$-WL. Since $(G,c)$ is identified by $1$-WL, there exists an isomorphism $\varphi$ between them.

We use $\varphi$ to transport $d'$ to a refinement $\widetilde d'$ of $c$ on $G$, defined by $\widetilde d'(u)=d'(\varphi(u))$.
The $1$-WL equivalence of $(G,c')$ and $(G,\widetilde d')$ gives a fractional isomorphism $X$ \citep{arvind2017graph}. Since both colorings refine $c$, the matrix $X$ is also a fractional automorphism of $(G,c)$. Moreover, $(G,c)$ is compact because it is identified by $1$-WL. We may therefore write
$X=\sum_t \alpha_t P_{\pi_t}$,
where $\alpha_t\geq 0$, $\sum_t\alpha_t=1$, and each $\pi_t$ is an automorphism of $(G,c)$.

Since $X$ is a fractional isomorphism between the two refinements, $X_{uv}=0$ whenever $c'(u)\neq \widetilde d'(v)$.
All coefficients in the decomposition of $X$ are nonnegative. Therefore, every $\pi_t$ with $\alpha_t>0$ satisfies $c'(u)=\widetilde d'(\pi_t(u))$ for every $u\in V_G$.
Such a $\pi_t$ is an isomorphism between $(G,c')$ and $(G,\widetilde d')$. Composing it with $\varphi$ gives an isomorphism between $(G,c')$ and $(G',d')$. Hence $(G,c')$ is identified by $1$-WL.
\end{proof}

We now use the assumption that distinct underlying graphs in the
family have distinct universal covers \citep{krebs2015universal}. Importantly, this assumption
concerns the uncolored graphs rather than the initial instances or
their subsequent partial colorings.

\begin{lemma}[Rooted types in a cover-separated family]
\label{lem:cover-separated-rooted-wl}
Let $\mathcal K$ be a collection of connected amenable graphs such
that no two nonisomorphic graphs in $\mathcal K$ have isomorphic
universal covers. Let $G,G'\in\mathcal K$, let $c,c'$ be arbitrary
partial colorings of $G,G'$, and let $u\in V_G$ and $u'\in V_{G'}$.
If $u$ and $u'$ have the same stable $1$-WL label, then
$(G,c,u)$ and $(G',c',u')$ are isomorphic as rooted partially colored
graphs.
\end{lemma}
\begin{proof}
The colorings $c$ and $c'$ refine the empty colorings of the amenable
graphs $G$ and $G'$. Hence
Lemma~\ref{lem:amenable-refinement} shows that $(G,c)$ and $(G',c')$
are identified by $1$-WL.

Equality of the stable labels of $u$ and $u'$ implies that their
rooted color-labeled universal covers are isomorphic
\citep{krebs2015universal}. Forgetting the color labels gives
isomorphic universal covers of the underlying graphs $G$ and $G'$.
The assumption on $\mathcal K$ therefore implies that $G$ and $G'$
are isomorphic. In particular,
$|V_G|=|V_{G'}|$.

The isomorphic rooted universal covers induce corresponding stable
cells $V_1,\ldots,V_m$ and $V'_1,\ldots,V'_m$ with the same
equitable-quotient parameters. Let $d_{pq}$ denote the number of
neighbors in $V_q$ of a node in $V_p$. We have
\[
|V_p|d_{pq}=|V_q|d_{qp}
\qquad\text{and}\qquad
|V'_p|d_{pq}=|V'_q|d_{qp}.
\]
Since the graphs are connected, their equitable quotients are
connected. These equations therefore determine the cell sizes up to
a common positive factor. Hence there exists $\lambda>0$ such that
$|V'_p|=\lambda|V_p|$ for every $p$. Summing over the cells and using
$|V_G|=|V_{G'}|$ gives $\lambda=1$.

Thus $(G,c)$ and $(G',c')$ have the same stable $1$-WL label
multiset and are not distinguished by $1$-WL. Since they are
identified by $1$-WL, there is a color-preserving isomorphism
$\varphi$ from $(G,c)$ to $(G',c')$.

The nodes $\varphi(u)$ and $u'$ have the same stable $1$-WL label in
$(G',c')$. Moreover, every graph identified by $1$-WL is Godsil
\citep{arvind2017graph}, so each stable cell is an orbit of a subgroup
of color-preserving automorphisms. There is therefore an automorphism
$\alpha$ of $(G',c')$ such that
$\alpha(\varphi(u))=u'$. The composition
$\alpha\circ\varphi$ is an isomorphism from $(G,c)$ to $(G',c')$
that maps $u$ to $u'$.
\end{proof}

We can now prove the main expressivity result.

\newtheorem*{thm:express}{Theorem~\ref{thm:express}}
\begin{thm:express}
Let $n,k\in\mathbb N$, and let $\mathcal G$ be a cover-amenable family of graphs with at most $n$ nodes. There exists a parameter setting of \textsc{Scone} s.t. every $k$-colorable $(G,c_0)$ with $G\in\mathcal G$ can be solved.
\end{thm:express}
\begin{proof}
For an uncolored node $u$, let $K_u$ denote its connected component,
let $c_u$ denote the restriction of $c$ to $K_u$, and let
$c_u^{u\leftarrow i}$ be obtained by assigning color $i$ to $u$.
Define the target score function by
\[
s(G,c,u)_i
=
\begin{cases}
1,
&
\text{if $(K_u,c_u^{u\leftarrow i})$ is extendable},
\\
0,
&
\text{otherwise}.
\end{cases}
\]

By Lemma~\ref{lem:cover-separated-rooted-wl}, the stable rooted
$1$-WL type of $u$ determines the rooted partially colored component
$(K_u,c_u,u)$ up to isomorphism. Consequently, the target score
function is well defined on stable rooted $1$-WL types. It is also
$S_k$-equivariant, since permuting the color labels maps proper
completions to proper completions with the same permutation applied.
Lemma~\ref{lem:name-s-wl} therefore gives a single parameter setting
of \textsc{Scone} that realizes this scoring rule.

Consider an extendable state $(G,c)$ and any uncolored node $u$. A
proper completion $c^\star$ of $c$ assigns some color
$i=c^\star(u)$ to $u$. Its restriction to $K_u$ shows that
$(K_u,c_u^{u\leftarrow i})$ is extendable. Hence the maximum score at
$u$ is $1$, and every maximizing color preserves the extendability of
the component containing $u$. The other components are unchanged, so
the complete state also remains extendable.

The decoder may therefore select any uncolored node and break every
color tie arbitrarily. Each step preserves extendability and colors
one additional node. By induction, decoding reaches a complete
extendable state, which is a proper coloring extending $c_0$.
\end{proof}

A sufficient condition for Theorem~\ref{thm:express} is that every
connected component be $1$-WL-discrete \citep{arvind2017graph}. 
Such components are amenable,
and two of them with isomorphic universal covers are necessarily
isomorphic. This condition holds with high probability in the dense
Erdős--Rényi model \citep{babai1980random}. Moreover, the proof gives
the stronger conclusion that every decoding trajectory solves the
instance, independently of node selection and tie-breaking.

\section{Experimental setup}

In this section, we discuss additional details about the experimental setup, including dataset preparation, details on baselines, on our implementation of \textsc{Scone}, and on the computing infrastructure.

\subsection{Datasets} \label{sec:datasets}

\paragraph{Sudoku instances.}
We evaluate on five collections of $9\times9$ Sudoku puzzles. The \texttt{sudoku\_satnet} and \texttt{sudoku\_rrn} datasets are taken from SATNet~\citep{wang2019satnet} and Recurrent Relational Networks~\citep{palm2018recurrent}, respectively. We additionally use \texttt{sudoku\_17}, a collection of non-equivalent puzzles with only 17 clues,\footnote{\url{https://web.archive.org/web/20131019184812if_/http://school.maths.uwa.edu.au/~gordon/sudokumin.php}} and \texttt{sudoku\_top1465}, a collection of 1,465 particularly difficult puzzles.\footnote{\url{http://magictour.free.fr/sudoku.htm}} 
Finally, we use \texttt{sudoku\_extreme}, using the splits introduced by
\citet{davis2026lattice} (1000 training instances and 4000 test instances), which is the most challenging  public Sudoku benchmark.
We independently divide \texttt{sudoku\_17} and \texttt{sudoku\_satnet} into $70\%/15\%/15\%$ training, validation, and test splits. In contrast, \texttt{sudoku\_rrn} and \texttt{sudoku\_top1465} are used only for testing. The optimal solutions used for training are either provided in the original data, or are obtained with an exact ILP solver. 

\paragraph{Register-allocation instances.}
We curate this benchmark from Appel and George's collection of 27,922 register-interference graphs \citep{george1996iterated} generated by Standard ML of New Jersey while compiling itself.\footnote{\url{https://www.cs.princeton.edu/~appel/graphdata/}} We parse the data into undirected graphs, remove self-loops, and remap vertex identifiers to a contiguous range. The original format specifies $K=21$ available registers and assumes an implicit clique (which we do not materialize) of $K$ precolored register nodes. We also discard the move edges, which encode preferences for assigning two nodes the same register rather than interference constraints. We retain graphs with $|V_G| \geq 4$ and filter out duplicated graphs. Finally, we compute a greedy clique lower bound and retain instances whose bound lies between 3 and 21, filtering graphs that are essentially trivially colourable or graphs whose clique lower bound already exceeds the original $K=21$ available registers budget. The resulting 8999 instances contain no initially assigned colors and are all $21$-colorable (or better). The optimal solutions used for training are obtained with an exact ILP solver. 

\paragraph{Toronto examination timetabling.}
We use version~I of the Carter--Laporte--Lee Toronto benchmark \footnote{\url{https://people.cs.nott.ac.uk/pszrq/data.htm}}, see \citet{carter1996examination}. Each file records the exams and their enrollments, and the exams taken by each student. We construct an undirected conflict graph in which vertices represent exams and two exams are adjacent if some student takes both. Then, the exams taken by each student induce a clique. We adopt the uncapacitated minimum-timeslot formulation, ignoring room capacities and proximity costs, so a proper coloring represents a feasible timetable. Duplicate entries in each student's exam list are removed before constructing edges, while isolated exams and the original exam numbering are retained. We set $k$ to the best-known feasible number of colors for each instance as reported in \citet[Table 11]{qu2009survey}, and these values are upper bounds on the chromatic number rather than certified optima. To reduce the size of the graphs without affecting the validity of the instance, we apply the standard $k$-core reduction, iteratively removing vertices of degree less than $k$, as these can be colored trivially. 
We partition the thirteen Toronto instances according to their original size before preprocessing. The four smallest instances, \texttt{hec-s-92}, \texttt{sta-f-83}, \texttt{yor-f-83}, and \texttt{ute-s-92}, form the validation set and are used only to select the decoding configuration. The remaining instances form the test set, which we divide into the smaller instances \texttt{ear-f-83}, \texttt{tre-s-92}, \texttt{lse-f-91}, and \texttt{kfu-s-93}, and the large instances \texttt{rye-s-93}, \texttt{car-f-92}, \texttt{uta-s-92}, \texttt{car-s-91}, and \texttt{pur-s-93}. No Toronto instance is used for training, as we apply the same synthetically trained model throughout.

\paragraph{Queen graphs.} For each $p$, \texttt{queens\_$p$} is the queen graph of a $p\times p$ chessboard, with one node per square and edges connecting squares in the same row, column, or diagonal \citep{vasquez2004new}. A proper coloring therefore partitions the board into sets of mutually non-attacking queens. We use \texttt{queens\_7} exclusively for validation and reserve \texttt{queens\_8}, \texttt{queens\_9}, and \texttt{queens\_10} for testing. None of these instances is used for training, and we apply the same synthetically trained model throughout.

\paragraph{Structured graphs.}
Alongside the application benchmarks, we evaluate on synthetic families with chromatic numbers known exactly by construction. These instances are straightforward for classical algorithms but isolate potential failure modes of learned colorers. The cycle graphs contain hundreds of nodes, with equal numbers of even and odd cycles requiring two and three colors, respectively, and thus test long-range parity reasoning. The grid graphs are 4-regular bipartite lattices with hundreds of nodes. The complete graphs are $K_n$ for $32\leq n\leq64$ and admit no slack, since every vertex must receive a distinct color. 

\paragraph{Synthetic training data.}
In Section~\ref{sec:synth_train}, the results for \textsc{Scone} are obtained using a single generalist checkpoint trained exclusively on synthetic graphs, and no benchmark instance is used for training. 
We generate graphs with planted proper colorings by partitioning the nodes into color classes and adding edges only between classes. The dataset spans diverse graph sizes, color budgets, densities, and degree distributions, combining random, preferential-attachment, and clique-augmented constructions.
In particular, it contains $10000$ graphs, split into
$8000/1000/1000$ training, validation, and test instances. We
draw $n$ log-uniformly from $[20,200]$ and $k$ uniformly from
$[3,\min(30,n)]$. Graph families are sampled as uniform-edge,
preferential-attachment, and clique-augmented constructions with equal
probabilities. We retain connected candidates with an
edge between every pair of planted color classes for which the
generator's deterministic \textsc{DSatur} requires at least $k+1$
colors.
This provides valid supervision while preventing the training set from being dominated by instances that a standard constructive heuristic solves immediately. 
Importantly, the planted proper coloring is generated jointly with each graph,
providing ground-truth supervision without requiring an additional
graph-coloring solve, which would be computationally expensive.

\subsection{Baselines} \label{sec:baselines}

\textbf{Greedy, \textsc{Dsatur} and \textsc{RLF}.}$\ $ We adapt  all three heuristics to the fixed-budget setting considered in this work. Greedy \citep{welsh1967upper} processes nodes in a fixed order, while \textsc{Dsatur} \citep{brelaz1979new} repeatedly selects an uncolored node with the largest number of distinct colors among its colored neighbors, breaking ties by residual degree. \textsc{RLF} \citep{leighton1979graph} constructs one color class at a time by greedily building a maximal independent set. In their classical formulations, all three methods assign the first available color and introduce a new color whenever all existing colors are blocked; they therefore always return a proper coloring, with the number of colors used serving as their output. In our setting, by contrast, the palette size $k$ is fixed. We restrict all three methods to these $k$ colors and, whenever every color already appears in the selected node's neighborhood, assign a color that conflicts with the fewest colored neighbors. Thus, the methods always return a complete coloring, which may be improper and can be evaluated using the same solved-instance and monochromatic-edge metrics as the learning-based methods. On feasible bipartite, chordal, and complete graphs, where \textsc{Dsatur} is exact, the palette cap does not bind and this modification leaves its classical behavior unchanged.

\textbf{\textsc{ILP}.}$\ $
We use a standard integer linear programming (ILP) formulation with binary
variables assigning one of the $k$ colors to each node and auxiliary
variables indicating monochromatic edges. The objective minimizes the
total number of monochromatic edges, with initial colors enforced as
fixed assignments. The ILP uses color-symmetry-breaking constraints to ensure an efficient representation. 
We solve the resulting problem using Gurobi \citep{gurobi}.

\textbf{\textsc{ACE}.}$\ $ ACE \citep{lecoutre2023ace} is a frontier general-purpose constraint solver for finite-domain constraint-satisfaction problems (CSP). We represent each node by an integer variable with domain \(\{1,\ldots,k\}\), restrict precolored nodes to singleton domains, and impose a binary inequality constraint for every edge. We run ACE with its default search configuration and the same dataset-specific time limits as the ILP baseline. Since ACE searches directly for a feasible coloring, it either returns a proper coloring or terminates without a solution within the time limit. Therefore, either its Mono\% is $0$ or is not defined.
To provide ACE with the most favorable comparison, we report only its internal
solve time, excluding JVM startup and instance-loading overhead, which would add an additional $0.5s$ per instance.

\textbf{\textsc{PI-SAGE} and \textsc{ROS}.}$\ $ Both methods optimize a separate GNN from scratch on each test instance using a relaxed coloring objective, and then round its predictions to obtain a discrete coloring. Thus, neither method uses a pretrained checkpoint or requires a training split. For \textsc{PI-SAGE} \citep{schuetz2022graph}, we use two mean-aggregating \textsc{SAGEConv} layers with hidden dimension $32$ and minimize the soft Potts objective $\sum_{{u,v}\in E}\langle p_u,p_v\rangle$. We optimize with learning rate $10^{-4}$ for at most $100{,}000$ epochs, with tolerance $10^{-4}$ and early-stopping patience $1{,}000$, and retain the best argmax coloring encountered during optimization. For \textsc{ROS} \citep{qiu2025ros}, we use the authors' two-layer \textsc{GraphConv} architecture with \textsc{GraphNorm} and hidden dimension $100$, and optimize the QUBO objective $\operatorname{tr}(XWX^\top)$ with learning rate $10^{-2}$ for at most $10{,}000$ epochs, using tolerance $10^{-2}$ and patience $100$. We obtain discrete solutions by drawing $100$ multinomial samples and retaining the best coloring. All hyperparameters follow the authors' published defaults. Our implementations faithfully port \textsc{PI-SAGE} from DGL to PyG and reproduce the original \textsc{ROS} GCN.

\textbf{\textsc{AnyCSP}.}$\ $
\textsc{AnyCSP} \citep{tonshoff2023anycsp} is a pretrained anytime solver for constraint-satisfaction problems. We train its graph-coloring model using the authors' provided code and retain the default \texttt{num\_boost}=8, which replicates each instance within a batch to perform eight parallel search restarts. Since the original checkpoint was not released, we compare our retrained model with the runtimes reported by the authors as a sanity check. They report $5$ to $17$ seconds for queen graphs of sizes $8$ to $10$, while our model solves the same sizes in $0.4$ to $11$ seconds. Although runtimes across different hardware are not directly comparable, these results suggest that our checkpoint is comparable to, and potentially stronger than, the original. Training takes more than 24 hours on an H200 GPU.
We run \textsc{AnyCSP} with a time limit of 5 seconds on Sudokus to limit the runtime, and with a time limit of 30s per instance otherwise, albeit sometimes it exceeds this timelimit. Note that this limit is only reached on some large \texttt{toronto} instances and on \texttt{queens\_10\_10}, and this time limit is comfortably higher than the time used by \textsc{Scone}.
The original method does not support fixed colors. We represent each pinned node using a singleton domain, making clue violations impossible, rather than introducing unary constraints that the solver could violate and whose penalties would be mixed with edge violations in its objective. Since the released model was not trained on instances with pinned nodes, we retrain it on \texttt{sudoku\_17} and on \texttt{sudoku\_extreme} using the authors' training procedure. 

\textbf{\textsc{ConsFormer}.}$\ $ \textsc{ConsFormer} \citep{xu2025selfsupervised} is a self-supervised transformer that iteratively refines a complete assignment. Since no pretrained checkpoints are provided, we train all models ourselves. We use a candidate pool of one, corresponding to the standard model in the authors' main experiments; we do not report the optional multi-start extension, for which additional candidates trade against refinement iterations under a fixed time budget. Following the authors' ablations, we use masked relative positional encodings rather than the repository default of no relative encoding. The remaining settings follow the published configurations: seven layers, three attention heads, hidden dimension $128$, the ABSE loss, and $5000$ training epochs. For Sudoku, we use \texttt{ape\_dim}=$2$ and threshold $0.5$; for general coloring, we use \texttt{ape\_dim}=$1$ and threshold $0.7$.

We formulate Sudoku as graph coloring rather than using \textsc{ConsFormer}'s native all-different constraints. Each puzzle is represented as an $81$-node, $20$-regular constraint graph, with edges encoding pairwise inequalities and clues fixed as one-hot assignments. We retain the two-dimensional Sudoku positional encoding, so the model receives the grid geometry while using the same constraint formulation and evaluation metrics as the other coloring datasets. The native all-different formulation yields only marginally better results. We train the model on \texttt{sudoku\_satnet} as in the original paper, as training on \texttt{sudoku\_17} yields significantly worse results both in terms of training loss and of validation solution quality.

\textbf{\textsc{ReLCol}.}$\ $ \textsc{ReLCol} \citep{watkins2023generating} uses deep Q-learning and a GNN to learn a node ordering for greedy coloring, minimizing the number of colors. We adapt it to a fixed palette by retaining its first $k$ color classes and assigning each remaining node its least-conflicting color among them. %
At inference, we repeat episodes under the same nominal per-graph budgets as \textsc{AnyCSP}: 5\,s on registers, Sudoku and cycles, and 30\,s on Toronto, queens and DIMACS. We retain the best coloring and stop once a proper $k$-coloring is found. On four of the five large Toronto instances, a single episode already exceeds the budget. We therefore report one episode and its full runtime, reaching approximately 3200\,s on \texttt{pur-s-93}.

\textbf{\textsc{LDT}.}$\ $
The Lattice Deduction Transformer (\textsc{LDT}) \citep{davis2026lattice} combines a recurrent Transformer with stochastic search over a lattice of candidate assignments. It cannot be applied directly to general graph-coloring instances and is therefore included only in the Sudoku experiments. Since no checkpoint is available, we train the model using the authors' unmodified implementation. We follow the configuration from their Table~1, using 4000 training steps while retaining all other defaults. At inference, we use the default 64 parallel chains and at most 1000 search rounds. We verified that the training and test instances exactly match the split used for \textsc{Scone}. 

\textbf{\textsc{TRM}.}$\ $
The Tiny Recursive Model (\textsc{TRM}) \citep{jolicoeur2025less} is a recurrent Transformer-based refinement model. It cannot be applied directly to general graph-coloring instances and is therefore included only in the Sudoku experiments. Since no checkpoint is available, we train the official model on the same 1000 puzzles used by \textsc{Scone} and \textsc{LDT}, following the authors' \texttt{sudoku\_extreme\_1k\_aug\_1k} configuration with 50000 training steps and 1000 augmentations per puzzle. The model and optimization procedure are unchanged. At inference, the number of refinement steps is controlled by \texttt{n\_supervision}. We report 128 steps, the largest setting within the matched runtime budget and eight times the 16 steps used during training, thereby giving \textsc{TRM} a favorable test-time compute allowance.

\subsection{Setup for \textsc{Scone}} \label{sec:setup}

\textbf{Architecture and training.}$\ $
We implement \textsc{Scone} in PyTorch Geometric \citep{fey2019fast}. Following
Section~\ref{sec:model}, each node maintains a $k\times d$
representation with one slot per color, and all updates are equivariant to
permutations of these slots. Unless otherwise stated, we use $\mathcal L=10$
message-passing layers, hidden dimension $d=128$, and the degree-normalized
aggregation
$\sum_{v\in\mathcal N(u)} H_v / \sqrt{|\mathcal N(u)|}$, as it retains degree information while being stable across scales. 
See Section~\ref{sec:sweeps} for a comparison with different depths and widths.
We use LayerNorm \citep{ba2016layer} and residual connections \citep{he2016deep}, as they have been shown to make training more stable \citep{luo2025can}.

We generate training states by sampling a masking fraction uniformly for each
graph and hiding the corresponding fraction of non-fixed colors from a target
solution. Clues provided with the instance are retained. We train against
the permutation-aware targets of Section~\ref{sec:training} using AdamW \citep{kingma2014adam, loshchilov2017decoupled} with
learning rate $10^{-3}$, weight decay $10^{-5}$, batch size $64$, and a cosine
schedule with five warm-up epochs. We train for $50$ epochs on \texttt{sudoku\_17} and
\texttt{registers}, which takes less than 30 minutes, for $300$ epochs on the synthetic data, which takes roughly 6 hours, and for $2000$ epochs on \texttt{sudoku\_extreme}, given its small size, which takes roughly 2.5 hours. We select the checkpoint that maximizes the fraction of solved instances (Solved\%), and in case of ties the ones that minimizes the fraction of monochromatic edges (Mono\%) on
\textit{validation} instances.

\textbf{Inference strategies.}$\ $
For beam search, each state scores all $k$ color slots for the selected
node. After removing
children equivalent under $S_k$ color permutations, we retain the
global top $B$ states across the full candidate set. There
is no per-parent quota, so a state may contribute multiple children or
none. Beam states are ranked by the cumulative sum of assignment
log-probabilities, while forced assignments contribute no score.
Ensembling uses $E$ independent replicas, with replica $r$ initialized
from the $r$-th highest-degree uncolored node.  
The final output is the coloring with the
fewest monochromatic edges across all beams or replicas.

\textbf{Adaptive test-time scaling.}$\ $
Our decoder has two width axes: a beam size $B$ within each decoding replica
and a number $E$ of independent replicas, for total width $W=B\, E$. These axes
are not interchangeable. Additional replicas never compete with previous ones,
whereas increasing the beam size changes which partial colorings are retained
and is therefore not monotone. We use the \textit{validation split only} to determine
whether additional width should be allocated to beams, replicas, or a mixture
of both, see Section~\ref{sec:sweeps} for results. 

At test time, each graph (or batch of graphs) is first decoded at the smallest width. %
If the returned
coloring is proper, decoding stops. Otherwise, the width is doubled according
to the selected allocation rule and the graph is decoded again, up to the fixed
cap. Since correctness can be verified directly by counting monochromatic
edges, this stopping rule requires neither ground-truth colorings nor validation
labels. 
This adaptive scaling puts \textsc{Scone} in the same broad category as
\textsc{AnyCSP}, \textsc{Consformer}, \textsc{LDT}, and \textsc{TRM}, which all
employ anytime-style inference procedures that terminate either upon reaching a
search limit or upon solving the instance at hand.
We retain the best coloring found across all attempted widths because a
wider beam may discard a trajectory completed by a narrower one. The reported
decoding cost is the total time consumed across all rungs entered by each
graph, averaged over the benchmark. In the experiments, we set a maximum width of
$512$ to avoid out-of-memory errors, but this value is reached only sporadically.

\textbf{Architectures for ablations.}$\ $
All equivariance ablations use the same training configuration as the full model. The
\emph{non-equivariant} variant embeds a one-hot encoding of the current color and uses MLPs as update functions, while retaining the same message-passing depth
and aggregation. 
We evaluate every variant using greedy decoding, i.e. setting $B=1, E=1$, with
forced-assignment propagation disabled, so that additional search does not mask
differences between the models.

\subsection{Computing infrastructure} \label{sec:infrastructure}

The experiments are run on a cluster equipped with AMD EPYC 9374F CPUs and NVIDIA H200 GPUs (143 GB). The code is based on PyTorch 2.11 and PyTorch-Geometric 2.8, with CUDA 12.8.

\section{Hyperparameter studies} \label{sec:sweeps}

In this section, we discuss the sensitivity of \textsc{Scone} to its hyperparameters.
In particular, we analyze the importance of the node selection strategy,
the effect of using $E$ ensembles or $B$ beams for beam search on different datasets, as well as the
tradeoff between solution quality and inference times as the embedding dimension $d$, the number of GNN layers $\mathcal{L}$ and the decoding width $W = B \, E$.
Importantly, all the following experiments are run exclusively on the \textit{validation} splits, so that the selection of the best model configurations are not influenced by test set results.

We start by analyzing the effect of the node selection strategy. The simplest strategy is to just color the nodes in their natural order, i.e. the order in which the nodes are provided in input. By contrast, as described in Section~\ref{sec:inference}, our decoder instead follows the common most-
constrained-first (MCF) heuristic \citep{haralick1980increasing} from constraint satisfaction.

\begin{figure}[h]
    \centering
    \includegraphics[width=0.95\linewidth]{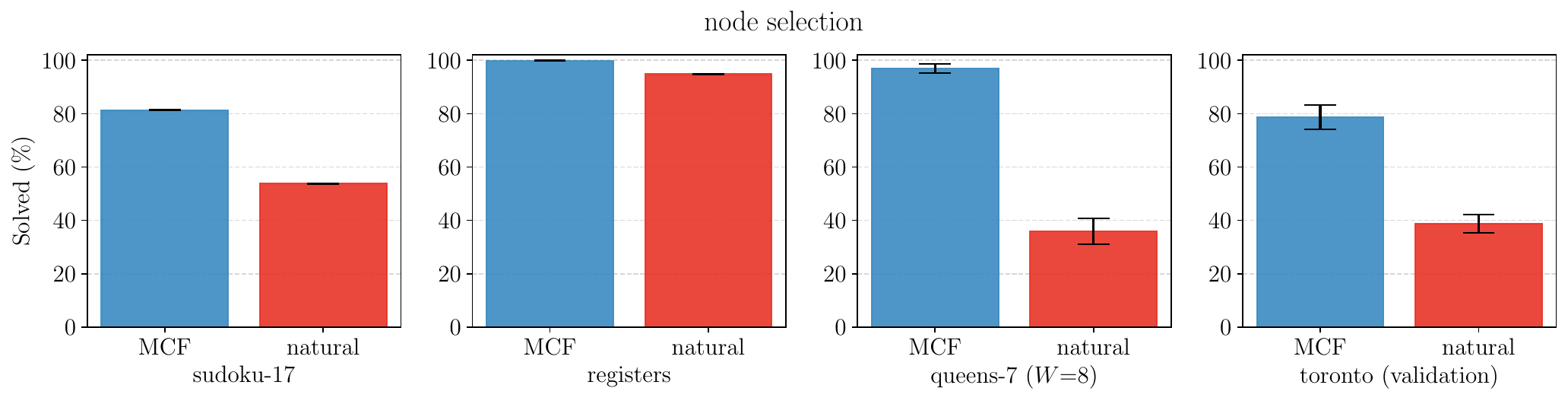} \vspace{-0.5ex}
    \caption{Validation Solved\% for coloring nodes in their natural order and in MCF order.}
    \label{fig:nodeselection}
\end{figure}

We randomly permute the input node order in every run and average over five repeats for Sudoku and registers. For \texttt{queens\_7} and Toronto, which contain fewer validation graphs, we use 20 repeats. Since at width 1 \texttt{queens\_7} cannot be solved, we use $B=8$ beams. Figure~\ref{fig:nodeselection} compares the validation Solved\% for the two strategies, showing that most-constrained-first performs consistently better than the natural order, confirming the validity of \textsc{Scone}'s design.

\begin{figure}[h]
    \centering
    \includegraphics[width=0.95\linewidth]{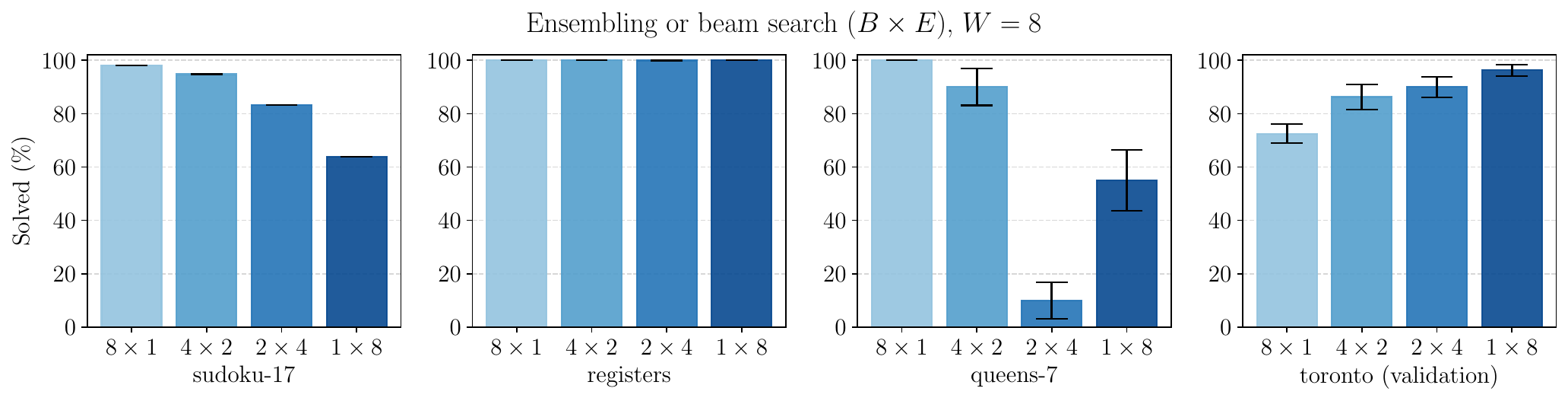} \vspace{-0.5ex}
    \caption{Validation Solved\% for different ratios between independent runs $E$ and beam width $B$.}
    \label{fig:width}
\end{figure}

We next study how a fixed decoding width is allocated between an ensemble of $E$ independent runs and beam width $B$. Figure~\ref{fig:width} shows that, at total width $W =B\, E=8$, Sudoku consistently benefits more from beam search than from ensembling, while Toronto exhibits the opposite trend. Results on queens are noisier but also favor beams, while register allocation is nearly saturated across all mixtures, even at smaller widths. Based exclusively on these validation results, we use beams for Sudoku, queens, and registers, and ensembling for Toronto throughout the paper.

Finally, we study how the embedding dimension $d$, the number of message-passing layers $\mathcal{L}$, and the decode width $W$ affect solution quality and inference time. We focus on Sudoku and Toronto, which exhibit the clearest trends in the preceding hyperparameter studies. For both datasets, we retrain the model with $d\in\{96,128,192\}$ while fixing $\mathcal{L}=10$, and with $\mathcal{L}\in\{5,10,15\}$ while fixing $d=128$. These models are decoded at width $W=1$. We then vary $W\in\{1,4,16\}$ while fixing $d=128$ and $\mathcal{L}=10$, using beams for Sudoku and ensembles for Toronto. As in the main experiments, Sudoku is trained in-distribution, while Toronto is trained on synthetic graphs. All results are reported exclusively on the validation splits.

\begin{figure}[h]
    \centering
    \includegraphics[width=0.95\linewidth]{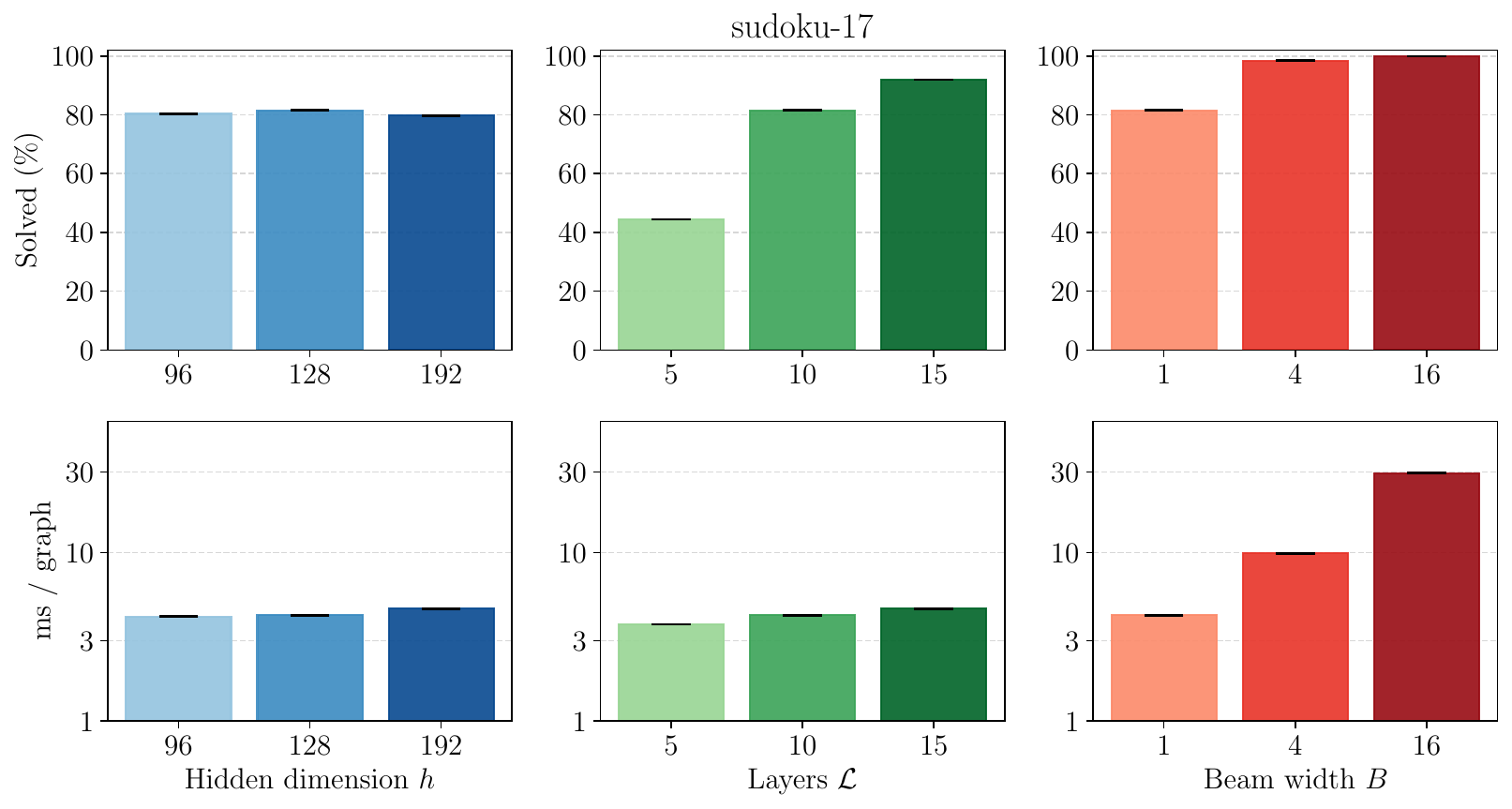} \vspace{-0.5ex}
    \caption{Validation Solved\% and average inference times on \texttt{sudoku\_17} for different values of embedding dimension $d$, message-passing layers $\mathcal{L}$ and beam width $B$.}
    \label{fig:sweep_sudoku}
\end{figure}
\begin{figure}[h]
    \centering
    \includegraphics[width=0.95\linewidth]{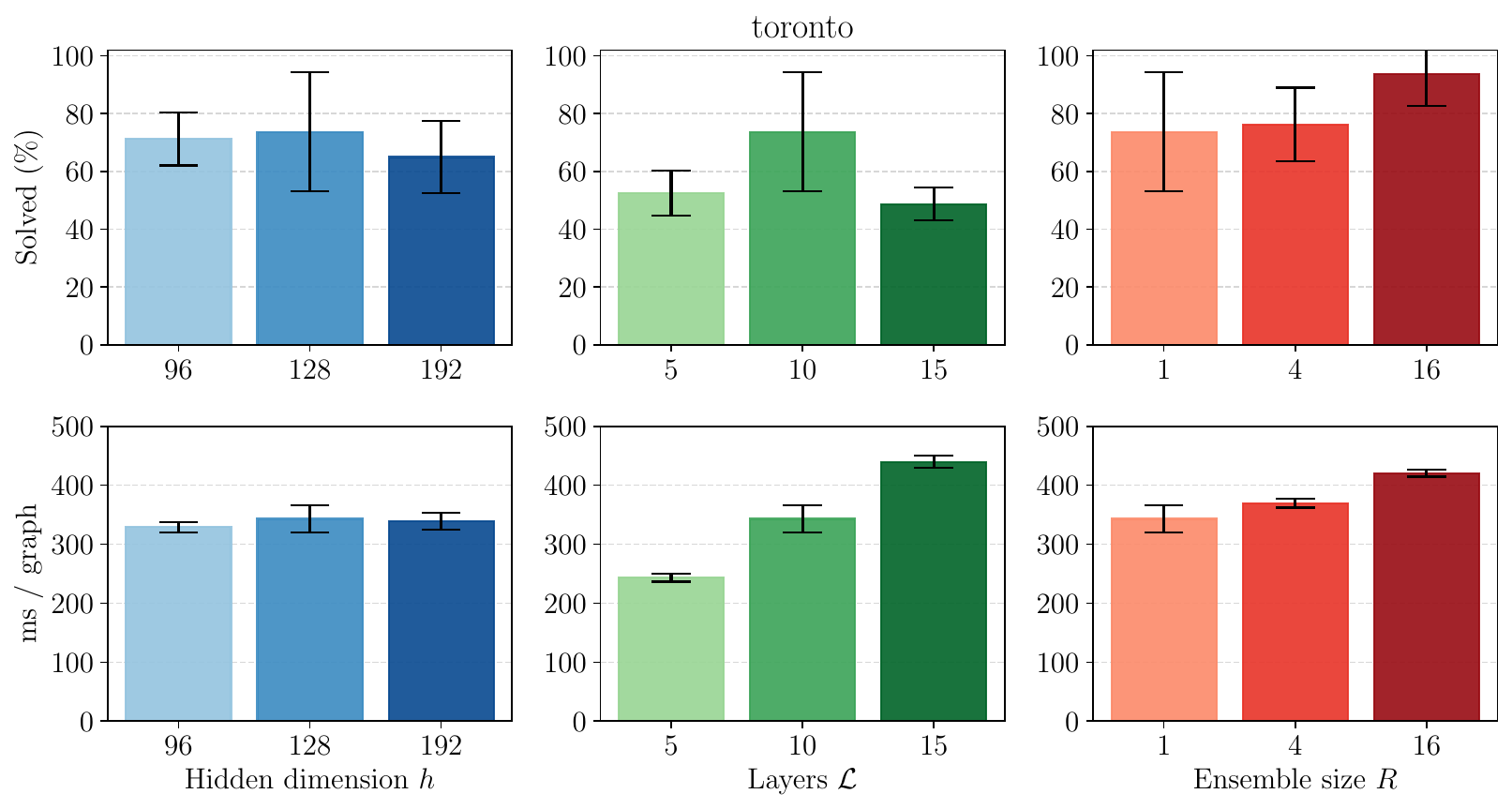} \vspace{-0.5ex}
    \caption{Validation Solved\% and average inference times on \texttt{toronto} for different values of embedding dimension $d$, message-passing layers $\mathcal{L}$ and ensemble size $E$.}
    \label{fig:sweep_toronto}
\end{figure}

Figures~\ref{fig:sweep_sudoku} and~\ref{fig:sweep_toronto} show that both solution quality and inference time vary only marginally with $d$, although $d=128$ achieves the best quality. Sudoku benefits from greater depth, while Toronto performs best at $\mathcal{L}=10$. As expected, inference time grows approximately linearly with $\mathcal{L}$, with a particularly noticeable effect on Toronto. Increasing the decode width consistently improves solution quality in these experiments, while also increasing inference time approximately linearly. Based on these results, we fix $\mathcal{L}=10$ and $d=128$ throughout, as they provide a good trade-off between solution quality and inference time across datasets. Tuning these values separately for each dataset could yield slightly better results but would require additional training.

\section{Additional experimental results} \label{sec:additional_exp}

In this section, we report some additional experimental results that were omitted from the main text for space reasons. 

\paragraph{Transfer across Sudokus.}
First, we evaluate \textsc{Scone} across Sudoku benchmarks in a transfer-learning setting. In the main results, the model trained on \texttt{sudoku\_17} was evaluated in-distribution and transferred to \texttt{sudoku\_satnet}, \texttt{sudoku\_rrn}, and \texttt{sudoku\_top1465}, while a separate model was trained and evaluated on \texttt{sudoku\_extreme}. Here, we evaluate both checkpoints on all five datasets. We additionally consider a checkpoint trained on the training split of \texttt{sudoku\_satnet}, which was used, e.g., in \citet{xu2025selfsupervised}.

Table~\ref{tab:sudoku_transfer} reports the results for the three checkpoints, decoded with up to $B=512$ beams. The model trained on \texttt{sudoku\_satnet} transfers least effectively, failing to solve every instance on several benchmarks and requiring on average wider beams, which then increases inference time. As expected, each checkpoint performs best on its training distribution, where it solves every instance with the lowest inference time. Under transfer, the models trained on \texttt{sudoku\_17} and \texttt{sudoku\_extreme} perform similarly, although the former is slightly stronger, possibly because it was trained on more instances.

\begin{table}[h!]
\centering
\resizebox{\textwidth}{!}{%
{\setlength{\tabcolsep}{6pt}
\begin{tabular}{l*{5}{cc}}
\toprule
\multirow{2}{*}{Method} & \multicolumn{2}{c}{sudoku\_17} & \multicolumn{2}{c}{sudoku\_satnet} & \multicolumn{2}{c}{sudoku\_rrn} & \multicolumn{2}{c}{sudoku\_top1465} & \multicolumn{2}{c}{sudoku\_extreme} \\
\cmidrule(lr){2-3} \cmidrule(lr){4-5} \cmidrule(lr){6-7} \cmidrule(lr){8-9} \cmidrule(lr){10-11}
 & Solved\% & Time (s) & Solved\% & Time (s) & Solved\% & Time (s) & Solved\% & Time (s) & Solved\% & Time (s) \\
\midrule
\textsc{Ours} (sudoku\_17) & \trb{100.00}{0.00} & \tr{0.009}{0.000} & \trb{100.00}{0.00} & \tr{0.000}{0.000} & \trb{100.00}{0.00} & \tr{0.003}{0.000} & \trb{100.00}{0.00} & \tr{0.017}{0.000} & \trb{100.00}{0.00} & \tr{0.054}{0.000} \\
\textsc{Ours} (sudoku\_extreme) & \tr{99.94}{0.01} & \tr{0.043}{0.000} & \trb{100.00}{0.00} & \tr{0.000}{0.000} & \trb{100.00}{0.00} & \tr{0.005}{0.000} & \trb{100.00}{0.00} & \tr{0.022}{0.000} & \trb{100.00}{0.00} & \tr{0.020}{0.000} \\
\textsc{Ours} (sudoku\_satnet) & \tr{99.85}{0.00} & \tr{0.057}{0.000} & \trb{100.00}{0.00} & \tr{0.000}{0.000} & \tr{99.98}{0.00} & \tr{0.007}{0.000} & \trb{100.00}{0.00} & \tr{0.023}{0.001} & \tr{99.97}{0.00} & \tr{0.059}{0.000} \\
\bottomrule
\end{tabular}%
}
}
\caption{Solution quality on Sudoku benchmarks with multiple checkpoints: Solved\% ($\uparrow$) and inference time per graph ($\downarrow$).}
\label{tab:sudoku_transfer}
\end{table}

\textbf{Per-instance results.}$\ $
We then report the per-instance results for the Toronto timetabling and queens benchmarks. Each of these instances corresponds to a single graph, so pooling instances into one column, although necessary due to space constraints, averages over problems of substantially different difficulty. Tables~\ref{tab:toronto_small_per_instance}, \ref{tab:toronto_large_per_instance}, and~\ref{tab:queens_per_instance} decompose the three pooled columns of Table~\ref{tab:toronto_queens_cycles} into their constituent graphs. Since each column now represents one graph, Solved\% denotes the fraction of 20 repeated runs that find a proper coloring, rather than the fraction of graphs solved.

Consistent with the pooled results, \textsc{Scone} and \textsc{AnyCSP} are closely matched on the smaller Toronto instances, with each method holding a slight advantage on some graphs and falling behind on others. On the larger Toronto instances, \textsc{Scone} consistently achieves both better solution quality and lower inference times. On queens, the two methods are essentially tied for sizes $8$ and $9$, while \textsc{Scone} is faster for size $10$.

\begin{table}[h!]
\centering
\resizebox{0.95\textwidth}{!}{%
{\setlength{\tabcolsep}{6pt}
\begin{tabular}{l*{4}{ccc}}
\toprule
\multirow{2}{*}{Method} & \multicolumn{3}{c}{ear-f-83} & \multicolumn{3}{c}{tre-s-92} & \multicolumn{3}{c}{lse-f-91} & \multicolumn{3}{c}{kfu-s-93} \\
 & \multicolumn{3}{c}{$n{=}159$, $k{=}22$} & \multicolumn{3}{c}{$n{=}204$, $k{=}20$} & \multicolumn{3}{c}{$n{=}132$, $k{=}17$} & \multicolumn{3}{c}{$n{=}196$, $k{=}19$} \\
 \cmidrule(lr){2-4} \cmidrule(lr){5-7} \cmidrule(lr){8-10} \cmidrule(lr){11-13}
 & Solved\% & Mono\% & Time (s) & Solved\% & Mono\% & Time (s) & Solved\% & Mono\% & Time (s) & Solved\% & Mono\% & Time (s) \\
\midrule
\textsc{Greedy} & \tr{0.0}{0.0} & \tr{0.38}{0.00} & \tr{0.00}{0.00} & \tr{0.0}{0.0} & \tr{0.49}{0.00} & \tr{0.00}{0.00} & \tr{0.0}{0.0} & \tr{0.38}{0.00} & \tr{0.00}{0.00} & \tr{0.0}{0.0} & \tr{0.35}{0.00} & \tr{0.00}{0.00} \\
\textsc{DSatur} & \tr{0.0}{0.0} & \tr{0.02}{0.00} & \tr{0.00}{0.00} & \tr{0.0}{0.0} & \tr{0.07}{0.00} & \tr{0.00}{0.00} & \trb{100.0}{0.0} & \trb{0.00}{0.00} & \tr{0.00}{0.00} & \trb{100.0}{0.0} & \trb{0.00}{0.00} & \tr{0.00}{0.00} \\
\textsc{ILP} & \tr{0.0}{0.0} & \tr{0.04}{0.00} & \tr{600.00}{0.00} & \tr{0.0}{0.0} & \tr{0.15}{0.00} & \tr{600.00}{0.00} & \trb{100.0}{0.0} & \trb{0.00}{0.00} & \tr{51.10}{0.00} & \trb{100.0}{0.0} & \trb{0.00}{0.00} & \tr{64.60}{0.00} \\
\textsc{ACE} & \trb{100.0}{0.0} & \trb{0.00}{0.00} & \tr{0.75}{0.03} & \trb{100.0}{0.0} & \trb{0.00}{0.00} & \tr{0.76}{0.01} & \trb{100.0}{0.0} & \trb{0.00}{0.00} & \tr{0.60}{0.01} & \trb{100.0}{0.0} & \trb{0.00}{0.00} & \tr{0.69}{0.03} \\
\midrule
\textsc{PI-SAGE} & \tr{0.0}{0.0} & \tr{0.64}{0.10} & \tr{6.89}{0.19} & \tr{0.0}{0.0} & \tr{0.82}{0.06} & \tr{7.52}{0.46} & \tr{0.0}{0.0} & \tr{0.62}{0.04} & \tr{6.90}{1.02} & \tr{0.0}{0.0} & \tr{0.55}{0.03} & \tr{7.06}{0.39} \\
\textsc{ROS} & \tr{0.0}{0.0} & \tr{0.58}{0.06} & \tr{0.63}{0.10} & \tr{0.0}{0.0} & \tr{0.62}{0.06} & \tr{0.75}{0.12} & \tr{0.0}{0.0} & \tr{0.47}{0.08} & \tr{0.61}{0.05} & \tr{0.0}{0.0} & \tr{0.41}{0.06} & \tr{0.65}{0.07} \\
\textsc{AnyCSP} & \trb{100.0}{0.0} & \trb{0.00}{0.00} & \tr{1.12}{0.05} & \trb{100.0}{0.0} & \trb{0.00}{0.00} & \tr{1.45}{0.09} & \trb{100.0}{0.0} & \trb{0.00}{0.00} & \tr{0.43}{0.10} & \trb{100.0}{0.0} & \trb{0.00}{0.00} & \tr{0.75}{0.02} \\
\textsc{ConsFormer} & \tr{0.0}{0.0} & \tr{1.71}{0.09} & \tr{24.59}{0.09} & \tr{0.0}{0.0} & \tr{2.40}{0.08} & \tr{24.86}{0.12} & \tr{0.0}{0.0} & \tr{2.02}{0.16} & \tr{24.91}{0.12} & \tr{0.0}{0.0} & \tr{1.98}{0.08} & \tr{24.97}{0.10} \\
\midrule
\textsc{Ours} & \trb{100.0}{0.0} & \trb{0.00}{0.00} & \tr{0.67}{0.01} & \trb{100.0}{0.0} & \trb{0.00}{0.00} & \tr{0.90}{0.01} & \trb{100.0}{0.0} & \trb{0.00}{0.00} & \tr{0.64}{0.00} & \trb{100.0}{0.0} & \trb{0.00}{0.00} & \tr{0.84}{0.00} \\
\bottomrule
\end{tabular}%
}
}
\caption{Per-instance breakdown of the results on \texttt{toronto(small)} instances. Solved\% ($\uparrow$), Mono\% ($\downarrow$) and inference time per graph ($\downarrow$).}
\label{tab:toronto_small_per_instance}
\end{table}

\begin{table}[h!]
\centering
\resizebox{\textwidth}{!}{%
{\setlength{\tabcolsep}{6pt}
\begin{tabular}{l*{5}{ccc}}
\toprule
\multirow{2}{*}{Method} & \multicolumn{3}{c}{rye-s-93} & \multicolumn{3}{c}{car-f-92} & \multicolumn{3}{c}{uta-s-92} & \multicolumn{3}{c}{car-s-91} & \multicolumn{3}{c}{pur-s-93} \\
 & \multicolumn{3}{c}{$n{=}236$, $k{=}21$} & \multicolumn{3}{c}{$n{=}413$, $k{=}27$} & \multicolumn{3}{c}{$n{=}482$, $k{=}30$} & \multicolumn{3}{c}{$n{=}556$, $k{=}28$} & \multicolumn{3}{c}{$n{=}1374$, $k{=}35$} \\
 \cmidrule(lr){2-4} \cmidrule(lr){5-7} \cmidrule(lr){8-10} \cmidrule(lr){11-13} \cmidrule(lr){14-16}
 & Solved\% & Mono\% & Time (s) & Solved\% & Mono\% & Time (s) & Solved\% & Mono\% & Time (s) & Solved\% & Mono\% & Time (s) & Solved\% & Mono\% & Time (s) \\
\midrule
\textsc{Greedy} & \tr{0.0}{0.0} & \tr{0.29}{0.00} & \tr{0.00}{0.00} & \tr{0.0}{0.0} & \tr{0.49}{0.00} & \tr{0.00}{0.00} & \tr{0.0}{0.0} & \tr{0.24}{0.00} & \tr{0.00}{0.00} & \tr{0.0}{0.0} & \tr{0.50}{0.00} & \tr{0.01}{0.00} & \tr{0.0}{0.0} & \tr{0.21}{0.00} & \tr{0.02}{0.00} \\
\textsc{DSatur} & \tr{0.0}{0.0} & \tr{0.08}{0.00} & \tr{0.00}{0.00} & \tr{0.0}{0.0} & \tr{0.10}{0.00} & \tr{0.01}{0.00} & \tr{0.0}{0.0} & \trb{0.00}{0.00} & \tr{0.01}{0.00} & \tr{0.0}{0.0} & \tr{0.03}{0.00} & \tr{0.02}{0.00} & \trb{100.0}{0.0} & \trb{0.00}{0.00} & \tr{0.07}{0.00} \\
\textsc{ILP} & \trb{100.0}{0.0} & \trb{0.00}{0.00} & \tr{395.10}{0.00} & \tr{0.0}{0.0} & \tr{0.43}{0.00} & \tr{600.00}{0.00} & \tr{0.0}{0.0} & \tr{0.54}{0.00} & \tr{600.00}{0.00} & \tr{0.0}{0.0} & \tr{0.82}{0.00} & \tr{600.00}{0.00} & \tr{0.0}{0.0} & \tr{0.93}{0.00} & \tr{600.00}{0.00} \\
\textsc{ACE} & \trb{100.0}{0.0} & \trb{0.00}{0.00} & \tr{0.70}{0.02} & \trb{100.0}{0.0} & \trb{0.00}{0.00} & \tr{306.01}{7.03} & \trb{100.0}{0.0} & \trb{0.00}{0.00} & \tr{2.06}{0.13} & \trb{100.0}{0.0} & \trb{0.00}{0.00} & \tr{11.24}{0.69} & \trb{100.0}{0.0} & \trb{0.00}{0.00} & \tr{9.87}{1.22} \\
\midrule
\textsc{PI-SAGE} & \tr{0.0}{0.0} & \tr{0.45}{0.07} & \tr{7.47}{0.27} & \tr{0.0}{0.0} & \tr{0.70}{0.04} & \tr{10.53}{1.09} & \tr{0.0}{0.0} & \tr{0.58}{0.02} & \tr{10.45}{0.66} & \tr{0.0}{0.0} & \tr{0.76}{0.02} & \tr{12.12}{1.08} & \tr{0.0}{0.0} & \tr{0.40}{0.01} & \tr{29.78}{4.35} \\
\textsc{ROS} & \tr{0.0}{0.0} & \tr{0.37}{0.04} & \tr{0.65}{0.04} & \tr{0.0}{0.0} & \tr{0.68}{0.04} & \tr{0.84}{0.12} & \tr{0.0}{0.0} & \tr{0.52}{0.04} & \tr{0.75}{0.08} & \tr{0.0}{0.0} & \tr{0.72}{0.04} & \tr{0.83}{0.09} & \tr{0.0}{0.0} & \tr{0.38}{0.01} & \tr{1.01}{0.21} \\
\textsc{AnyCSP} & \trb{100.0}{0.0} & \trb{0.00}{0.00} & \tr{1.22}{0.09} & \tr{60.0}{49.0} & \trb{0.00}{0.00} & \tr{29.11}{5.07} & \trb{100.0}{0.0} & \trb{0.00}{0.00} & \tr{18.56}{3.27} & \tr{10.0}{30.0} & \tr{0.02}{0.01} & \tr{35.90}{1.97} & \tr{80.0}{40.0} & \trb{0.00}{0.00} & \tr{44.76}{4.78} \\
\textsc{ConsFormer} & \tr{0.0}{0.0} & \tr{1.77}{0.08} & \tr{24.82}{0.09} & \tr{0.0}{0.0} & \tr{1.98}{0.05} & \tr{24.94}{0.28} & \tr{0.0}{0.0} & \tr{1.71}{0.03} & \tr{24.56}{0.12} & \tr{0.0}{0.0} & \tr{2.11}{0.05} & \tr{24.75}{0.10} & -- & -- & -- \\
\midrule
\textsc{Ours} & \trb{100.0}{0.0} & \trb{0.00}{0.00} & \tr{1.38}{0.63} & \trb{100.0}{0.0} & \trb{0.00}{0.00} & \tr{6.34}{3.19} & \trb{100.0}{0.0} & \trb{0.00}{0.00} & \tr{3.14}{1.48} & \trb{100.0}{0.0} & \trb{0.00}{0.00} & \tr{3.42}{1.52} & \trb{100.0}{0.0} & \trb{0.00}{0.00} & \tr{10.10}{0.05} \\
\bottomrule
\end{tabular}%
}
}
\caption{Per-instance breakdown of the results on \texttt{toronto(large)} instances. Solved\% ($\uparrow$), Mono\% ($\downarrow$) and inference time per graph ($\downarrow$).}
\label{tab:toronto_large_per_instance}
\end{table}

\begin{table}[t!]
\centering
\resizebox{0.9\textwidth}{!}{%
{\setlength{\tabcolsep}{6pt}
\begin{tabular}{l*{3}{ccc}}
\toprule
\multirow{2}{*}{Method} & \multicolumn{3}{c}{queens\_8\_8} & \multicolumn{3}{c}{queens\_9\_9} & \multicolumn{3}{c}{queens\_10\_10} \\
 & \multicolumn{3}{c}{$n{=}64$, $k{=}9$} & \multicolumn{3}{c}{$n{=}81$, $k{=}10$} & \multicolumn{3}{c}{$n{=}100$, $k{=}11$} \\
 \cmidrule(lr){2-4} \cmidrule(lr){5-7} \cmidrule(lr){8-10}
 & Solved\% & Mono\% & Time (s) & Solved\% & Mono\% & Time (s) & Solved\% & Mono\% & Time (s) \\
\midrule
\textsc{Greedy} & \tr{0.0}{0.0} & \tr{1.24}{0.00} & \tr{0.00}{0.00} & \tr{0.0}{0.0} & \tr{1.52}{0.00} & \tr{0.00}{0.00} & \tr{0.0}{0.0} & \tr{1.56}{0.00} & \tr{0.00}{0.00} \\
\textsc{DSatur} & \tr{0.0}{0.0} & \tr{1.24}{0.00} & \tr{0.00}{0.00} & \tr{0.0}{0.0} & \tr{1.04}{0.00} & \tr{0.00}{0.00} & \tr{0.0}{0.0} & \tr{1.43}{0.00} & \tr{0.00}{0.00} \\
\textsc{ILP} & \trb{100.0}{0.0} & \trb{0.00}{0.00} & \tr{9.99}{0.00} & \trb{100.0}{0.0} & \trb{0.00}{0.00} & \tr{2.94}{0.00} & \trb{100.0}{0.0} & \trb{0.00}{0.00} & \tr{499.53}{0.00} \\
\textsc{ACE} & \trb{100.0}{0.0} & \trb{0.00}{0.00} & \tr{2.05}{0.02} & \trb{100.0}{0.0} & \trb{0.00}{0.00} & \tr{20.06}{0.15} & \tr{0.0}{0.0} & -- & \tr{597.54}{0.07} \\
\midrule
\textsc{PI-SAGE} & \tr{0.0}{0.0} & \tr{1.88}{0.33} & \tr{5.43}{0.66} & \tr{0.0}{0.0} & \tr{1.83}{0.29} & \tr{5.39}{0.46} & \tr{0.0}{0.0} & \tr{1.67}{0.19} & \tr{6.13}{0.78} \\
\textsc{ROS} & \tr{0.0}{0.0} & \tr{1.45}{0.32} & \tr{0.69}{0.14} & \tr{0.0}{0.0} & \tr{1.37}{0.20} & \tr{0.67}{0.01} & \tr{0.0}{0.0} & \tr{1.26}{0.21} & \tr{0.71}{0.04} \\
\textsc{AnyCSP} & \trb{100.0}{0.0} & \trb{0.00}{0.00} & \tr{0.42}{0.23} & \trb{100.0}{0.0} & \trb{0.00}{0.00} & \tr{0.63}{0.39} & \tr{90.0}{30.0} & \tr{0.01}{0.02} & \tr{10.82}{7.22} \\
\textsc{ConsFormer} & \tr{30.0}{47.0} & \tr{0.10}{0.08} & \tr{20.40}{7.14} & \tr{5.0}{22.4} & \tr{0.13}{0.06} & \tr{24.21}{2.67} & \tr{0.0}{0.0} & \tr{2.78}{0.42} & \tr{24.90}{0.17} \\
\midrule
\textsc{Ours} & \trb{100.0}{0.0} & \trb{0.00}{0.00} & \tr{0.38}{0.01} & \trb{100.0}{0.0} & \trb{0.00}{0.00} & \tr{0.39}{0.02} & \trb{100.0}{0.0} & \trb{0.00}{0.00} & \tr{0.84}{0.41} \\
\bottomrule
\end{tabular}%
}
}
\caption{Per-instance breakdown of the results on \texttt{queens(8-10)} instances. Solved\% ($\uparrow$), Mono\% ($\downarrow$) and inference time per graph ($\downarrow$).}
\label{tab:queens_per_instance}
\end{table}

\paragraph{Synthetic structured graphs.}
Finally, we report results on synthetic structured graphs to examine potential failure modes of the different methods. In addition to the cycle benchmarks presented in the main paper, we consider complete graphs with up to $64$ nodes, regular bipartite grid graphs, a separate collection containing only even cycles, and the Mycielski graphs $M_5$, $M_6$, and $M_7$. The latter are triangle-free despite having chromatic numbers $5$, $6$, and $7$, respectively.
As a stress test, \textsc{Scone} uses width $1$ with forced assignments disabled. Its performance therefore reflects the learned policy itself, without beam search, ensembling, or constraint propagation.

\begin{table}[h!]
\centering
\resizebox{\textwidth}{!}{%
{\setlength{\tabcolsep}{6pt}
\begin{tabular}{l*{4}{ccc}}
\toprule
\multirow{2}{*}{Method} & \multicolumn{3}{c}{complete} & \multicolumn{3}{c}{grid} & \multicolumn{3}{c}{even\_cycles} & \multicolumn{3}{c}{myciel(5-7)} \\
\cmidrule(lr){2-4} \cmidrule(lr){5-7} \cmidrule(lr){8-10} \cmidrule(lr){11-13}
 & Solved\% & Mono\% & Time (s) & Solved\% & Mono\% & Time (s) & Solved\% & Mono\% & Time (s) & Solved\% & Mono\% & Time (s) \\
\midrule
\textsc{Greedy} & \trb{100.00}{0.00} & \trb{0.00}{0.00} & \tr{0.00}{0.00} & \trb{100.00}{0.00} & \trb{0.00}{0.00} & \tr{0.00}{0.00} & \trb{100.00}{0.00} & \trb{0.00}{0.00} & \tr{0.00}{0.00} & \trb{100.00}{0.00} & \trb{0.00}{0.00} & \tr{0.00}{0.00} \\
\textsc{DSatur} & \trb{100.00}{0.00} & \trb{0.00}{0.00} & \tr{0.00}{0.00} & \trb{100.00}{0.00} & \trb{0.00}{0.00} & \tr{0.00}{0.00} & \trb{100.00}{0.00} & \trb{0.00}{0.00} & \tr{0.00}{0.00} & \trb{100.00}{0.00} & \trb{0.00}{0.00} & \tr{0.00}{0.00} \\
\textsc{ILP} & \trb{100.00}{0.00} & \trb{0.00}{0.00} & \tr{0.07}{0.00} & \trb{100.00}{0.00} & \trb{0.00}{0.00} & \tr{0.01}{0.00} & \trb{100.00}{0.00} & \trb{0.00}{0.00} & \tr{0.01}{0.00} & \trb{100.00}{0.00} & \trb{0.00}{0.00} & \tr{0.01}{0.00} \\
\textsc{ACE} & \trb{100.00}{0.00} & \trb{0.00}{0.00} & \tr{0.54}{0.00} & \trb{100.00}{0.00} & \trb{0.00}{0.00} & \tr{0.47}{0.00} & \trb{100.00}{0.00} & \trb{0.00}{0.00} & \tr{0.45}{0.00} & \trb{100.00}{0.00} & \trb{0.00}{0.00} & \tr{0.50}{0.01} \\
\midrule
\textsc{PI-SAGE} & \tr{98.75}{1.53} & \trb{0.00}{0.00} & \tr{4.24}{0.01} & \tr{2.18}{1.88} & \tr{7.18}{0.26} & \tr{4.71}{0.06} & \tr{0.00}{0.00} & \tr{8.96}{0.23} & \tr{3.77}{0.01} & \tr{40.00}{38.87} & \tr{0.12}{0.11} & \tr{5.14}{0.30} \\
\textsc{ROS} & \trb{100.00}{0.00} & \trb{0.00}{0.00} & \tr{0.30}{0.00} & \tr{10.31}{2.90} & \tr{4.37}{0.24} & \tr{1.03}{0.00} & \tr{0.00}{0.00} & \tr{8.18}{0.04} & \tr{0.81}{0.00} & \tr{60.00}{13.33} & \tr{0.03}{0.02} & \tr{0.73}{0.07} \\
\textsc{AnyCSP} & \trb{100.00}{0.00} & \trb{0.00}{0.00} & \tr{0.55}{0.02} & \trb{100.00}{0.00} & \trb{0.00}{0.00} & \tr{0.20}{0.01} & \trb{100.00}{0.00} & \trb{0.00}{0.00} & \tr{0.36}{0.02} & \trb{100.00}{0.00} & \trb{0.00}{0.00} & \tr{0.08}{0.00} \\
\textsc{ConsFormer} & -- & -- & -- & \tr{50.7}{4.7} & \tr{0.95}{0.15} & \tr{2.16}{0.00} & \tr{19.1}{4.1} & \tr{0.68}{0.04} & \tr{2.19}{0.01} & \trb{100.0}{0.0} & \trb{0.00}{0.00} & \tr{0.08}{0.16} \\
\midrule
\textsc{Ours} & \trb{100.00}{0.00} & \trb{0.00}{0.00} & \tr{0.02}{0.00} & \trb{100.00}{0.00} & \trb{0.00}{0.00} & \tr{0.03}{0.00} & \trb{100.00}{0.00} & \trb{0.00}{0.00} & \tr{0.03}{0.00} & \trb{100.00}{0.00} & \trb{0.00}{0.00} & \tr{0.36}{0.00} \\
\bottomrule
\end{tabular}%
}
}
\caption{Solution quality on additional structured graphs: Solved\% ($\uparrow$), Mono\% ($\downarrow$) and inference time per graph ($\downarrow$).}
\label{tab:structured}
\end{table}

\textsc{Scone} and the classical heuristics solve every instance. \textsc{AnyCSP} also solves all four families and is particularly fast on the sparse Mycielski graphs. However, it takes roughly an order of magnitude longer than \textsc{Scone} on grids and even cycles. These connected bipartite graphs have an essentially unique $2$-coloring up to color permutations and require consistency to be propagated over long distances, which may explain the additional search. \textsc{ConsFormer} cannot be evaluated on the complete graphs because their palettes contain up to $64$ colors, whereas its architecture does not support palette sizes beyond the maximum of $30$ seen during training. Moreover, it can only solve roughly half of the grid graphs, and few even-size cycles.
Finally, the relaxation-based methods frequently fail on grids and Mycielski graphs, and none solves any even cycle.

\paragraph{Additional baselines}

Due to space constraints, we report some comparisons with three additional baselines here. Results are reported in Tables~\ref{tab:add_baselines_sudoku} and~\ref{tab:add_baselines_toronto}. \textsc{CSO} (Cat Swarm Optimization) \citep{saeed2024comparative} is a population-based metaheuristic without a learned component, alternating between local seeking moves and tracing moves guided by the best solution found. \textsc{GBS} \citep{arora2026graph} constructs color classes through GNN-guided extraction of large independent sets, while \textsc{VColRL} \citep{anand2025vcolrl} uses reinforcement learning to assign colors to multiple vertices per step, reverting conflicting assignments. Since \textsc{GBS} and \textsc{VColRL} minimize the number of colors, we adapt them to our fixed palette by retaining their first $k$ color classes and assigning each remaining node its least-conflicting color among them. All three methods receive the same nominal per-graph budgets as \textsc{AnyCSP}: 5\,s on Sudoku and registers, and 30\,s on Toronto and queens. 
\textsc{VColRL} repeats episodes within the budget, retains the best colouring, and stops once a proper $k$-colouring is found.
\textsc{GBS} performs a single search per graph, capped at the budget.
None of these methods solves an instance from Toronto or \texttt{sudoku\_17}. \textsc{GBS} solves \texttt{queens\_8\_8} in 35\% of repeats and \texttt{queens\_9\_9} in 5\%.

\begin{table}[h!]
\centering 
\resizebox{\textwidth}{!}{%
{\setlength{\tabcolsep}{6pt}
\begin{tabular}{l*{4}{ccc}}
\toprule
\multirow{2}{*}{Method} & \multicolumn{3}{c}{sudoku\_17} & \multicolumn{3}{c}{sudoku\_satnet} & \multicolumn{3}{c}{sudoku\_rrn} & \multicolumn{3}{c}{sudoku\_top1465} \\
\cmidrule(lr){2-4} \cmidrule(lr){5-7} \cmidrule(lr){8-10} \cmidrule(lr){11-13}
 & Solved\% & Mono\% & Time (s) & Solved\% & Mono\% & Time (s) & Solved\% & Mono\% & Time (s) & Solved\% & Mono\% & Time (s) \\
\midrule
\textsc{CSO} & \tr{0.0}{0.0} & \tr{1.73}{0.01} & \tr{5.00}{0.00} & \tr{53.6}{0.6} & \tr{0.54}{0.01} & \tr{3.04}{0.02} & \tr{0.7}{0.2} & \tr{1.47}{0.01} & \tr{4.98}{0.01} & \tr{0.0}{0.0} & \tr{1.56}{0.01} & \tr{5.00}{0.00} \\
\textsc{GBS} & \tr{0.0}{0.0} & \tr{1.12}{0.01} & \tr{4.12}{0.01} & \tr{57.8}{0.4} & \tr{0.22}{0.00} & \tr{0.16}{0.00} & \tr{1.2}{0.1} & \tr{1.09}{0.01} & \tr{2.62}{0.01} & \tr{0.0}{0.0} & \tr{0.92}{0.01} & \tr{2.69}{0.02} \\
\textsc{VColRL} & \tr{0.0}{0.0} & \tr{0.42}{0.00} & \tr{5.07}{0.04} & \tr{82.3}{0.1} & \tr{0.05}{0.00} & \tr{1.47}{0.01} & \tr{13.7}{0.3} & \tr{0.37}{0.01} & \tr{4.75}{0.02} & \tr{0.0}{0.0} & \tr{0.43}{0.00} & \tr{5.21}{0.04} \\
\midrule
\textsc{Ours} & \trb{100.0}{0.0} & \trb{0.00}{0.00} & \tr{0.01}{0.00} & \trb{100.0}{0.0} & \trb{0.00}{0.00} & \tr{0.00}{0.00} & \trb{100.0}{0.0} & \trb{0.00}{0.00} & \tr{0.00}{0.00} & \trb{100.0}{0.0} & \trb{0.00}{0.00} & \tr{0.02}{0.00} \\
\bottomrule
\end{tabular}%
}
}
\caption{ Solution quality of additional baselines on Sudoku-graph coloring.}
\label{tab:add_baselines_sudoku}
\end{table}

\begin{table}[h!]
\centering 
\resizebox{\textwidth}{!}{%
{\setlength{\tabcolsep}{6pt}
\begin{tabular}{l*{4}{ccc}}
\toprule
\multirow{2}{*}{Method} & \multicolumn{3}{c}{toronto (small)} & \multicolumn{3}{c}{toronto (large)} & \multicolumn{3}{c}{queens} & \multicolumn{3}{c}{registers} \\
\cmidrule(lr){2-4} \cmidrule(lr){5-7} \cmidrule(lr){8-10} \cmidrule(lr){11-13}
 & Solved\% & Mono\% & Time (s) & Solved\% & Mono\% & Time (s) & Solved\% & Mono\% & Time (s) & Solved\% & Mono\% & Time (s) \\
\midrule
\textsc{CSO} & \tr{0.0}{0.0} & \tr{0.51}{0.04} & \tr{30.00}{0.01} & \tr{0.0}{0.0} & \tr{0.51}{0.01} & \tr{30.01}{0.01} & \tr{0.0}{0.0} & \tr{0.95}{0.13} & \tr{30.00}{0.00} & \tr{93.26}{0.16} & \tr{0.14}{0.01} & \tr{0.45}{0.01} \\
\textsc{GBS} & \tr{0.0}{0.0} & \tr{0.16}{0.02} & \tr{30.17}{0.06} & \tr{0.0}{0.0} & \tr{0.11}{0.00} & \tr{30.34}{0.06} & \tr{13.3}{16.3} & \tr{0.49}{0.13} & \tr{24.97}{0.36} & \tr{93.39}{0.06} & \tr{0.10}{0.00} & \tr{1.91}{0.01} \\
\textsc{VColRL} & \tr{0.0}{0.0} & \tr{0.50}{0.03} & \tr{31.61}{0.52} & \tr{0.0}{0.0} & \tr{0.43}{0.01} & \tr{33.05}{0.56} & \tr{0.0}{0.0} & \tr{0.94}{0.07} & \tr{30.98}{0.30} & \tr{96.38}{0.35} & \tr{0.07}{0.01} & \tr{0.70}{0.02} \\
\midrule
\textsc{Ours} & \trb{100.0}{0.0} & \trb{0.00}{0.00} & \tr{0.76}{0.00} & \trb{100.0}{0.0} & \trb{0.00}{0.00} & \tr{4.88}{0.82} & \trb{100.0}{0.0} & \trb{0.00}{0.00} & \tr{0.54}{0.14} & \trb{100.00}{0.00} & \trb{0.00}{0.00} & \tr{0.00}{0.00} \\
\bottomrule
\end{tabular}%
}
}
\caption{ Solution quality of additional baselines on isolated instances and registers. }
\label{tab:add_baselines_toronto}
\end{table}

\paragraph{Additional datasets}

Our main evaluations focus on graph instances drawn from practical applications.
Here, we additionally evaluate on three DIMACS\footnote{https://mat.tepper.cmu.edu/COLOR/instances.html} families: book graphs connecting characters that encounter one another, Erd\H{o}s--R\'enyi \texttt{DSJC} graphs, and random \texttt{le450} graphs with guaranteed $k$-colorings.
On these graphs, we use the same synthetically-trained \textsc{Scone} checkpoint used in Table~\ref{tab:toronto_queens_cycles}, and we use ensembling. 

Table~\ref{tab:dimacs_books} reports the results on book graphs, where $k$ is set to their chromatic number. These graphs are relatively easy to color, and \textsc{Scone} solves 100\% of them. However, for such simple instances, heuristics like \textsc{Dsatur} might be preferable, as they also solve them while being faster.

\begin{table}[h!]
\centering 
\resizebox{\textwidth}{!}{%
{\setlength{\tabcolsep}{6pt}
\begin{tabular}{l*{5}{ccc}} 
\toprule
\multirow{2}{*}{Method} & \multicolumn{3}{c}{anna} & \multicolumn{3}{c}{david} & \multicolumn{3}{c}{homer} & \multicolumn{3}{c}{huck} & \multicolumn{3}{c}{jean} \\
\cmidrule(lr){2-4} \cmidrule(lr){5-7} \cmidrule(lr){8-10} \cmidrule(lr){11-13} \cmidrule(lr){14-16}
 & \multicolumn{3}{c}{$n{=}138$, $k{=}11$} & \multicolumn{3}{c}{$n{=}87$, $k{=}11$} & \multicolumn{3}{c}{$n{=}561$, $k{=}13$} & \multicolumn{3}{c}{$n{=}74$, $k{=}11$} & \multicolumn{3}{c}{$n{=}80$, $k{=}10$} \\
 & Solved\% & Mono\% & Time (s) & Solved\% & Mono\% & Time (s) & Solved\% & Mono\% & Time (s) & Solved\% & Mono\% & Time (s) & Solved\% & Mono\% & Time (s) \\
\midrule
\textsc{Greedy} & \tr{80.0}{40.0} & \tr{0.04}{0.08} & \tr{0.00}{0.00} & \tr{20.0}{40.0} & \tr{0.20}{0.10} & \tr{0.00}{0.00} & \tr{60.0}{49.0} & \tr{0.04}{0.05} & \tr{0.00}{0.00} & \trb{100.0}{0.0} & \trb{0.00}{0.00} & \tr{0.00}{0.00} & \tr{80.0}{40.0} & \tr{0.08}{0.16} & \tr{0.00}{0.00} \\
\textsc{DSatur} & \trb{100.0}{0.0} & \trb{0.00}{0.00} & \tr{0.00}{0.00} & \trb{100.0}{0.0} & \trb{0.00}{0.00} & \tr{0.00}{0.00} & \trb{100.0}{0.0} & \trb{0.00}{0.00} & \tr{0.01}{0.00} & \trb{100.0}{0.0} & \trb{0.00}{0.00} & \tr{0.00}{0.00} & \trb{100.0}{0.0} & \trb{0.00}{0.00} & \tr{0.00}{0.00} \\
\textsc{RLF} & \trb{100.0}{0.0} & \trb{0.00}{0.00} & \tr{0.00}{0.00} & \trb{100.0}{0.0} & \trb{0.00}{0.00} & \tr{0.00}{0.00} & \trb{100.0}{0.0} & \trb{0.00}{0.00} & \tr{0.01}{0.00} & \trb{100.0}{0.0} & \trb{0.00}{0.00} & \tr{0.00}{0.00} & \trb{100.0}{0.0} & \trb{0.00}{0.00} & \tr{0.00}{0.00} \\
\textsc{ILP} & \trb{100.0}{0.0} & \trb{0.00}{0.00} & \tr{0.13}{0.00} & \trb{100.0}{0.0} & \trb{0.00}{0.00} & \tr{0.12}{0.00} & \trb{100.0}{0.0} & \trb{0.00}{0.00} & \tr{0.30}{0.00} & \trb{100.0}{0.0} & \trb{0.00}{0.00} & \tr{0.11}{0.00} & \trb{100.0}{0.0} & \trb{0.00}{0.00} & \tr{0.12}{0.00} \\
\textsc{ACE} & \trb{100.0}{0.0} & \trb{0.00}{0.00} & \tr{0.77}{0.00} & \trb{100.0}{0.0} & \trb{0.00}{0.00} & \tr{0.74}{0.00} & \trb{100.0}{0.0} & \trb{0.00}{0.00} & \tr{0.51}{0.00} & \trb{100.0}{0.0} & \trb{0.00}{0.00} & \tr{0.72}{0.01} & \trb{100.0}{0.0} & \trb{0.00}{0.00} & \tr{0.41}{0.00} \\
\midrule
\textsc{AnyCSP} & \trb{100.0}{0.0} & \trb{0.00}{0.00} & \tr{0.10}{0.15} & \trb{100.0}{0.0} & \trb{0.00}{0.00} & \tr{0.07}{0.06} & \trb{100.0}{0.0} & \trb{0.00}{0.00} & \tr{0.24}{0.11} & \trb{100.0}{0.0} & \trb{0.00}{0.00} & \tr{0.05}{0.03} & \trb{100.0}{0.0} & \trb{0.00}{0.00} & \tr{0.06}{0.04} \\
\textsc{ConsFormer} & \tr{0.0}{0.0} & \tr{1.96}{0.28} & \tr{24.27}{0.31} & \tr{0.0}{0.0} & \tr{2.28}{0.41} & \tr{24.84}{0.10} & \tr{0.0}{0.0} & \tr{2.08}{0.23} & \tr{24.38}{0.11} & \tr{0.0}{0.0} & \tr{1.83}{0.40} & \tr{24.76}{0.08} & \trb{100.0}{0.0} & \trb{0.00}{0.00} & \tr{0.13}{0.34} \\
\midrule
\textsc{Ours} & \trb{100.0}{0.0} & \trb{0.00}{0.00} & \tr{0.75}{0.00} & \trb{100.0}{0.0} & \trb{0.00}{0.00} & \tr{0.35}{0.00} & \trb{100.0}{0.0} & \trb{0.00}{0.00} & \tr{2.27}{0.00} & \trb{100.0}{0.0} & \trb{0.00}{0.00} & \tr{0.29}{0.00} & \trb{100.0}{0.0} & \trb{0.00}{0.00} & \tr{0.32}{0.00} \\
\bottomrule
\end{tabular}%
}
}
\caption{Solution quality on book graphs: Solved\% ($\uparrow$), Mono\% ($\downarrow$) and inference time ($\downarrow$).}
\label{tab:dimacs_books}
\end{table}

Table~\ref{tab:dimacs_dsjc} reports the results on \texttt{DSJC} graphs, where $k$ is set to the best-known feasible palette size. These graphs are much harder to solve compared to book graphs. \textsc{Scone} solves \texttt{DSJC125.1} and \texttt{DSJC250.1} in all runs, but reaches a 20\% solve rate on \texttt{DSJC125.9}. \textsc{AnyCSP} solves the latter consistently, while \textsc{Scone} is faster and more reliable on \texttt{DSJC250.1}. None of the evaluated methods solves \texttt{DSJC125.5}.

\begin{table}[h!]
\centering 
\resizebox{\textwidth}{!}{%
{\setlength{\tabcolsep}{6pt}
\begin{tabular}{l*{4}{ccc}}
\toprule
\multirow{2}{*}{Method} & \multicolumn{3}{c}{DSJC125.1} & \multicolumn{3}{c}{DSJC125.5} & \multicolumn{3}{c}{DSJC125.9} & \multicolumn{3}{c}{DSJC250.1} \\
\cmidrule(lr){2-4} \cmidrule(lr){5-7} \cmidrule(lr){8-10} \cmidrule(lr){11-13}
 & \multicolumn{3}{c}{$n{=}125$, $k{=}5$} & \multicolumn{3}{c}{$n{=}125$, $k{=}17$} & \multicolumn{3}{c}{$n{=}125$, $k{=}44$} & \multicolumn{3}{c}{$n{=}250$, $k{=}8$} \\
 & Solved\% & Mono\% & Time (s) & Solved\% & Mono\% & Time (s) & Solved\% & Mono\% & Time (s) & Solved\% & Mono\% & Time (s) \\
\midrule
\textsc{Greedy} & \tr{0.0}{0.0} & \tr{3.32}{0.32} & \tr{0.00}{0.00} & \tr{0.0}{0.0} & \tr{0.89}{0.04} & \tr{0.00}{0.00} & \tr{0.0}{0.0} & \tr{0.30}{0.03} & \tr{0.00}{0.00} & \tr{0.0}{0.0} & \tr{2.03}{0.29} & \tr{0.00}{0.00} \\
\textsc{DSatur} & \tr{0.0}{0.0} & \tr{1.36}{0.00} & \tr{0.00}{0.00} & \tr{0.0}{0.0} & \tr{0.98}{0.00} & \tr{0.00}{0.00} & \tr{0.0}{0.0} & \tr{0.29}{0.00} & \tr{0.00}{0.00} & \tr{0.0}{0.0} & \tr{1.55}{0.00} & \tr{0.00}{0.00} \\
\textsc{RLF} & \tr{0.0}{0.0} & \tr{1.36}{0.00} & \tr{0.02}{0.02} & \tr{0.0}{0.0} & \tr{0.51}{0.00} & \tr{0.02}{0.02} & \tr{0.0}{0.0} & \tr{0.19}{0.00} & \tr{0.04}{0.01} & \tr{0.0}{0.0} & \tr{0.84}{0.00} & \tr{0.07}{0.09} \\
\textsc{ILP} & \trb{100.0}{0.0} & \trb{0.00}{0.00} & \tr{24.98}{0.00} & \tr{0.0}{0.0} & \tr{0.57}{0.00} & \tr{600.22}{0.00} & \tr{0.0}{0.0} & \tr{0.06}{0.00} & \tr{600.42}{0.00} & \tr{0.0}{0.0} & \tr{0.44}{0.00} & \tr{600.22}{0.00} \\
\textsc{ACE} & \trb{100.0}{0.0} & \trb{0.00}{0.00} & \tr{0.90}{0.02} & \tr{0.0}{0.0} & -- & \tr{594.29}{0.40} & \tr{0.0}{0.0} & -- & \tr{593.98}{0.17} & \tr{0.0}{0.0} & -- & \tr{594.80}{0.82} \\
\midrule
\textsc{AnyCSP} & \trb{100.0}{0.0} & \trb{0.00}{0.00} & \tr{0.43}{0.20} & \tr{0.0}{0.0} & \trb{0.06}{0.02} & \tr{30.66}{0.12} & \trb{100.0}{0.0} & \trb{0.00}{0.00} & \tr{9.71}{5.14} & \tr{90.0}{30.0} & \trb{0.00}{0.01} & \tr{13.90}{9.45} \\
\textsc{ConsFormer} & \tr{20.0}{40.0} & \tr{0.11}{0.05} & \tr{23.38}{4.99} & \tr{0.0}{0.0} & \tr{2.73}{0.21} & \tr{24.66}{0.18} & -- & -- & -- & \tr{0.0}{0.0} & \tr{0.17}{0.04} & \tr{24.97}{0.12} \\
\textsc{ReLCol} & \tr{0.0}{0.0} & \tr{1.78}{0.12} & \tr{30.54}{0.08} & \tr{0.0}{0.0} & \tr{0.65}{0.05} & \tr{30.74}{0.15} & \tr{0.0}{0.0} & \tr{0.22}{0.01} & \tr{30.50}{0.10} & \tr{0.0}{0.0} & \tr{1.52}{0.00} & \tr{33.25}{0.68} \\
\midrule
\textsc{Ours} & \trb{100.0}{0.0} & \trb{0.00}{0.00} & \tr{0.76}{0.01} & \tr{0.0}{0.0} & \tr{0.17}{0.03} & \tr{7.10}{0.20} & \tr{20.0}{40.0} & \tr{0.01}{0.01} & \tr{18.96}{4.75} & \trb{100.0}{0.0} & \trb{0.00}{0.00} & \tr{4.47}{2.63} \\
\bottomrule
\end{tabular}%
}
}
\caption{ Solution quality on \texttt{DSJC} graphs: Solved\% ($\uparrow$), Mono\% ($\downarrow$) and inference time ($\downarrow$).}
\label{tab:dimacs_dsjc}
\end{table}

Tables~\ref{tab:dimacs_le450_15} and \ref{tab:dimacs_le450_25} report the results on \texttt{le450} graphs, with the palette sizes set to $k=15$ and $k=25$ respectively. \textsc{Scone} solves all 15-color instances, including \texttt{le450\_15c} and \texttt{le450\_15d}, where \textsc{AnyCSP} succeeds in only 5\% of runs and all other reported baselines fail. For 25 colors, \textsc{Scone} and several baselines consistently solve the \texttt{a/b} variants, while the \texttt{c/d} variants remain unsolved by all methods. Nevertheless, \textsc{Scone} achieves the lowest monochromatic-edge rates on these two instances.

\begin{table}[h!]
\centering 
\resizebox{\textwidth}{!}{%
{\setlength{\tabcolsep}{6pt}
\begin{tabular}{l*{4}{ccc}}
\toprule
\multirow{2}{*}{Method} & \multicolumn{3}{c}{le450\_15a} & \multicolumn{3}{c}{le450\_15b} & \multicolumn{3}{c}{le450\_15c} & \multicolumn{3}{c}{le450\_15d} \\
\cmidrule(lr){2-4} \cmidrule(lr){5-7} \cmidrule(lr){8-10} \cmidrule(lr){11-13}
 & \multicolumn{3}{c}{$n{=}450$, $k{=}15$} & \multicolumn{3}{c}{$n{=}450$, $k{=}15$} & \multicolumn{3}{c}{$n{=}450$, $k{=}15$} & \multicolumn{3}{c}{$n{=}450$, $k{=}15$} \\
 & Solved\% & Mono\% & Time (s) & Solved\% & Mono\% & Time (s) & Solved\% & Mono\% & Time (s) & Solved\% & Mono\% & Time (s) \\
\midrule
\textsc{Greedy} & \tr{0.0}{0.0} & \tr{0.85}{0.04} & \tr{0.00}{0.00} & \tr{0.0}{0.0} & \tr{0.85}{0.06} & \tr{0.00}{0.00} & \tr{0.0}{0.0} & \tr{1.92}{0.05} & \tr{0.00}{0.00} & \tr{0.0}{0.0} & \tr{2.00}{0.02} & \tr{0.00}{0.00} \\
\textsc{DSatur} & \tr{0.0}{0.0} & \tr{0.17}{0.00} & \tr{0.01}{0.00} & \tr{0.0}{0.0} & \tr{0.09}{0.00} & \tr{0.01}{0.00} & \tr{0.0}{0.0} & \tr{1.90}{0.00} & \tr{0.01}{0.00} & \tr{0.0}{0.0} & \tr{1.90}{0.00} & \tr{0.01}{0.00} \\
\textsc{RLF} & \tr{0.0}{0.0} & \tr{0.16}{0.00} & \tr{0.02}{0.00} & \tr{0.0}{0.0} & \tr{0.15}{0.00} & \tr{0.02}{0.00} & \tr{0.0}{0.0} & \tr{1.51}{0.00} & \tr{0.04}{0.00} & \tr{0.0}{0.0} & \tr{1.58}{0.00} & \tr{0.04}{0.00} \\
\textsc{ILP} & \tr{0.0}{0.0} & \tr{0.37}{0.00} & \tr{600.27}{0.00} & \tr{0.0}{0.0} & \tr{0.38}{0.00} & \tr{600.23}{0.00} & \tr{0.0}{0.0} & \tr{2.59}{0.00} & \tr{600.73}{0.00} & \tr{0.0}{0.0} & \tr{2.49}{0.00} & \tr{600.42}{0.00} \\
\textsc{ACE} & \trb{100.0}{0.0} & \trb{0.00}{0.00} & \tr{4.45}{0.06} & \trb{100.0}{0.0} & \trb{0.00}{0.00} & \tr{2.67}{0.11} & \tr{0.0}{0.0} & -- & \tr{593.83}{0.26} & \tr{0.0}{0.0} & -- & \tr{594.00}{0.40} \\
\midrule
\textsc{AnyCSP} & \trb{100.0}{0.0} & \trb{0.00}{0.00} & \tr{10.02}{4.31} & \trb{100.0}{0.0} & \trb{0.00}{0.00} & \tr{5.81}{2.52} & \tr{5.0}{21.8} & \tr{0.35}{0.50} & \tr{31.70}{1.65} & \tr{5.0}{21.8} & \tr{0.19}{0.38} & \tr{31.36}{2.89} \\
\textsc{ConsFormer} & \tr{0.0}{0.0} & \tr{2.84}{0.17} & \tr{24.42}{0.21} & \tr{0.0}{0.0} & \tr{2.81}{0.18} & \tr{24.68}{0.34} & \tr{0.0}{0.0} & \tr{3.18}{0.18} & \tr{24.65}{0.18} & \tr{0.0}{0.0} & \tr{3.13}{0.12} & \tr{25.06}{0.20} \\
\midrule
\textsc{Ours} & \trb{100.0}{0.0} & \trb{0.00}{0.00} & \tr{1.74}{0.01} & \trb{100.0}{0.0} & \trb{0.00}{0.00} & \tr{1.73}{0.01} & \trb{100.0}{0.0} & \trb{0.00}{0.00} & \tr{5.65}{5.11} & \trb{100.0}{0.0} & \trb{0.00}{0.00} & \tr{8.54}{5.89} \\
\bottomrule
\end{tabular}%
}
}
\caption{Solution quality on \texttt{le450\_15} graphs: Solved\% ($\uparrow$), Mono\% ($\downarrow$) and inference time ($\downarrow$).}
\label{tab:dimacs_le450_15}
\end{table}

\begin{table}[h!]
\centering 
\resizebox{\textwidth}{!}{%
{\setlength{\tabcolsep}{6pt}
\begin{tabular}{l*{4}{ccc}}
\toprule
\multirow{2}{*}{Method} & \multicolumn{3}{c}{le450\_25a} & \multicolumn{3}{c}{le450\_25b} & \multicolumn{3}{c}{le450\_25c} & \multicolumn{3}{c}{le450\_25d} \\
\cmidrule(lr){2-4} \cmidrule(lr){5-7} \cmidrule(lr){8-10} \cmidrule(lr){11-13}
 & \multicolumn{3}{c}{$n{=}450$, $k{=}25$} & \multicolumn{3}{c}{$n{=}450$, $k{=}25$} & \multicolumn{3}{c}{$n{=}450$, $k{=}25$} & \multicolumn{3}{c}{$n{=}450$, $k{=}25$} \\
 & Solved\% & Mono\% & Time (s) & Solved\% & Mono\% & Time (s) & Solved\% & Mono\% & Time (s) & Solved\% & Mono\% & Time (s) \\
\midrule
\textsc{Greedy} & \tr{0.0}{0.0} & \tr{0.13}{0.02} & \tr{0.00}{0.00} & \tr{0.0}{0.0} & \tr{0.10}{0.02} & \tr{0.00}{0.00} & \tr{0.0}{0.0} & \tr{0.57}{0.01} & \tr{0.00}{0.00} & \tr{0.0}{0.0} & \tr{0.55}{0.02} & \tr{0.00}{0.00} \\
\textsc{DSatur} & \trb{100.0}{0.0} & \trb{0.00}{0.00} & \tr{0.01}{0.00} & \trb{100.0}{0.0} & \trb{0.00}{0.00} & \tr{0.01}{0.00} & \tr{0.0}{0.0} & \tr{0.36}{0.00} & \tr{0.01}{0.00} & \tr{0.0}{0.0} & \tr{0.22}{0.00} & \tr{0.01}{0.00} \\
\textsc{RLF} & \trb{100.0}{0.0} & \trb{0.00}{0.00} & \tr{0.02}{0.00} & \trb{100.0}{0.0} & \trb{0.00}{0.00} & \tr{0.02}{0.00} & \tr{0.0}{0.0} & \tr{0.20}{0.00} & \tr{0.05}{0.00} & \tr{0.0}{0.0} & \tr{0.22}{0.00} & \tr{0.05}{0.00} \\
\textsc{ILP} & \trb{100.0}{0.0} & \trb{0.00}{0.00} & \tr{4.66}{0.00} & \trb{100.0}{0.0} & \trb{0.00}{0.00} & \tr{8.68}{0.00} & \tr{0.0}{0.0} & \tr{0.53}{0.00} & \tr{600.69}{0.00} & \tr{0.0}{0.0} & \tr{0.51}{0.00} & \tr{600.70}{0.00} \\
\textsc{ACE} & \trb{100.0}{0.0} & \trb{0.00}{0.00} & \tr{1.92}{0.07} & \trb{100.0}{0.0} & \trb{0.00}{0.00} & \tr{1.85}{0.09} & \tr{0.0}{0.0} & -- & \tr{593.69}{0.30} & \tr{0.0}{0.0} & -- & \tr{593.78}{0.30} \\
\midrule
\textsc{AnyCSP} & \trb{100.0}{0.0} & \trb{0.00}{0.00} & \tr{1.83}{0.02} & \trb{100.0}{0.0} & \trb{0.00}{0.00} & \tr{1.82}{0.03} & \tr{0.0}{0.0} & \tr{0.22}{0.01} & \tr{33.56}{0.13} & \tr{0.0}{0.0} & \tr{0.22}{0.01} & \tr{33.58}{0.04} \\
\textsc{ConsFormer} & \tr{0.0}{0.0} & \tr{2.04}{0.05} & \tr{24.52}{0.11} & \tr{0.0}{0.0} & \tr{2.07}{0.04} & \tr{25.01}{0.09} & \tr{0.0}{0.0} & \tr{2.56}{0.04} & \tr{24.46}{0.10} & \tr{0.0}{0.0} & \tr{2.56}{0.04} & \tr{24.46}{0.11} \\
\midrule
\textsc{Ours} & \trb{100.0}{0.0} & \trb{0.00}{0.00} & \tr{2.47}{0.00} & \trb{100.0}{0.0} & \trb{0.00}{0.00} & \tr{2.49}{0.00} & \tr{0.0}{0.0} & \trb{0.06}{0.01} & \tr{21.05}{0.34} & \tr{0.0}{0.0} & \trb{0.05}{0.01} & \tr{21.34}{0.25} \\
\bottomrule
\end{tabular}%
}
}
\caption{Solution quality on \texttt{le450\_25} graphs: Solved\% ($\uparrow$), Mono\% ($\downarrow$) and inference time ($\downarrow$).}
\label{tab:dimacs_le450_25}
\end{table}

\paragraph{Searching for the best feasible palette size}

In all main experiments, \textsc{Scone} takes the palette size $k$ as input, as is natural in several applications with a fixed resource budget. This setting is also considered by \textsc{AnyCSP} and \textsc{ConsFormer}, and is standard in CSP and ILP formulations of graph coloring. 
Here, we also consider the setting where $k$ is not prescribed, and the goal is to obtain a coloring using as few colors as possible. Starting from a \textsc{Dsatur} coloring, we repeatedly attempt a proper coloring with one fewer color, stopping at the first failure and retaining the best proper coloring found.
Retaining the best proper coloring makes this an anytime procedure. 

\begin{figure}[h]
    \centering
    \includegraphics[width=\linewidth]{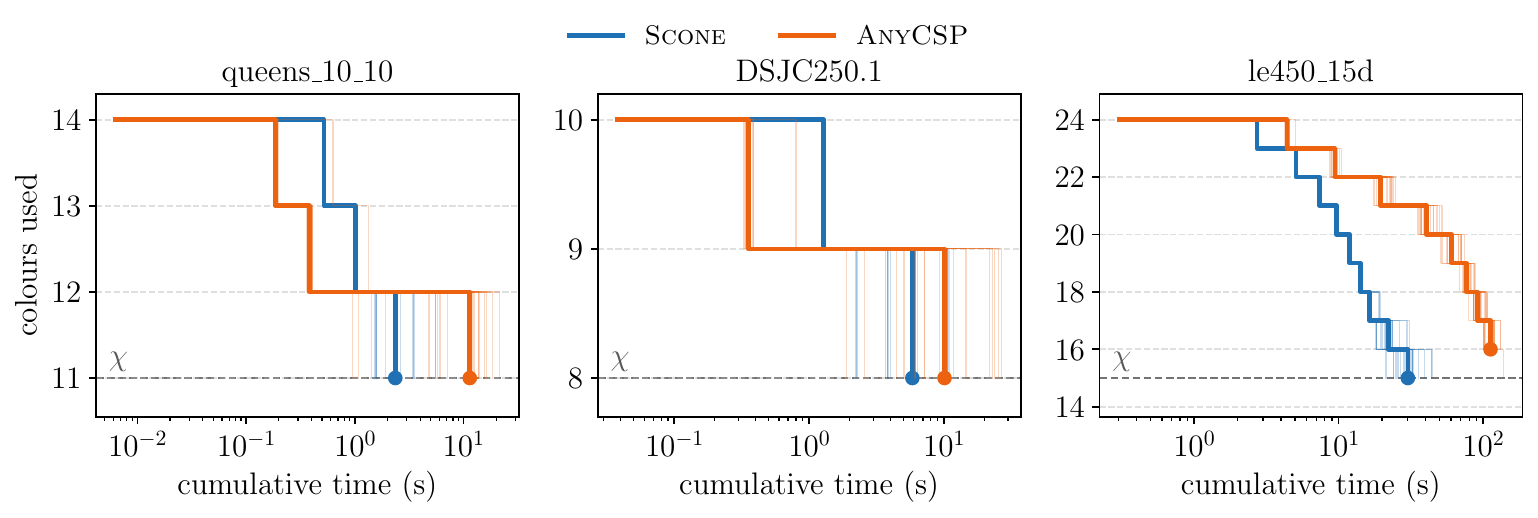}
    \caption{ Improving chromatic upper bounds from a \textsc{Dsatur} coloring. Curves show cumulative time to successive proper colorings over 20 independent searches. Dashed lines mark the best-known upper bounds $\chi$.}
    \label{fig:chi_search}
\end{figure}

For this experiment, we consider three hard instances, \texttt{queens\_10\_10}, \texttt{DSJC250.1} and \texttt{le450\_15d}, and use \textsc{AnyCSP} as a reference point, as it is the strongest neural baseline.
Figure~\ref{fig:chi_search} compares 20 independent searches per method against cumulative wall-clock time. \textsc{Scone} reaches the best-known color counts in every search: from 14 to 11 on \texttt{queens\_10\_10}, from 10 to 8 on \texttt{DSJC250.1}, and from 24 to 15 on \texttt{le450\_15d}. Under the same search protocol, with a 30\,s per-budget limit, \textsc{AnyCSP} reaches these counts only in 18 of 20 searches on each of the first two instances, taking slightly longer than \textsc{Scone}. On \texttt{le450\_15d}, only one of its searches reaches the 15 colors, while most runs can only reach a coloring with 16. Moreover, it takes significantly longer than \textsc{Scone}.

\paragraph{Infeasible palettes.}

While our training procedure and main experiments focus on producing proper colorings, we also examine the number of monochromatic edges when the available palette makes a proper coloring impossible.
As a proof of concept, we consider register-interference graphs, where insufficient registers can necessitate spilling. To test at different levels of infeasibility, we set $k=\chi(G)-1$ or $k=\lceil\chi(G)/2\rceil$ and evaluate the resulting approximate colorings using Mono\%. 
As reference points, we use \textsc{AnyCSP} with a timelimit of 1\,s, which is reached on each instance, and our ILP formulation with a maximum timelimit of 5\,s.
Our \textsc{ACE} encoding searches for proper colorings and can certify infeasibility, but does not minimize monochromatic edges, so it is omitted from this comparison. We run \textsc{Scone} with 256 replicas.
On the easier $k=\chi(G)-1$, the ILP with a timelimit of 5\,s certifies optimality on 97\% of the graphs. Both \textsc{Scone} and \textsc{AnyCSP} achieve Mono\% values comparable to the ILP, with the former being slightly better than the latter and twice as fast. 
On the harder $k=\lceil\chi(G)/2\rceil$ setting, the ILP can only certify optimality on 85\% of the graphs. 
Here, \textsc{Scone} produces better solutions than both baselines, while also having the lowest inference times. 
Since \textsc{Scone} is trained on proper colorings, specializing its training to infeasible palettes and conflict minimization may further improve approximate solution quality.

\begin{table}[h!]
\centering 
{\setlength{\tabcolsep}{6pt}
\begin{tabular}{l*{2}{cc}}
\toprule
\multirow{2}{*}{Method} & \multicolumn{2}{c}{\texttt{registers} ($\chi-1$)} & \multicolumn{2}{c}{\texttt{registers} ($\lceil\chi/2\rceil$)} \\
\cmidrule(lr){2-3} \cmidrule(lr){4-5}
 & Mono\% & Time (s) & Mono\% & Time (s) \\
\midrule
\textsc{ILP} & \trb{0.793}{0.000} & \tr{0.245}{0.000} & \tr{4.528}{0.000} & \tr{1.034}{0.000} \\
\textsc{AnyCSP} & \tr{0.797}{0.003} & \tr{1.017}{0.004} & \tr{4.523}{0.003} & \tr{1.013}{0.003} \\
\midrule
\textsc{Ours} & \tr{0.795}{0.001} & \tr{0.432}{0.000} & \trb{4.469}{0.003} & \tr{0.248}{0.000} \\
\bottomrule
\end{tabular}
}
\caption{ Mono\% ($\downarrow$) and inference time ($\downarrow$) on \texttt{registers} with infeasible palettes.}
\label{tab:infeasible}
\end{table}

\color{black}

\paragraph{Additional details on the equivariance ablations}
The ablations in the main paper compare \textsc{Scone} with a sequential model that does not enforce color equivariance, both with and without color-permutation augmentation. Here, we provide additional details on the experimental protocol and capacity controls. All variants use the training configuration described in Sec.~\ref{sec:setup} and the same most-constrained-first decoding procedure, ensuring that the color representation is the only difference between the primary models.
Each configuration is trained three times, we then report the mean and standard deviation of the evaluations across training runs. 
The augmented and unaugmented models share the same data pipeline. Data augmentation samples a fresh uniformly random color permutation on every fetch. Augmentation does not change the number of graph presentations or optimization updates.

For the \texttt{sudoku\_17} experiment in Fig.~\ref{fig:ablation}, we hold the training budget fixed at $10^6$ graph presentations for every subset size $N$, setting the number of epochs to $10^6/N$. We evaluate the validation objective at $100$ equally spaced points during training, so all subset sizes receive the same number of checkpoint-selection opportunities. All models are evaluated on the same held-out test set.
For the experiments varying the target chromatic number, we generate graphs with a planted $\chi$-coloring and $n\in[128,512]$ vertices.
The planted coloring certifies that $\chi(G)\leq\chi$, but not equality. Since \textsc{Dsatur} fails to find a $\chi$-coloring for every retained graph, the datasets comprise challenging, predominantly tight instances as the palette size $k$ grows.
For $\chi\in\{3,5,7,9\}$, we generate $2000$ graphs and split them into $80\%/10\%/10\%$ training, validation, and test instances. We use that same synthetic generator as described in Section~\ref{sec:datasets}, fixing the number of colors to be exactly $\chi$. For each dataset, the model is trained with $k = \chi$.

\begin{figure}[h!]
    \centering
    \includegraphics[width=0.99\linewidth]{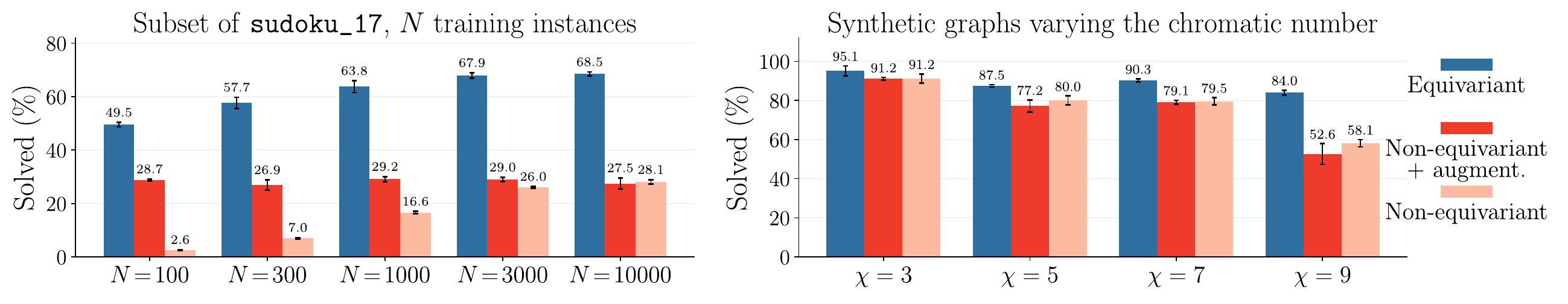}
    \caption{Ablation study on Sudoku and synthetic graphs at equal hidden width ($d=128$).}
    \label{fig:equivariance-colors-h128}
\end{figure}

Note that at equal hidden width ($d=128$), the non-equivariant model has roughly half the number of trainable parameters than \textsc{Scone}. For the experiments reported in the main paper in Figure~\ref{fig:ablation}, we therefore set the non-equivariant model at width $d=256$, giving the baseline approximately twice as many parameters as \textsc{Scone}. 
For completeness, Figure~\ref{fig:equivariance-colors-h128} gives the full results when all models use hidden width $d=128$, complementing the results in Figure~\ref{fig:ablation}.  
Nevertheless, the equal-width and parameter-favorable comparisons yield the same qualitative trends.

\paragraph{Scalability}

\begin{figure}[h]
    \centering
    \includegraphics[width=0.5\linewidth]{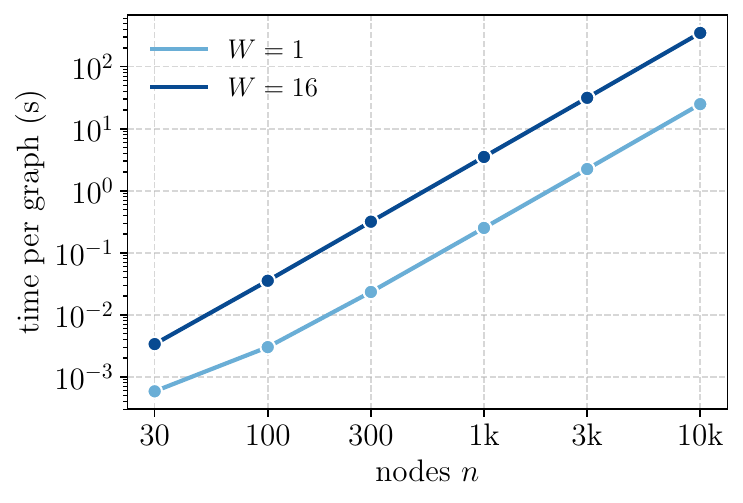}
    \caption{ Inference time per graph versus node count on random graphs, using greedy decoding ($W=1$) and beam search ($W=16$).}
    \label{fig:scalability}
\end{figure}

We study scalability on random graphs with 30 to 10,000 nodes, using greedy decoding ($W=1$) and beam search ($W=16$). As reported in Figure~\ref{fig:scalability},
runtime grows approximately quadratically over the tested range. This is in line with the theory, as for a graph with $n$ nodes \textsc{Scone} takes $O(n)$ forward passes to color each node sequentially, and each forward pass takes linear time in the number of edges, which is linear in $n$ for sparse graphs. 
Greedy decoding remains below one second at 1,000 nodes but reaches tens of seconds at 10,000 nodes.

\end{document}